\documentclass[runningheads]{llncs}

\usepackage{eccv}

\usepackage{eccvabbrv}
\usepackage[table]{xcolor}
\usepackage{multirow}
\usepackage{graphicx}
\usepackage{booktabs}
\usepackage{makecell}
\usepackage{multirow}
\usepackage{subcaption}
\usepackage{caption}   % 推荐
\usepackage{tabularx}

\usepackage[accsupp]{axessibility}  % Improves PDF readability for those with disabilities.

\usepackage{hyperref}

\usepackage{orcidlink}

\begin{document}

\title{Spectral Probing of Feature Upsamplers in 2D-to-3D Scene Reconstruction} 

\titlerunning{Abbreviated paper title}


\author{
Ling Xiao\inst{1}\orcidlink{0000-0002-4650-8841} \and
Yuliang Xiu\inst{2} \orcidlink{0000-0003-0165-5909} \and
Yue Chen\inst{2} \and
Guoming Wang\inst{3} \and
Toshihiko Yamasaki\inst{4} \orcidlink{0000-0002-1784-2314}
}

\authorrunning{L. Xiao et al.}

\institute{
Hokkaido University, Sapporo, Japan\\
\email{ling@ist.hokudai.ac.jp}
\and
Westlake University, Hangzhou, China\\
\email{\{xiuyuliang,faneggchen\}@westlake.edu.cn}
\and
Zhejiang University, Hangzhou, China\\
\email{nb21013@zju.edu.cn}
\and
The University of Tokyo, Tokyo, Japan\\
\email{yamasaki@cvm.t.u-tokyo.ac.jp}
}

% Springer Heidelberg, Tiergartenstr.~17, 69121 Heidelberg, Germany

\maketitle

\begin{abstract}
A typical 2D-to-3D pipeline takes multi-view images as input, where a Vision Foundation Model (VFM) extracts features that are spatially upsampled to dense representations for 3D reconstruction. 
If dense features across views preserve geometric consistency, differentiable rendering can recover an accurate 3D representation, making the feature upsampler a critical component. 
Recent learnable upsampling methods mainly aim to enhance spatial details, such as sharper geometry or richer textures, yet their impact on 3D awareness remains underexplored. 
To address this gap, we introduce a spectral diagnostic framework with six complementary metrics that characterize amplitude redistribution, structural spectral alignment, and directional stability.
Across classical interpolation and learnable upsampling methods on CLIP and DINO backbones, we observe three key findings. 
First, structural spectral consistency (SSC/CSC) is the strongest predictor of NVS quality, whereas High-Frequency Spectral Slope Drift (HFSS) often correlates negatively with reconstruction performance, indicating that emphasizing high-frequency details alone does not necessarily improve 3D reconstruction.
Second, geometry and texture respond to different spectral properties: Angular Energy Consistency (ADC) correlates more strongly with geometry-related metrics, while SSC/CSC influence texture fidelity slightly more than geometric accuracy. 
Third, although learnable upsamplers often produce sharper spatial features, they rarely outperform classical interpolation in reconstruction quality, and their effectiveness depends on the reconstruction model.
Overall, our results indicate that reconstruction quality is more closely related to preserving spectral structure than to enhancing spatial detail, highlighting spectral consistency as an important principle for designing upsampling strategies in 2D-to-3D pipelines.
\end{abstract}

\section{Introduction}
\label{sec:intro}
2D-to-3D scene reconstruction has become a central paradigm for modern 3D modeling. 
Given multi-view images, a visual backbone extracts image representations that encode spatial correspondences across views. 
Since patch-based encoders produce coarse representations, feature upsampling is required to recover dense spatial features before constructing neural scene representations such as Gaussian-based models or implicit fields. 
These representations are optimized through differentiable rendering to enforce geometric and photometric consistency across views, and reconstruction quality is typically evaluated via novel view synthesis (NVS) using metrics such as PSNR, SSIM, and LPIPS.

Feature upsampling therefore plays a critical role in recovering high-resolution representations~\cite{fu2024featup,suri2024lift,huang2025loftup,couairon2025jafar,wimmer2025anyup}. 
Existing methods can be broadly categorized into classical and learnable approaches. 
Classical techniques include bilinear, nearest-neighbor, bicubic, and Lanczos interpolation, while learnable upsamplers include FeatUp~\cite{fu2024featup}, LoftUp~\cite{huang2025loftup}, LiFT~\cite{suri2024lift}, JAFAR~\cite{couairon2025jafar}, and AnyUp~\cite{wimmer2025anyup}. 
Existing learnable feature upsamplers mainly focus on enhancing spatial details, such as sharper boundaries, richer textures, or higher pixel-level fidelity. 
However, whether such improvements benefit 3D-aware inference remains largely unexplored.
Feat2GS~\cite{chen2025feat2gs} provides a unified and model-agnostic framework for probing the 3D awareness through NVS by disentangling geometry, texture, and overall 3D awareness.

Building on Feat2GS~\cite{chen2025feat2gs}, we introduce a spectral diagnosis framework with six complementary metrics that characterize amplitude redistribution, structural spectral alignment, and directional stability, enabling systematic comparison between classical interpolation and learnable upsampling methods. 
We also introduce Non-cropping Spatial Matching (NSM), a simple baseline that isolates the effect of spatial interpolation.
This work investigates three key questions:
(1) Is the common assumption of learnable upsamplers, enhancing spatial details such as sharper geometry or richer textures, sufficient for 2D-to-3D reconstruction?
(2) Which spectral characteristics most strongly predict 3D awareness across the joint (All), geometry, and texture probing modes?
(3) How do classical interpolation methods compare with learnable upsamplers in 3D reconstruction performance?
We provide the following contributions:

\begin{itemize}
\item We introduce a spectral diagnosis framework that characterizes amplitude redistribution, structural spectral alignment, and directional stability, enabling a systematic analysis of how feature upsampling modifies spectral structure in 2D-to-3D reconstruction pipelines.

\item We conduct a systematic comparison of classical interpolation and learnable upsampling methods, analyzing how their spectral transformations influence 3D awareness under overall, geometry-only, and texture-only probing modes.

\item Our analysis reveals three key insights that challenge the conventional spatial-detail intuition of feature upsampling:
(i) Structural spectral consistency (SSC/CSC) is the strongest predictor of NVS quality, while High-Frequency Spectral Slope Drift (HFSS) often correlates negatively with reconstruction performance;
(ii) geometry and texture depend on different spectral properties: Angular Energy Consistency (ADC) correlates more strongly with geometry-related metrics, whereas SSC/CSC correlate more consistently with texture fidelity;
(iii) although learnable upsamplers often produce sharper spatial features, they rarely outperform classical interpolation in reconstruction quality, and their effectiveness depends on the reconstruction model.
\end{itemize}

\section{Related Works}
\label{sec:related}

\subsection{Feature Upsampling}
Many methods have been proposed to convert low-resolution features into high-resolution representations. Classical approaches include bilinear, nearest-neighbor, bicubic, and Lanczos interpolation, which rely on fixed kernels for resampling. 
Joint Bilateral Upsampling (JBU)~\cite{kopf2007joint} refines features using image guidance but may introduce blur and interpolation artifacts. 
Other directions, such as GAN-based densification~\cite{tan2018feature} and distillation-based upsampling~\cite{chen2022super}, require complex task-specific training pipelines. However, their fixed kernels cannot adapt to image content or feature semantics, often resulting in oversmoothing and limited ability to recover fine spatial structures.
 
Recent works overcome the limitations of classical methods by learning content-aware upsampling modules that enhance spatial details such as sharper boundaries, richer textures, or higher pixel-level fidelity.
LiFT~\cite{suri2024lift} proposes a lightweight ViT-compatible upsampler that fuses image and feature representations through a U-Net-style architecture. 
FeatUp~\cite{fu2024featup} provides both a fast JBU-based feedforward variant and a per-image implicit variant for sharper detail. 
LoftUp~\cite{huang2025loftup} models upsampling as a coordinate-to-feature mapping with cross-attention for global reasoning. 
JAFAR~\cite{couairon2025jafar} formulates upsampling as global attention between high-resolution image features and low-resolution patch features, enabling arbitrary-resolution outputs. 
AnyUp~\cite{wimmer2025anyup} introduces a feature-agnostic framework with channel-independent projection and window-based attention to generalize across backbone feature types.
However, they primarily focus on enhancing spatial details, while largely overlooking whether such enhancements are beneficial for 3D-aware inference. 
Therefore, systematically analyzing which factors influence overall reconstruction quality, geometry, and texture is crucial for guiding the design of future feature upsamplers.

\subsection{Measuring 3D Awareness}
Previous studies have examined the 3D awareness of VFMs~\cite{bommasani2021opportunities} through tasks such as geometric cue estimation~\cite{bae2024rethinking,fu2024geowizard,hu2025depthcrafter,ke2024repurposing,khirodkar2024sapiens,ye2024stablenormal}, 6D pose estimation~\cite{ornek2024foundpose}, spatial reasoning~\cite{chen2024spatialvlm}, and visual tracking~\cite{tumanyan2024dino}. 
While these works demonstrate the geometric capability of VFMs, many evaluations focus on coarse spatial reasoning, such as object consistency~\cite{bonnen2024evaluating}, spatial visual question answering~\cite{fu2024blink,zuo2025towards}, and visual perspective taking~\cite{linsley20243d}. 
More fine-grained evaluations, including depth or normal estimation and two-view correspondence~\cite{aydemir2024can,el2024probing}, typically rely on labeled 3D data, limiting scalability. 

Recent work explores probing 3D awareness without explicit 3D supervision. 
Feat2GS~\cite{chen2025feat2gs} follows this direction by regressing 3D Gaussian Splatting representations from VFM features and evaluating them via novel view synthesis (NVS), enabling dense and label-free probing of 3D awareness. 
Crucially, it enables disentangled evaluation of overall, geometry-only, and texture-only reconstruction fidelity, providing a controlled testbed for analyzing how different upsampling strategies affect distinct aspects of 3D reconstruction.

\section{Methods}
\subsection{Motivations and Workflow}
In 2D-to-3D pipelines, Vision Foundation Models (VFMs) extract patch-level features that must be upsampled into dense maps before pixel-level rendering and scene reconstruction. Feature upsampling therefore influences spatial continuity and local feature relationships, which directly affect the fidelity of reconstructed 3D scene.
Despite its importance, the role of feature upsampling in 3D perception remains largely unexplored. Most pipelines rely on simple interpolation such as bilinear upsampling, while recently proposed learnable upsamplers mainly focus on enhancing spatial details such as sharper boundaries or richer textures. 
However, it remains unclear whether such enhancements are beneficial for 3D-aware inference.
To address this gap, we investigate how different upsampling strategies reshape the spectral structure of 2D features and how these spectral changes influence 3D awareness, measured via NVS quality.

\begin{figure*}
\centering
\includegraphics[width=\linewidth]{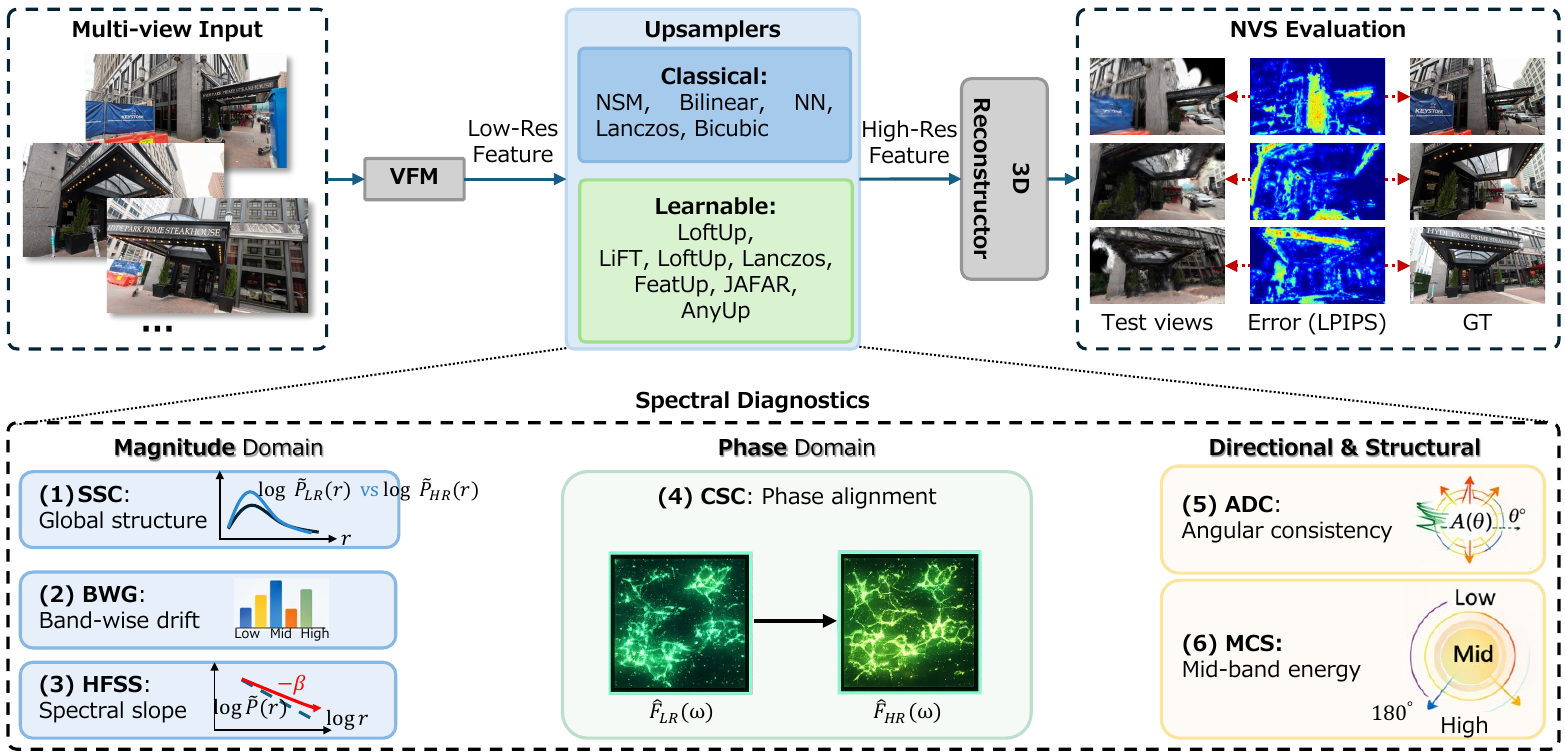}
\caption{Overview of the spectral probing pipeline. Feature upsamplers reshape spectral structure, which in turn
affects reconstruction quality. Our framework measures this
relationship through six spectral diagnostics and NVS evaluation. Specifically, multi-view images are resized to $H\times W$ and encoded into patch-grid features ($h\times w$). Different upsampling strategies produce dense maps at $H\times W$, which are used to regress 3D Gaussian parameters via differentiable rendering. Spectral changes are correlated with NVS quality (PSNR, SSIM, LPIPS) to probe 3D awareness.}
\label{fig:pipeline}
\end{figure*}

The workflow is shown in Fig.~\ref{fig:pipeline}. Using Feat2GS~\cite{chen2025feat2gs} as the baseline, each input image is resized to a fixed resolution $H\times W$ (($H=W=256$)). A VFM extracts the low-resolution feature map 
$F_{\text{LR}}\in\mathbb{R}^{h\times w\times C}$, which is then upsampled to 
$F_{\text{HR}}\in\mathbb{R}^{H\times W\times C}$. 
To isolate the effect of upsampling, all methods operate on the same $F_{\text{LR}}$ and differ only in the transformation from $h\times w$ to $H\times W$.
The resulting features are regressed into 3D Gaussian Splatting parameters and optimized via differentiable rendering. 
To analyze what information is preserved in the upsampled features, we adopt three probing modes that selectively read out Gaussian parameters: \textbf{Geometry (G):} predicts geometric parameters $(\mathbf{x}_i, \alpha_i, \Sigma_i)$; \textbf{Texture (T):} predicts appearance parameters $\mathbf{c}_i$; \textbf{All (A):} jointly predicts both geometry and appearance parameters. Here $\mathbf{x}_i$ denotes the 3D position of the $i$-th Gaussian, $\alpha_i$ its opacity, $\Sigma_i$ its covariance matrix, and $\mathbf{c}_i$ its appearance parameters.

\subsection{Spectral Diagnostics}
Feature upsampling not only changes spatial resolution but also modifies the spectral structure of feature representations. 
In the Fourier domain, it redistributes spectral energy, alters phase relationships, and may introduce orientation-dependent distortions. 
To systematically characterize these effects, we propose six complementary spectral diagnostics.
Global Spectral Structure Preservation (SSC) and Band-wise Spectral Drift (BWG) capture global and localized spectral energy redistribution. 
High-Frequency Spectral Slope Drift (HFSS) measures deviations from the natural decay of high-frequency components. 
Complex Spectral Coherence (CSC) evaluates phase-aligned structural preservation, while Angular Energy Consistency (ADC) detects orientation-dependent distortions. 
Mid-band Concentration Stability (MCS) quantifies the stability of mid-frequency structural components.
Together, these diagnostics provide complementary perspectives for understanding how spectral transformations induced by upsampling influence downstream 3D reconstruction quality.

\paragraph{Fourier Representation.}
For a feature map $F\in\{F_{\text{LR}},F_{\text{HR}}\}$, we compute the 2D Fourier transform
\[
\hat{F}(\boldsymbol{\omega})=\mathcal{F}(F),
\]
where $\boldsymbol{\omega}=(\omega_x,\omega_y)$ denotes the spatial frequency coordinate defined on the discrete frequency grid $\Omega$. 
The power spectrum is defined as
\begin{equation}
P(\boldsymbol{\omega})=|\hat{F}(\boldsymbol{\omega})|^2.
\end{equation}

In practice, spectra are computed independently for each feature channel and then aggregated across channels (e.g., by averaging) to obtain a single spectrum per feature map. 
Applying this transform to $F_{\text{LR}}$ and $F_{\text{HR}}$ produces spectra $P_{\text{LR}}(\boldsymbol{\omega})$ and $P_{\text{HR}}(\boldsymbol{\omega})$.
To summarize the frequency distribution independent of orientation, we compute a radial spectrum by averaging $P(\boldsymbol{\omega})$ within discrete radial bins defined by $r=\|\boldsymbol{\omega}\|$:
\begin{equation}
\tilde{P}(r)=
\mathbb{E}_{\|\boldsymbol{\omega}\|\in[r,r+\Delta r)} 
P(\boldsymbol{\omega}),
\end{equation}
where $\Delta r$ denotes the radial bin width. 
This operation yields radial spectra $\tilde{P}_{\text{LR}}(r)$ and $\tilde{P}_{\text{HR}}(r)$. All diagnostics below quantify changes between the low-resolution and upsampled spectra.

\noindent\textbf{Magnitude-Based Diagnostics: (1) SSC.}
SSC measures how well overall frequency allocation is preserved after upsampling, capturing global structural similarity while being insensitive to uniform spectral scaling.
We compute Pearson correlation between log radial spectra:
\begin{equation}
SSC
=
\rho\!\left(
\log \tilde{P}_{\text{LR}}(r),
\log \tilde{P}_{\text{HR}}(r)
\right),
\end{equation}
where $\rho(\cdot,\cdot)$ denotes the Pearson correlation coefficient.

\noindent\textbf{Magnitude-Based Diagnostics: (2) BWG.}
While SSC summarizes global similarity, it may miss localized redistribution across specific frequency ranges. 
To capture such frequency-selective effects, we introduce BWG, which measures how spectral energy shifts across frequency bands after upsampling.
We divide the radial frequency axis into $K$ disjoint bands $\{\mathcal{R}_k\}_{k=1}^{K}$, where $k$ indexes the band. 
The energy of band $k$ is computed as
\begin{equation}
E_k^{(\cdot)}
=
\sum_{r\in\mathcal{R}_k}
\tilde{P}_{(\cdot)}(r),
\end{equation}
where $(\cdot)\in\{\text{LR},\text{HR}\}$ indicates the spectrum source.
We then compare the normalized band energy distributions:
\begin{equation}
BWG
=
\sum_{k=1}^{K}
\left|
\frac{E_k^{\text{LR}}}{\sum_{j=1}^{K} E_j^{\text{LR}}}
-
\frac{E_k^{\text{HR}}}{\sum_{j=1}^{K} E_j^{\text{HR}}}
\right|,
\end{equation}
where $j$ indexes all frequency bands.

\noindent\textbf{Magnitude-Based Diagnostics: (3) HFSS.}
HFSS measures how strongly upsampling alters the natural spectral decay; large deviations suggest excessive sharpening or smoothing effects.
Natural visual signals approximately follow a power-law spectral decay:
\begin{equation}
\tilde{P}(r)\propto r^{-\beta},
\end{equation}
where $r$ denotes radial frequency and $\beta$ represents the spectral decay slope.
To estimate $\beta$, we perform linear regression in the log--log domain over a predefined high-frequency range:
\begin{equation}
\log \tilde{P}(r) = -\beta \log r + c,
\end{equation}
where $c$ is a constant. 
The slope $\beta$ is obtained via least-squares fitting using radial spectrum samples $\{(r_i,\tilde{P}(r_i))\}$ within the selected high-frequency region.
The slope drift introduced by upsampling is then defined as
\begin{equation}
HFSS
=
|\beta_{\text{HR}}-\beta_{\text{LR}}|.
\end{equation}

\noindent\textbf{Phase-Based Diagnostic: (4) CSC.}
Magnitude statistics characterize spectral energy distribution, whereas phase encodes spatial alignment and structural consistency. 
We therefore evaluate phase-consistent structural preservation using normalized complex spectral coherence between LR and HR spectra:
\begin{equation}
CSC
=
\left|
\frac{
\sum_{\boldsymbol{\omega}\in\Omega}
\hat{F}_{\text{LR}}(\boldsymbol{\omega})
\overline{\hat{F}_{\text{HR}}(\boldsymbol{\omega})}
}{
\sqrt{
\sum_{\boldsymbol{\omega}\in\Omega}
|\hat{F}_{\text{LR}}(\boldsymbol{\omega})|^2
\sum_{\boldsymbol{\omega}\in\Omega}
|\hat{F}_{\text{HR}}(\boldsymbol{\omega})|^2
}
}
\right|.
\end{equation}

\noindent\textbf{Directional and Structural Diagnostics: (5) ADC.}
Radial statistics ignore orientation-dependent distortions. 
To analyze directional spectral structure, ADC quantifies how well the orientation distribution of spectral energy is preserved after upsampling.
We partition the angular domain into bins $\{\mathcal{B}_m\}$, where $m$ indexes the angular bin. 
The angular energy in bin $m$ is computed as
\begin{equation}
A_m^{(\cdot)}
=
\sum_{\boldsymbol{\omega}\in\mathcal{B}_m}
P_{(\cdot)}(\boldsymbol{\omega}),
\end{equation}
where $(\cdot)\in\{\text{LR},\text{HR}\}$.
The similarity between angular energy distributions is measured by
\begin{equation}
ADC
=
\rho(A^{\text{LR}},A^{\text{HR}}),
\end{equation}
where 
\[
A^{\text{LR}}=\{A_m^{\text{LR}}\}_m, 
\qquad
A^{\text{HR}}=\{A_m^{\text{HR}}\}_m .
\]

\noindent\textbf{Directional and Structural Diagnostics: (6) MCS.}
Mid-frequency components often encode structural edges and geometric contours. 
Let $\mathcal{R}_{mid}$ denote the mid-frequency region. 
The mid-band concentration is defined as
\begin{equation}
MCS
=
\frac{E_{mid}}{E_{total}},
\end{equation}
where
\[
E_{mid}
=
\sum_{r\in\mathcal{R}_{mid}}\tilde{P}(r),
\qquad
E_{total}
=
\sum_{r}\tilde{P}(r).
\]

The deviation introduced by upsampling is defined as
\begin{equation}
\Delta MCS
=
|MCS_{\text{HR}}-MCS_{\text{LR}}|.
\end{equation}

\section{Experimental Results}
\subsection{Experimental Settings}
\noindent\textbf{Implementation Details.}
We evaluate two widely used vision backbones, DINO and CLIP. 
Both adopt a patch size of $16 \times 16$ and provide publicly available pretrained checkpoints for most learnable upsamplers. 
The input images are resized to $256 \times 256$, which is evenly divisible by the patch size and ensures consistent feature map alignment across backbones.

We evaluate both classical interpolation and learnable upsampling methods. 
The classical baselines include bilinear, nearest-neighbor, bicubic, and Lanczos interpolation. 
For learnable upsamplers, we adopt publicly available pretrained checkpoints whenever possible, including JAFAR (DINO, CLIP)~\cite{couairon2025jafar}, AnyUp (DINO, CLIP)~\cite{wimmer2025anyup}, FeatUp (JBU) (CLIP)~\cite{fu2024featup}, LoftUp (CLIP)~\cite{huang2025loftup}, and LiFT (DINO)~\cite{suri2024lift}.
All methods are evaluated within the same framework with identical data processing, feature extraction, and reconstruction settings.

We introduce Non-cropping Spatial Matching (NSM), a simple baseline that isolates the effect of interpolation. 
Instead of resampling, NSM applies right- and bottom-side zero-padding to match the target resolution $(H, W)$. 
Given a low-resolution feature map $\mathbf{F}_{\mathrm{LR}} \in \mathbb{R}^{h \times w \times C}$, NSM produces
$\mathbf{F}_{\mathrm{NS}} \in \mathbb{R}^{H \times W \times C}$ as
\begin{equation}
\mathbf{F}_{\mathrm{NS}} =
\begin{cases}
\mathbf{F}_{\mathrm{LR}}, & \text{if } (h, w) = (H, W), \\[4pt]
\texttt{Pad}\!\left(\mathbf{F}_{\mathrm{LR}};\, (0,\, W-w,\, 0,\, H-h)\right), & \text{otherwise}.
\end{cases}
\label{eq:nsm}
\end{equation}
Here $\texttt{Pad}(\cdot)$ denotes zero-padding along the right and bottom edges to expand the feature map to $(H, W)$.

\noindent\textbf{Datasets and Metrics.}
We conduct experiments on six diverse multi-view datasets (Table~\ref{tab:datasets}). 
For each dataset, 2–7 sparse training views are sampled, while test viewpoints are placed farther away to enforce extrapolation. 
In total, we evaluate 30 scenes. 
For each scene, spectral diagnostics are computed on all training views and averaged to obtain scene-level statistics. 
NVS quality is evaluated per scene, and cross-scene correlations between spectral diagnostics and NVS metrics are analyzed to study how frequency-domain characteristics relate to 3D awareness.

We adopt three complementary NVS metrics: PSNR, SSIM, and LPIPS, capturing pixel fidelity, structural similarity, and perceptual similarity, respectively, providing a multi-scale assessment of geometry and appearance consistency across views.

\begin{table}[t]
\caption{Datasets for Evaluation.}
\label{tab:datasets}
\centering
\small
\setlength{\tabcolsep}{7pt}
\renewcommand{\arraystretch}{1.2}
\resizebox{.8\textwidth}{!}{
\begin{tabular}{@{}lcccc@{}}
\toprule
\rowcolor{gray!10} 
\textbf{Dataset} & \textbf{Scene Type} & \textbf{Complexity} & \textbf{View Range} & \textbf{Views} \\ 
\midrule
LLFF~\cite{mildenhall2019local} & Indoor & Simple & Small & 2 \\
DL3DV~\cite{ling2024dl3dv} & Indoor / Outdoor & Moderate & Medium & 5--6 \\
Casual~\cite{chen2025feat2gs} & Daily Scenario & Moderate & Medium & 4--7 \\
MipNeRF360~\cite{barron2022mip} & Unbounded & Moderate & 360 & 6 \\
MVImgNet~\cite{yu2023mvimgnet} & Outdoor Object & Moderate & 180--360 & 2--4 \\
T\&T~\cite{knapitsch2017tanks} & Indoor / Outdoor & High & Large & 6 \\
\bottomrule
\end{tabular}}
\end{table}

\setlength{\tabcolsep}{4pt}
\begin{table}[htbp]
\caption{Cross-dataset average NVS performance using DUSt3R~\cite{wang2024dust3r} as the 3D reconstructor. 
Results are reported under three probing modes: All (geometry+texture), Geometry-only, and Texture-only. 
Metrics are averaged over 30 scenes from six datasets. 
Best results are highlighted in bold.}
\label{tab:dust3r-all}
\centering
\resizebox{\textwidth}{!}{
\begin{tabular}{l|cccccccccc}
\toprule
\multicolumn{1}{c|}{} & \multicolumn{1}{c|}{} & \multicolumn{3}{c|}{All}  & \multicolumn{3}{c|}{Geometry}   & \multicolumn{3}{c|}{Texture}  \\
\midrule
Feature& Upsamplers & PSNR $\uparrow$ & SSIM $\uparrow$ & LPIPS $\downarrow$ & PSNR $\uparrow$ & SSIM $\uparrow$ & LPIPS $\downarrow$ & PSNR $\uparrow$ & SSIM $\uparrow$ & LPIPS $\downarrow$\\
\midrule
\multirow{8}{*}{DINO} & NSM  &  23.62 & 0.8383 &    0.1628  &  18.13 & 0.7811 &    0.2212   &  22.57 & 0.8204 &    0.1415  \\  \cline{2-11}
& Bilinear    &  \underline{24.08} & 0.8400 &    0.1601  &  \underline{24.32} & \textbf{0.8483} &    0.1368  &  22.58 & 0.8193 &    0.1413   \\  \cline{2-11}
& NN     &  23.94 & 0.8396 &    0.1606   &  22.54 & 0.8214 &    0.1790&  22.71 & 0.8211 &    0.1392    \\  \cline{2-11}
& Lanczos    &  24.07 & \underline{0.8408} &    0.1582   & 24.29 & 0.8475 &    \textbf{0.1342} &  22.71 & \underline{0.8215} &    0.1392   \\  \cline{2-11}
& Bicubic    &  \textbf{24.15} & \textbf{0.8416} &    \underline{0.1577}  &  \textbf{24.36} & \underline{0.8477} &    \underline{0.1343}  &  \textbf{22.83} & \textbf{0.8237} &    \textbf{0.1370}  \\  \cline{2-11}
& LiFT~\cite{suri2024lift}    &  24.04 & 0.8389 &    \textbf{0.1571} &  24.16 & 0.8400 &    0.1416  &  22.72 & 0.8201 &    0.1398   \\  \cline{2-11}
& JAFAR~\cite{couairon2025jafar}  &  23.61 & 0.8375 &    0.1630 &  24.15 & 0.8416 &    0.1420  &  21.52 & 0.8084 &    0.1568     \\   \cline{2-11}
& AnyUp~\cite{wimmer2025anyup}     &  24.04 & 0.8389 &    0.1625  & 24.13 & 0.8430 &    \textbf{0.1401} &  \underline{22.74} & 0.8209 &    \underline{0.1389}  \\  \midrule

\multirow{9}{*}{CLIP} & NSM &  \textbf{24.33} & \textbf{0.8479} &    \textbf{0.1516}  &  19.42 & 0.7988 &    0.2084  &  22.68 & 0.8210 &    0.1397     \\  \cline{2-11}
& Bilinear   &  23.91 & 0.8403 &    0.1594  &  \underline{24.21} & 0.8454 &    0.1388 &  22.56 & 0.8200 &    0.1409   \\  \cline{2-11}
& NN  &  24.05 & 0.8397 &    0.1592  &  22.56 & 0.8209 &    0.1808 &  22.62 & 0.8208 &    0.1405  \\  \cline{2-11}
& Lanczos   &  23.90 & 0.8393 &    0.1584  &  24.14 & 0.8431 &    0.1391  &  \underline{22.70} & \underline{0.8211} &    0.1397     \\  \cline{2-11}
& Bicubic  &  \underline{24.32} & \underline{0.8468} &   \underline{0.1522} &  24.13 & 0.8433 &    0.1393  &  22.67 & 0.8209 &    0.1399     \\  \cline{2-11}
& LoftUp~\cite{huang2025loftup}  &  24.05 & 0.8416 &    0.1601   &  24.19 & 0.8450 &    0.1390  &  \textbf{22.75} & \textbf{0.8213} &   \underline{0.1389}  \\  \cline{2-11}
& FeatUp~\cite{fu2024featup}  &  23.95 & 0.8391 &    0.1592  &  \textbf{24.53} & \textbf{0.8514} &    \textbf{0.1301}  &  22.62 & 0.8207 &    0.1402     \\  \cline{2-11}
& JAFAR~\cite{couairon2025jafar}    &  23.98 & 0.8401 &    0.1589   &  23.93 & 0.8402 &    0.1432 &  21.53 & 0.8051 &    0.1570   \\  \cline{2-11}
& AnyUp~\cite{wimmer2025anyup}  &  24.12 & 0.8416 &    0.1628  &  24.19 & \underline{0.8456} &    \underline{0.1386}  &  \underline{22.70} & 0.8205 &    \textbf{0.1388}   \\
\bottomrule
\end{tabular}}
\vspace{0.02em}
\caption{Cross-dataset average NVS performance using MASt3R~\cite{leroy2024grounding} as the 3D reconstructor. 
Results are reported under three probing modes: All (geometry+texture), Geometry-only, and Texture-only. Metrics are averaged over 30 scenes from six datasets. 
Best results are highlighted in bold.}
\label{tab:mast3r-all}
\centering
\resizebox{\textwidth}{!}{
\begin{tabular}{l|cccccccccc}
\toprule
\multicolumn{1}{c|}{} & \multicolumn{1}{c|}{} & \multicolumn{3}{c|}{All}  & \multicolumn{3}{c|}{Geometry}   & \multicolumn{3}{c|}{Texture}  \\
\midrule
Feature& Upsamplers & PSNR $\uparrow$ & SSIM $\uparrow$ & LPIPS $\downarrow$ & PSNR $\uparrow$ & SSIM $\uparrow$ & LPIPS $\downarrow$ & PSNR $\uparrow$ & SSIM $\uparrow$ & LPIPS $\downarrow$\\
\midrule 
\multirow{8}{*}{DINO} & NSM     &  17.93 & \textbf{0.7818} &    \textbf{0.2212}    &  17.87 & \textbf{0.7751} &    0.2256  &  16.25 & 0.7120 &    \textbf{0.2315} \\ \cline{2-11}
& Bilinear    &  \textbf{18.26} & 0.7628 &    0.2273 &  16.68 & 0.7272 &    0.2419 &  15.55 & 0.6928 &    0.2571   \\  \cline{2-11}
& NN   & 17.15 & 0.7591 &    0.2384 & 17.87 & \underline{0.7688} &    0.2225 &  16.25 & 0.7116 &   \underline{0.2317}     \\  \cline{2-11}
& Lanczos     &  17.78 & \underline{0.7753} &   0.2261  & \underline{17.95} & 0.7496 &   \textbf{0.2196}  &  16.50 & 0.7151 &  0.2334  \\  \cline{2-11}
& Bicubic  &  17.89 & 0.7580 &    0.2269  &  17.69 & 0.7468 &    \underline{0.2207} &  \underline{16.63} & \underline{0.7159} &    0.2322     \\  \cline{2-11}
& LiFT~\cite{suri2024lift}   &  17.67 & 0.7628 &    0.2330  &  17.62 & 0.7440 &    0.2216 &  15.82 & 0.7008 &    0.2433     \\  \cline{2-11}
& JAFAR~\cite{couairon2025jafar}   &  17.77 & 0.7573 &    0.2270  &  \textbf{17.97} & 0.7475 &    0.2219  &  \textbf{16.66} & 0.7118 &    0.2388    \\   \cline{2-11}
& AnyUp~\cite{wimmer2025anyup}    &  \underline{17.95} & 0.7739 &   \underline{0.2250} &  17.45 & 0.7384 &    0.2284   &  16.62 & \textbf{0.7163} &    0.2318 \\  \midrule

\multirow{9}{*}{CLIP} & NSM  &  18.02 & \underline{0.7774} &    \underline{0.2218}   &  \textbf{18.25} & \textbf{0.7844} &    \textbf{0.2148}  &  15.98 & 0.7015 &    0.2422 \\  \cline{2-11}
& Bilinear   &  17.90 & 0.7498 &    0.2378 &  17.38 & 0.7381 &    0.2256  &  16.02 & 0.7022 &    0.2429   \\  \cline{2-11}
& NN    &  18.09 & \textbf{0.7796} &    \textbf{0.2206}  & \underline{18.01} & \underline{0.7641} &   0.2301   &  16.37 & 0.7060 &    0.2428    \\  \cline{2-11}
& Lanczos    &  \underline{18.15} & 0.7713 &    0.2254 &  17.42 & 0.7436 &    0.2264   &  16.50 & 0.7151 &    \underline{0.2329}    \\  \cline{2-11}
& Bicubic   &  17.28 & 0.7453 &    0.2437   &  16.95 & 0.7298 &    0.2346  &  15.86 & 0.6980 &    0.2464   \\  \cline{2-11}
& LoftUp~\cite{huang2025loftup}  &  17.61 & 0.7519 &    0.2317  &  17.24 & 0.7366 &    0.2276 &  16.04 & 0.7042 &    0.2417  \\  \cline{2-11}
& FeatUp~\cite{fu2024featup}    &  \textbf{18.16} & 0.7580 &    0.2220  &  17.96 & 0.7453 &   \underline{0.2192}  &  \textbf{16.79} & \textbf{0.7168} &    \textbf{0.2320}  \\  \cline{2-11}
& JAFAR~\cite{couairon2025jafar}  &  17.13 & 0.7438 &    0.2374 &  17.64 & 0.7348 &    0.2321 &  16.38 & 0.7012 &    0.2461      \\  \cline{2-11}
& AnyUp~\cite{wimmer2025anyup}    &  18.09 & 0.7706 &    0.2267 & 17.73 & 0.7460 &    0.2202  &  \underline{16.52} & \underline{0.7153} &    0.2331   \\
\bottomrule
\end{tabular}}
\end{table}

%%% BEGIN AUTOGENERATED %%%
\setlength{\tabcolsep}{8pt}
\begin{table*}[t!]
\caption{Per-dataset NVS performance using DUSt3R~\cite{wang2024dust3r} with the CLIP backbone on six multi-view datasets under three probing modes: Geometry-only, Texture-only, and All (geometry+texture). 
Best results are highlighted in bold. \textbf{More results are provided in the Supplementary Material.}}
\label{tab:dust3r-clip}
\setlength{\tabcolsep}{4pt}
\centering
\resizebox{\textwidth}{!}{
\begin{tabular}{l|>{\raggedleft\arraybackslash}p{0.9cm}>{\raggedleft\arraybackslash}p{0.9cm}>{\raggedleft\arraybackslash}p{0.9cm}|>{\raggedleft\arraybackslash}p{0.9cm}>{\raggedleft\arraybackslash}p{0.9cm}>{\raggedleft\arraybackslash}p{0.9cm}|>{\raggedleft\arraybackslash}p{0.9cm}>{\raggedleft\arraybackslash}p{0.9cm}>{\raggedleft\arraybackslash}p{0.9cm}|>{\raggedleft\arraybackslash}p{0.9cm}>{\raggedleft\arraybackslash}p{0.9cm}>{\raggedleft\arraybackslash}p{0.9cm}|>{\raggedleft\arraybackslash}p{0.9cm}>{\raggedleft\arraybackslash}p{0.9cm}>{\raggedleft\arraybackslash}p{0.9cm}|>{\raggedleft\arraybackslash}p{0.9cm}>{\raggedleft\arraybackslash}p{0.9cm}>{\raggedleft\arraybackslash}p{0.9cm}}
\toprule
\multicolumn{1}{c|}{} & \multicolumn{9}{c|}{LLFF} & \multicolumn{9}{c}{DL3DV} \\
\midrule
\multicolumn{1}{c|}{} & \multicolumn{3}{c|}{\textbf{G}eometry} & \multicolumn{3}{c|}{\textbf{T}exture} & \multicolumn{3}{c|}{\textbf{A}ll} & \multicolumn{3}{c|}{\textbf{G}eometry} & \multicolumn{3}{c|}{\textbf{T}exture} & \multicolumn{3}{c}{\textbf{A}ll} \\
\midrule
Upsamplers & \fontsize{8.5pt}{9pt}\selectfont{PSNR$\uparrow$} & \fontsize{8.5pt}{9pt}\selectfont{SSIM$\uparrow$} & \fontsize{8.5pt}{9pt}\selectfont{LPIPS$\downarrow$} & \fontsize{8.5pt}{9pt}\selectfont{PSNR$\uparrow$} & \fontsize{8.5pt}{9pt}\selectfont{SSIM$\uparrow$} & \fontsize{8.5pt}{9pt}\selectfont{LPIPS$\downarrow$} & \fontsize{8.5pt}{9pt}\selectfont{PSNR$\uparrow$} & \fontsize{8.5pt}{9pt}\selectfont{SSIM$\uparrow$} & \fontsize{8.5pt}{9pt}\selectfont{LPIPS$\downarrow$} & \fontsize{8.5pt}{9pt}\selectfont{PSNR$\uparrow$} & \fontsize{8.5pt}{9pt}\selectfont{SSIM$\uparrow$} & \fontsize{8.5pt}{9pt}\selectfont{LPIPS$\downarrow$} & \fontsize{8.5pt}{9pt}\selectfont{PSNR$\uparrow$} & \fontsize{8.5pt}{9pt}\selectfont{SSIM$\uparrow$} & \fontsize{8.5pt}{9pt}\selectfont{LPIPS$\downarrow$} & \fontsize{8.5pt}{9pt}\selectfont{PSNR$\uparrow$} & \fontsize{8.5pt}{9pt}\selectfont{SSIM$\uparrow$} & \fontsize{8.5pt}{9pt}\selectfont{LPIPS$\downarrow$} \\
\midrule
NSM   &19.21 &.8350 &   .1593 &24.11 &.8904 &   .0779 &\underline{25.13} &\textbf{.9028} &   \textbf{.0858} &18.26 &.8207 &   .1967 &20.62 &.8334 &   .1555 &\underline{22.09} &.8616 &   .1596 \\  \midrule
Bilinear    &25.05 &.9021 &   \underline{.0745} &23.96 &.8882 &   .0803 &24.99 &.8899 &   .0919 &22.27 &.8614 &   .1514 &\underline{20.86} &\underline{.8373} &   \underline{.1528} &22.00 &.8618 &   \underline{.1590} \\
 \midrule
NN    &23.04 &.8529 &   .1256 &24.00 &.8882 &   .0802 &25.02 &.8887 &   .0923 &20.74 &.8474 &   .1712 &\textbf{21.03} &\textbf{.8388} &   \textbf{.1520} &22.03 &.8607 &   .1598 \\
  \midrule
Lanczos     &24.91 &.8996 &   .0750 &24.14 &.8906 &   .0779 &24.96 &.8880 &   .0915&\underline{22.36} &.8609 &   \underline{.1502} &20.63 &.8329 &   .1560 &21.93 &.8604 &   .1598 \\  \midrule
Bicubic   &24.93 &.9005 &   .0746 &24.11 &.8901 &   .0781 &\textbf{25.22} &\underline{.9015} &   \underline{.0870} &22.20 &.8607 &   .1515 &20.59 &.8334 &   .1553 &21.77 &.8581 &   .1632 \\ \midrule
LoftUp~\cite{huang2025loftup}  &24.97 &.9027 &   .0761 &24.37 &.8915 &   .0769 &24.88 &.8931 &   .0941 &22.33 &\underline{.8616} &   .1518 &20.79 &.8354 &   .1545 &21.85 &.8603 &   .1607 \\  \midrule
FeatUp~\cite{fu2024featup}  &\underline{25.10} &\textbf{.9080} &   \textbf{.0679} &24.23 &.8913 &   .0767 &25.04 &.8885 &   .0917 &\textbf{22.61} &\textbf{.8666} &   \textbf{.1439} &20.40 &.8314 &   .1581 &21.70 &.8570 &   .1617 \\ \midrule
JAFAR~\cite{couairon2025jafar}   &25.07 &.9041 &   .0757 &\textbf{24.39} &\textbf{.8929} &   \textbf{.0752} &24.96 &.8884 &   .0928 &21.98 &.8555 &   .1570 &18.54 &.8061 &   .1843 &\textbf{22.19} &\underline{.8637} &   \textbf{.1573} \\ \midrule
AnyUp~\cite{wimmer2025anyup}    &\textbf{25.16} &\underline{.9054} &   .0757 &\underline{24.38} &\underline{.8922} &   \underline{.0756} &25.08 &.8855 &   .0972 &22.30 &.8610 &   .1528 &20.66 &.8328 &   .1555 &21.94 &\textbf{.8639} &   .1605 \\
\midrule
\multicolumn{1}{c|}{} & \multicolumn{9}{c|}{Casual} & \multicolumn{9}{c}{MipNeRF360} \\
\midrule
\multicolumn{1}{c|}{} & \multicolumn{3}{c|}{\textbf{G}eometry} & \multicolumn{3}{c|}{\textbf{T}exture} & \multicolumn{3}{c|}{\textbf{A}ll} & \multicolumn{3}{c|}{\textbf{G}eometry} & \multicolumn{3}{c|}{\textbf{T}exture} & \multicolumn{3}{c}{\textbf{A}ll} \\
\midrule
Upsamplers & \fontsize{8.5pt}{9pt}\selectfont{PSNR$\uparrow$} & \fontsize{8.5pt}{9pt}\selectfont{SSIM$\uparrow$} & \fontsize{8.5pt}{9pt}\selectfont{LPIPS$\downarrow$} & \fontsize{8.5pt}{9pt}\selectfont{PSNR$\uparrow$} & \fontsize{8.5pt}{9pt}\selectfont{SSIM$\uparrow$} & \fontsize{8.5pt}{9pt}\selectfont{LPIPS$\downarrow$} & \fontsize{8.5pt}{9pt}\selectfont{PSNR$\uparrow$} & \fontsize{8.5pt}{9pt}\selectfont{SSIM$\uparrow$} & \fontsize{8.5pt}{9pt}\selectfont{LPIPS$\downarrow$} & \fontsize{8.5pt}{9pt}\selectfont{PSNR$\uparrow$} & \fontsize{8.5pt}{9pt}\selectfont{SSIM$\uparrow$} & \fontsize{8.5pt}{9pt}\selectfont{LPIPS$\downarrow$} & \fontsize{8.5pt}{9pt}\selectfont{PSNR$\uparrow$} & \fontsize{8.5pt}{9pt}\selectfont{SSIM$\uparrow$} & \fontsize{8.5pt}{9pt}\selectfont{LPIPS$\downarrow$} & \fontsize{8.5pt}{9pt}\selectfont{PSNR$\uparrow$} & \fontsize{8.5pt}{9pt}\selectfont{SSIM$\uparrow$} & \fontsize{8.5pt}{9pt}\selectfont{LPIPS$\downarrow$} \\
\midrule
NSM   &19.94 &.8041 &   .2258 &22.80 &.7974 &   .1736 &24.57 &\textbf{.8467} &   \underline{.1825} &19.42 &.7028 &   .2843 &23.43 &.7342 &   \textbf{.1709} &\textbf{24.95} &\textbf{.7585} &   \underline{.2090} \\  \midrule
Bilinear    &24.40 &.8424 &   .1708 &22.67 &.7961 &   .1759 &23.71 &.8427 &   .1881 &\underline{24.81} &\underline{.7576} &   .1761 &23.33 &.7341 &   .1716 &24.57 &.7490 &   .2152 \\
 \midrule
NN    &23.07 &.8260 &   .2029 &22.70 &.7963 &   .1758 &24.50 &.8444 &   .1866 &23.34 &.7327 &   .2481 &23.28 &.7332 &   .1723 &24.56 &.7477 &   .2145 \\
 \midrule
Lanczos    &24.60 &.8435 &   .1708 &22.81 &\underline{.7980} &   .1732 &23.96 &.8431 &   .1876 &24.70 &.7531 &   .1779 &\textbf{23.46} &\underline{.7344} &   \textbf{.1709} &24.50 &.7465 &   .2133 \\  \midrule
Bicubic    &24.63 &.8433 &   .1696 &22.82 &.7973 &   .1743 &\underline{24.65} &\underline{.8460} &   \textbf{.1816} &24.73 &.7531 &   .1782 &\underline{23.45} &\textbf{.7348} &   \underline{.1714} &\underline{24.85} &\underline{.7577} &   \textbf{.2083} \\ \midrule
LoftUp~\cite{huang2025loftup}    &\underline{24.75} &\underline{.8464} &   \underline{.1676} &22.87 &\textbf{.7981} &   \underline{.1723} &\textbf{24.66} &.8449 &   .1890 &24.67 &.7534 &   .1768 &23.29 &.7305 &   .1719 &24.59 &.7497 &   .2187 \\  \midrule
FeatUp~\cite{fu2024featup}   &\textbf{25.47} &\textbf{.8482} &   \textbf{.1580} &22.72 &.7969 &   .1754 &24.58 &.8450 &   .1861 &\textbf{24.90} &\textbf{.7660} &   \textbf{.1643} &23.25 &.7311 &   .1728 &24.51 &.7477 &   .2137 \\ \midrule
JAFAR~\cite{couairon2025jafar}    &24.26 &.8453 &   .1715 &\textbf{23.04} &.7977 &   .1724 &24.33 &.8446 &   .1881 &24.48 &.7490 &   .1817 &21.33 &.7076 &   .2046 &24.49 &.7483 &   .2135 \\ \midrule
AnyUp~\cite{wimmer2025anyup}    &24.31 &.8461 &   .1681 &\underline{22.92} &.7974 &   \textbf{.1722} &24.09 &.8414 &   .1924 &24.78 &.7565 &   \underline{.1753} &23.26 &.7297 &   .1720 &24.62 &.7524 &   .2241 \\
\midrule
\multicolumn{1}{c|}{} & \multicolumn{9}{c|}{MVImgNet} & \multicolumn{9}{c}{T\&T} \\
\midrule
\multicolumn{1}{c|}{} & \multicolumn{3}{c|}{\textbf{G}eometry} & \multicolumn{3}{c|}{\textbf{T}exture} & \multicolumn{3}{c|}{\textbf{A}ll} & \multicolumn{3}{c|}{\textbf{G}eometry} & \multicolumn{3}{c|}{\textbf{T}exture} & \multicolumn{3}{c}{\textbf{A}ll} \\
\midrule
Upsamplers & \fontsize{8.5pt}{9pt}\selectfont{PSNR$\uparrow$} & \fontsize{8.5pt}{9pt}\selectfont{SSIM$\uparrow$} & \fontsize{8.5pt}{9pt}\selectfont{LPIPS$\downarrow$} & \fontsize{8.5pt}{9pt}\selectfont{PSNR$\uparrow$} & \fontsize{8.5pt}{9pt}\selectfont{SSIM$\uparrow$} & \fontsize{8.5pt}{9pt}\selectfont{LPIPS$\downarrow$} & \fontsize{8.5pt}{9pt}\selectfont{PSNR$\uparrow$} & \fontsize{8.5pt}{9pt}\selectfont{SSIM$\uparrow$} & \fontsize{8.5pt}{9pt}\selectfont{LPIPS$\downarrow$} & \fontsize{8.5pt}{9pt}\selectfont{PSNR$\uparrow$} & \fontsize{8.5pt}{9pt}\selectfont{SSIM$\uparrow$} & \fontsize{8.5pt}{9pt}\selectfont{LPIPS$\downarrow$} & \fontsize{8.5pt}{9pt}\selectfont{PSNR$\uparrow$} & \fontsize{8.5pt}{9pt}\selectfont{SSIM$\uparrow$} & \fontsize{8.5pt}{9pt}\selectfont{LPIPS$\downarrow$} & \fontsize{8.5pt}{9pt}\selectfont{PSNR$\uparrow$} & \fontsize{8.5pt}{9pt}\selectfont{SSIM$\uparrow$} & \fontsize{8.5pt}{9pt}\selectfont{LPIPS$\downarrow$} \\
\midrule
NSM    &18.83 &.7728 &   .2130 &21.28 &.7947 &   .1469 &\textbf{24.25} &\textbf{.8257} &   \textbf{.1416} &20.85 &.8574 &   .1712 &23.82 &.8760 &   .1132 &\underline{25.00} & \underline{.8924} &    \underline{.1309} \\ \midrule
Bilinear    &\underline{24.09} &\underline{.8244} &   \underline{.1313} &21.06 &.7920 &   .1494 &23.82 &.8155 &   .1635 &\underline{24.65} &.8841 &   .1284 &23.51 &.8724 &   .1154 &24.38 &.8830 &   .1387 \\
 \midrule
NN    &21.93 &.7910 &   .1944 &21.04 &.7920 &   .1493 &23.81 &.8145 &   .1637 &23.25 &.8756 &   .1425 &23.67 &.8761 &   .1132 &24.36 &.8822 &   .1386 \\
 \midrule
Lanczos   &23.89 &.8208 &   .1321 &21.27 &.7945 &   .1473 &23.70 &.8147 &   .1610 &24.36 &.8805 &   .1288 &\textbf{23.87} &.8762 &   .1132 &24.35 &.8831 &   .1373 \\ \midrule
Bicubic   &24.02 &.8224 &   .1317 &21.25 &.7938 &   .1471 &\textbf{24.25} &\underline{.8251} &   \underline{.1423} &24.30 &.8798 &   .1303 &23.80 &.8762 &   .1132 &\textbf{25.16}&\textbf{.8928} &   \textbf{.1308} \\ \midrule
LoftUp~\cite{huang2025loftup}    &23.95 &.8218 &   .1332 &\textbf{21.35} &\underline{.7957} &   \underline{.1455} &23.86 &.8157 &   .1607 &24.48 &.8842 &   .1282 &\underline{23.84} &\underline{.8765} &   \underline{.1123} &24.46 &.8858 &   .1377 \\  \midrule
FeatUp~\cite{fu2024featup}    &\textbf{24.24} &\textbf{.8343} &   \textbf{.1221} &\underline{21.31} &\textbf{.7959} &   \textbf{.1450} &23.67 &.8145 &   .1627 &\textbf{24.86} &\textbf{.8854} &   \textbf{.1248} &23.80 &\textbf{.8777} &   .1135 &24.19 &.8819 &   .1395 \\ \midrule
JAFAR~\cite{couairon2025jafar}    &23.69 &.8101 &   .1397 &20.03 &.7676 &   .1719 &23.64 &.8139 &   .1625 &24.11 &.8773 &   .1336 &21.87 &.8589 &   .1334 &24.29 &.8816 &   .1393 \\ \midrule
AnyUp~\cite{wimmer2025anyup}  &24.01 &.8202 &   .1339 &21.25 &.7945 &   .1458 &\underline{24.03} &.8151 &   .1691 &24.58 &\underline{.8846} &   \underline{.1260} &23.69 &.8764 &   \textbf{.1119} &24.94 &.8913 &   .1336 \\
\bottomrule
\end{tabular}
}
\end{table*}

\begin{figure*}
    \centering
    \includegraphics[width=\linewidth]{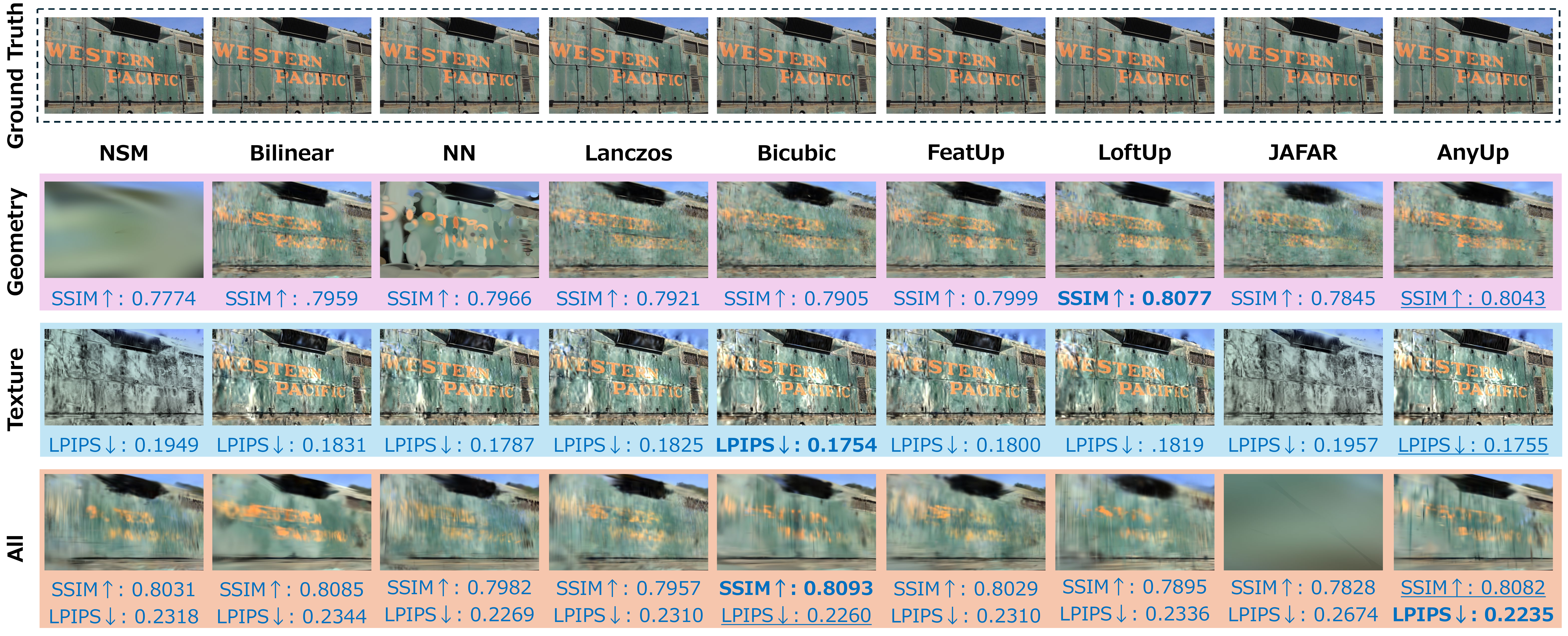}
    \caption{NVS visualizations using CLIP+DUSt3R~\cite{wang2024dust3r}. Classical interpolation methods often achieve performance comparable to learned upsamplers. Best and second-best results are shown in bold and \underline{underlined}, respectively.}
    \label{fig:NVS}
\end{figure*}

\begin{figure}[htbp]
\centering
\captionsetup[subfigure]{font=small, labelfont=bf}

% ===================== (a) =====================
\begin{subfigure}[t]{\linewidth}
    \centering
    \includegraphics[width=0.49\linewidth]{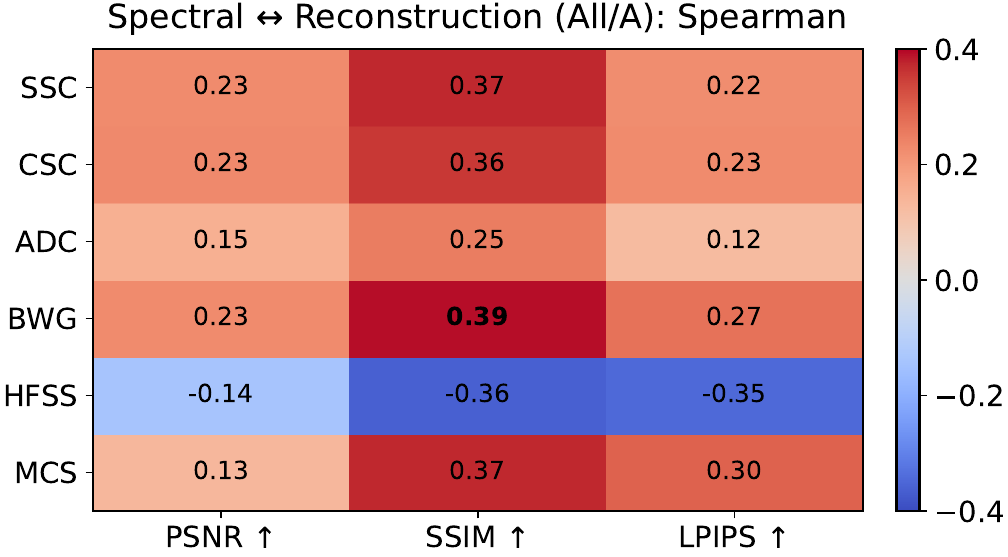}
    \hfill
   \includegraphics[width=0.49\linewidth]{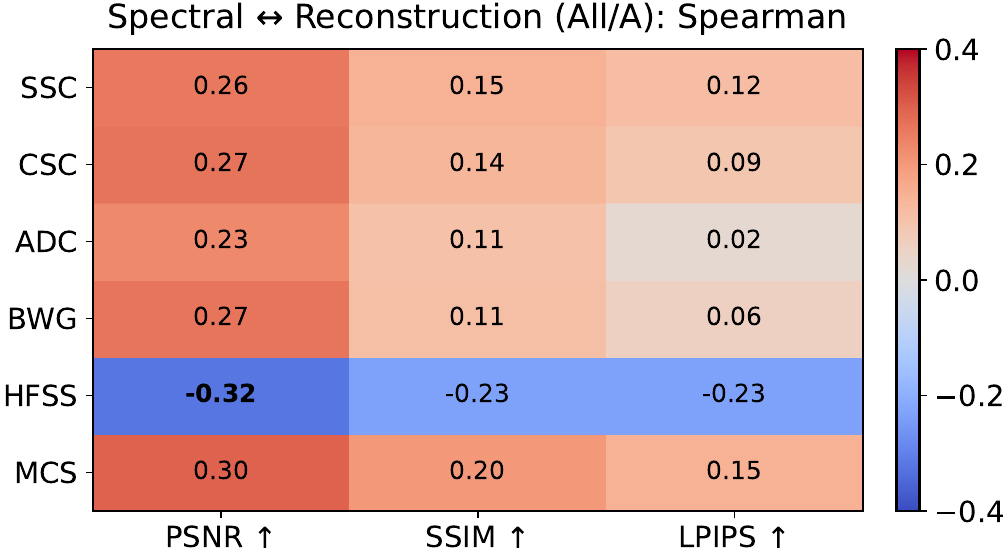}
    \vspace{-0.3em}
    \caption{Left: CLIP+Lanczos+DUSt3R~\cite{wang2024dust3r}; Right: DINO+Lanczos+DUSt3R~\cite{wang2024dust3r}.}
\end{subfigure}
\vspace{0.6em}
% ===================== (b) =====================
\begin{subfigure}[t]{\linewidth}
    \centering
    \includegraphics[width=0.49\linewidth]{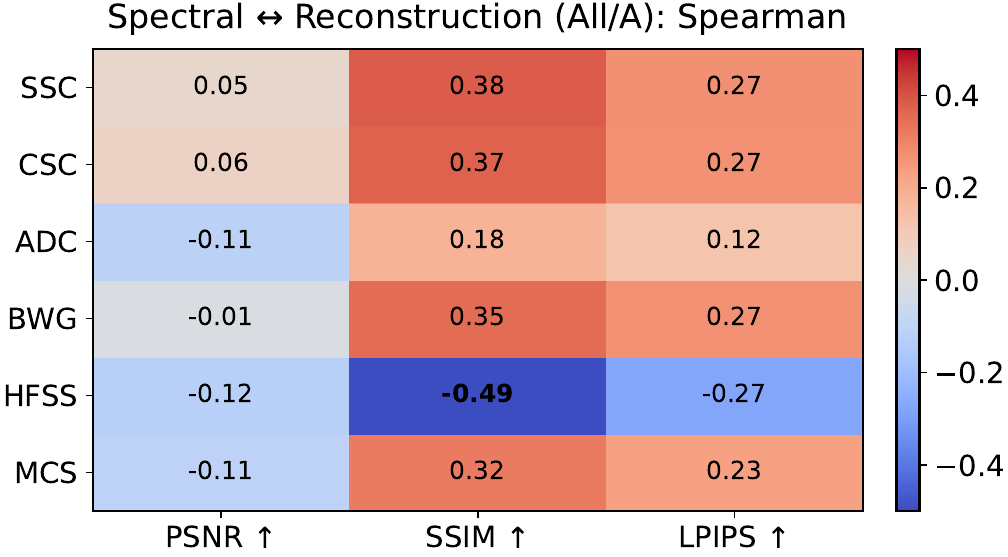}
    \hfill
    \includegraphics[width=0.49\linewidth]{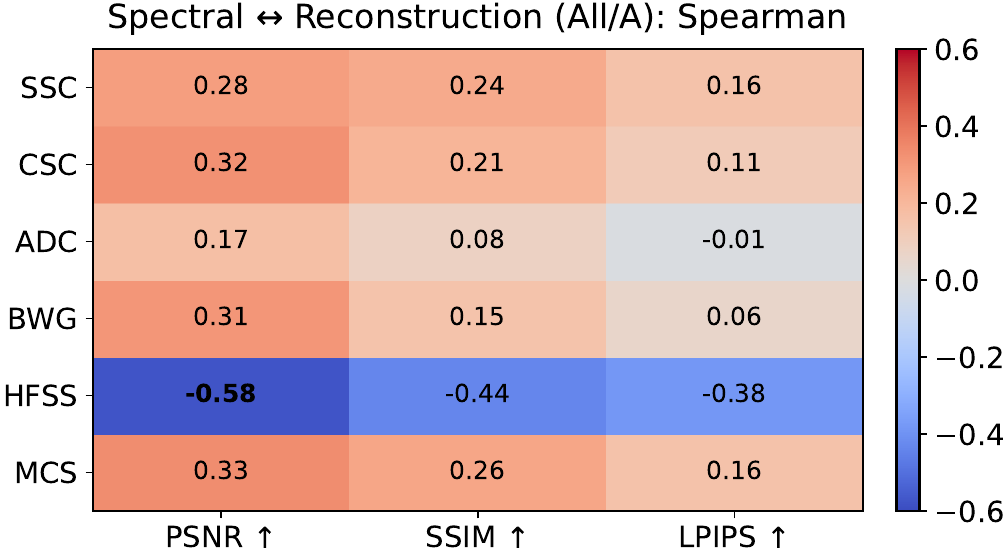}
    \vspace{-0.3em}
    \caption{Left: CLIP+Lanczos+MASt3R~\cite{leroy2024grounding}; Right: DINO+Lanczos+MASt3R~\cite{leroy2024grounding}.}
\end{subfigure}

\vspace{0.6em}
\begin{subfigure}[t]{\linewidth}
  \centering
  \includegraphics[width=0.49\linewidth]{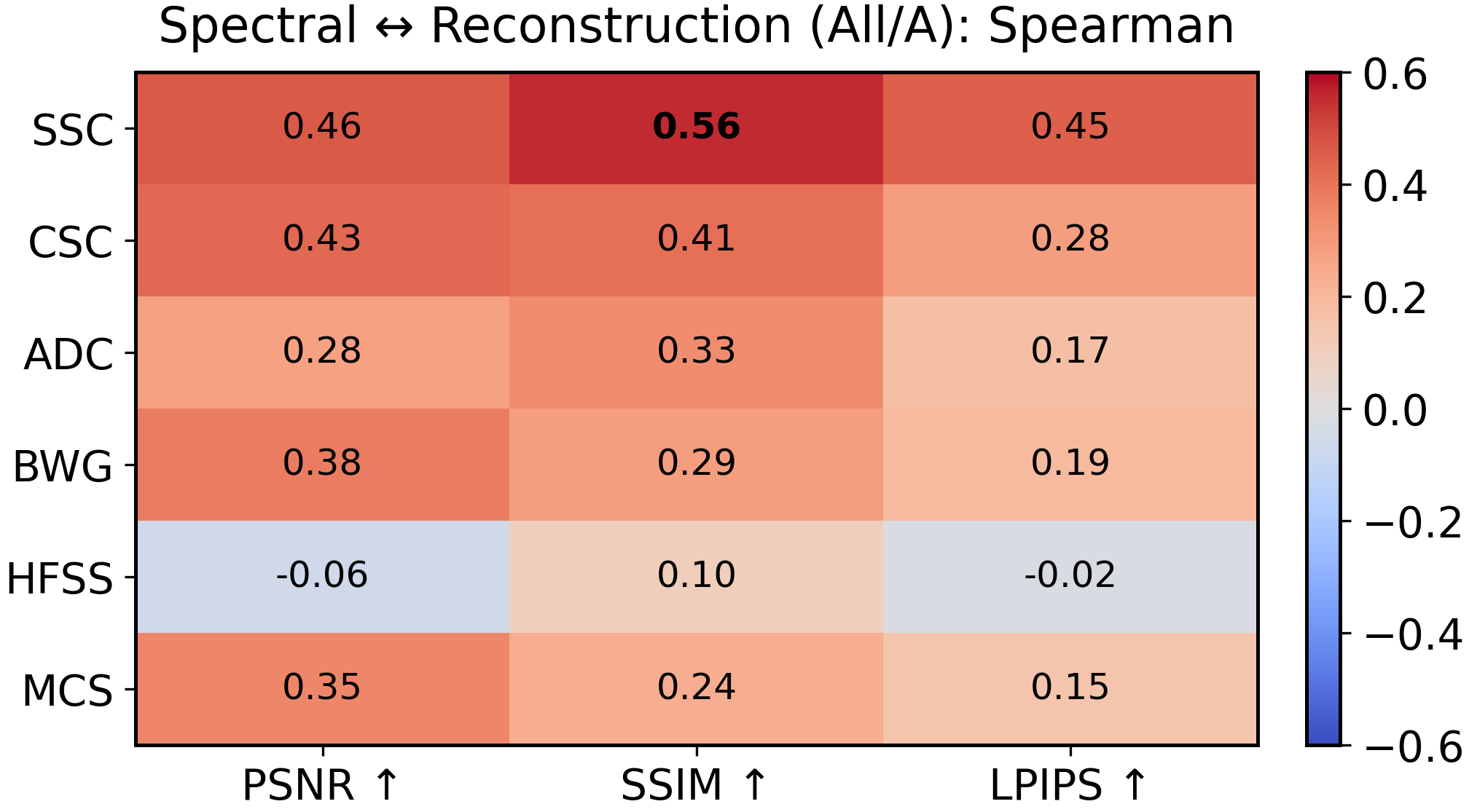}
  \hfill
  \includegraphics[width=0.49\linewidth]{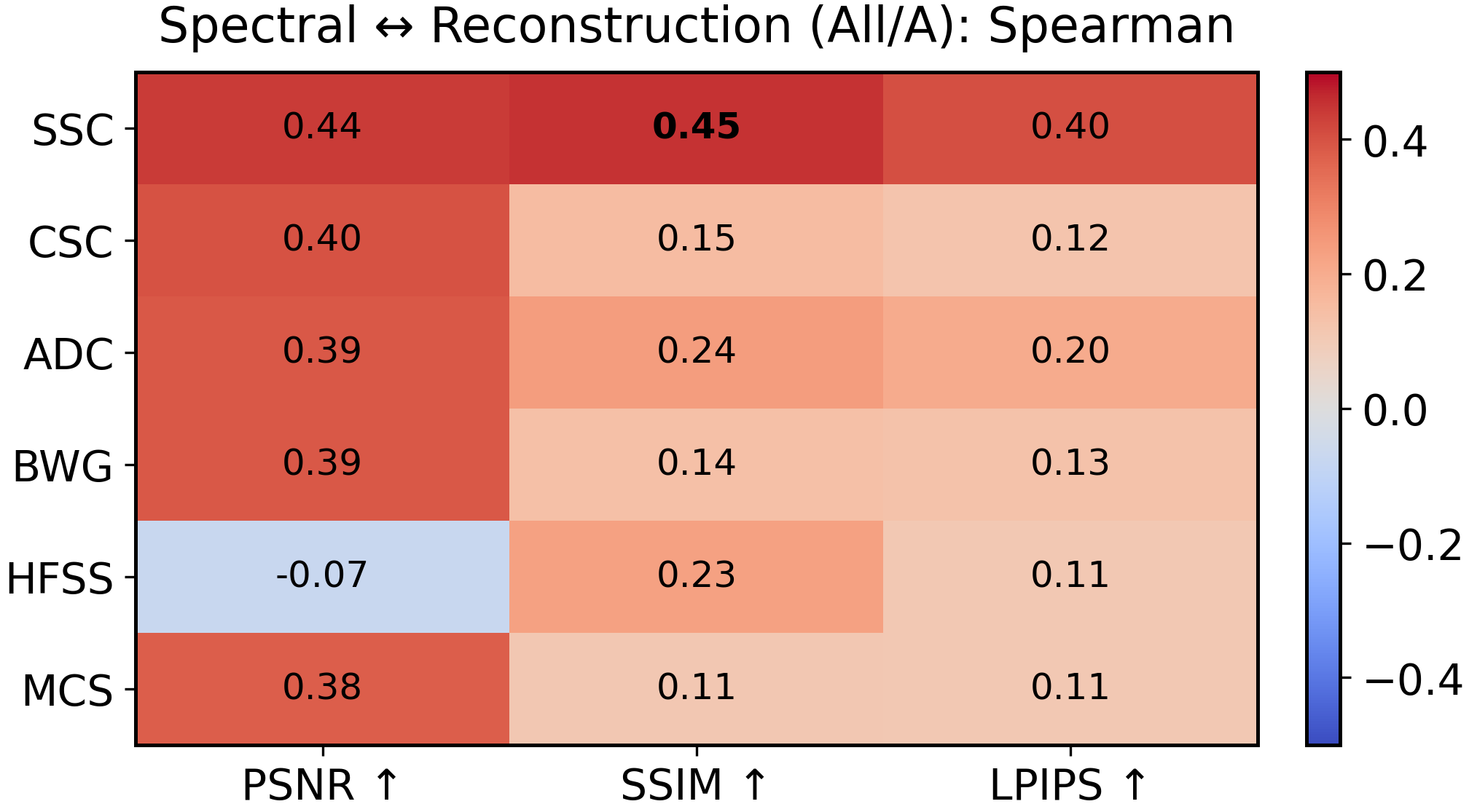}
  \vspace{-0.3em}
  \caption{Left: CLIP+JAFAR~\cite{couairon2025jafar}+DUSt3R~\cite{wang2024dust3r}; Right: DINO+JAFAR~\cite{couairon2025jafar}+DUSt3R~\cite{wang2024dust3r}.}
\end{subfigure}

\vspace{0.6em}
\begin{subfigure}[t]{\linewidth}
  \centering
  \includegraphics[width=0.49\linewidth]{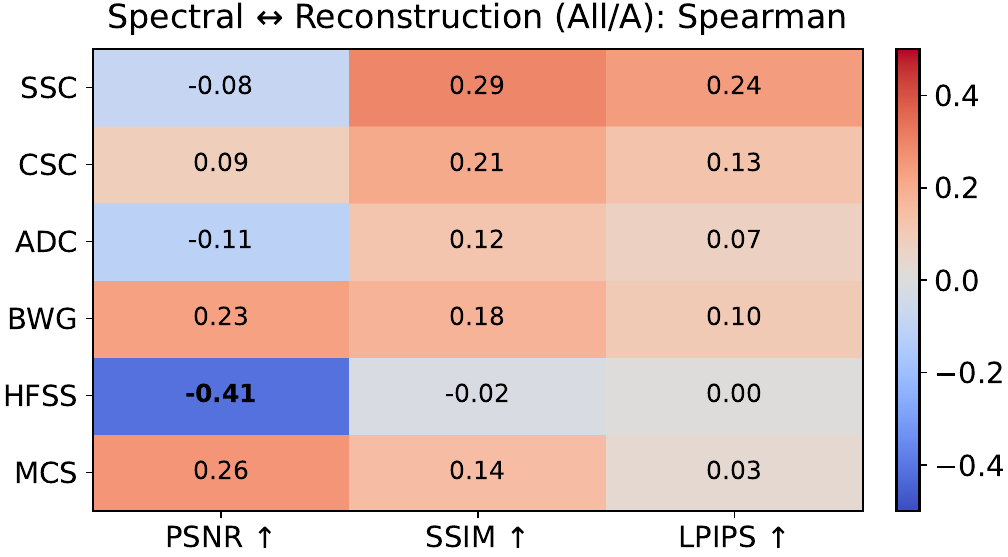}
  \hfill
  \includegraphics[width=0.49\linewidth]{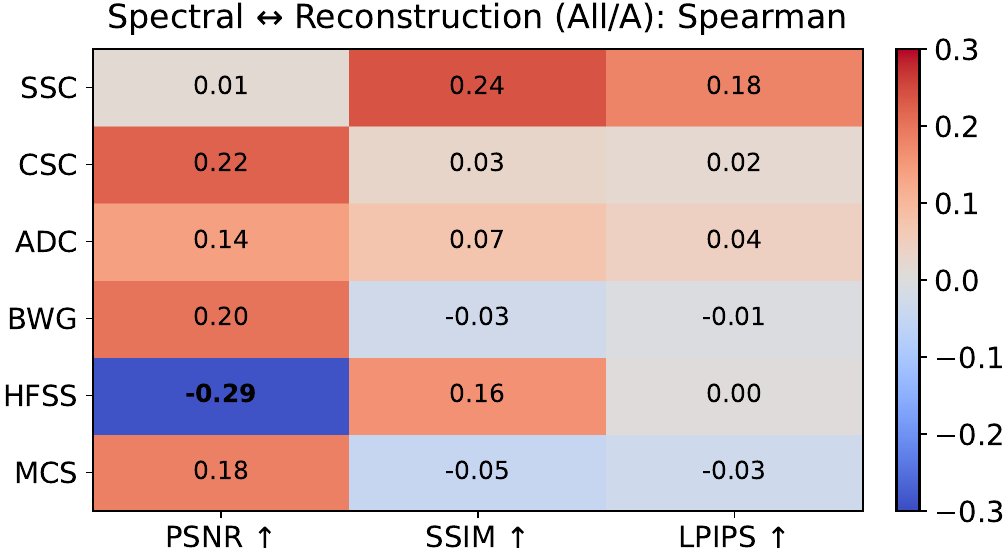}
  \vspace{-0.3em}
  \caption{Left: CLIP+JAFAR~\cite{couairon2025jafar}+MASt3R~\cite{leroy2024grounding}; Right: DINO+JAFAR~\cite{couairon2025jafar}+MASt3R~\cite{leroy2024grounding}.}
\end{subfigure}
\caption{Spectral–reconstruction correlations.
Each row shows the Spearman correlation heatmap.
\textbf{Results of other upsamplers are provided in the Supplementary Material.}
}
\label{fig:heatmap}
\end{figure}

\begin{figure}[htbp]
\centering
\captionsetup[subfigure]{font=small, labelfont=bf}
\centering
\captionsetup[subfigure]{font=small, labelfont=bf}
% ===================== (c) =====================
\begin{subfigure}[t]{\linewidth}
  \centering
  \includegraphics[width=0.49\linewidth]{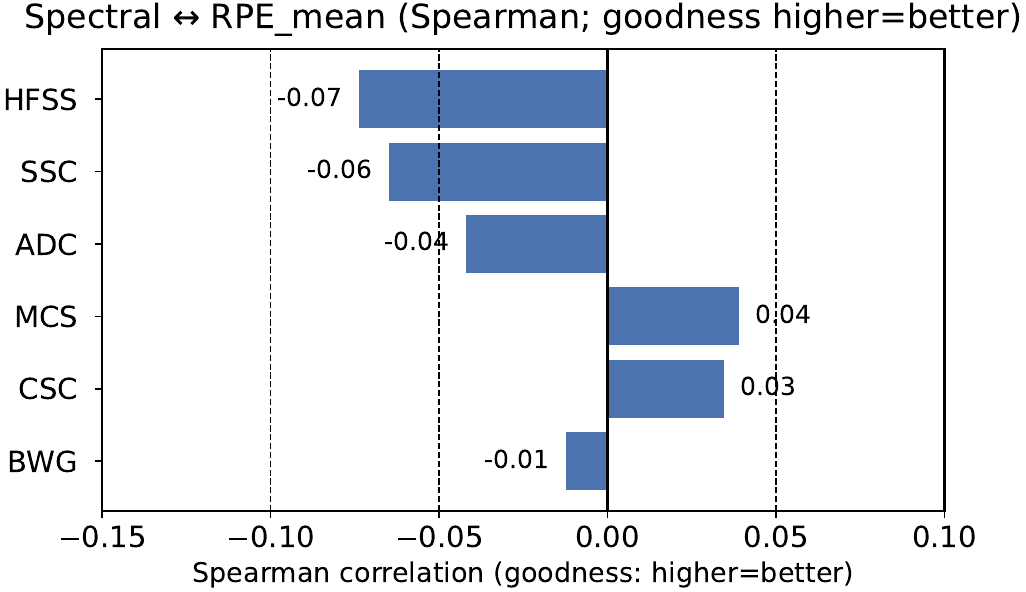}
  \hfill
  \includegraphics[width=0.49\linewidth]{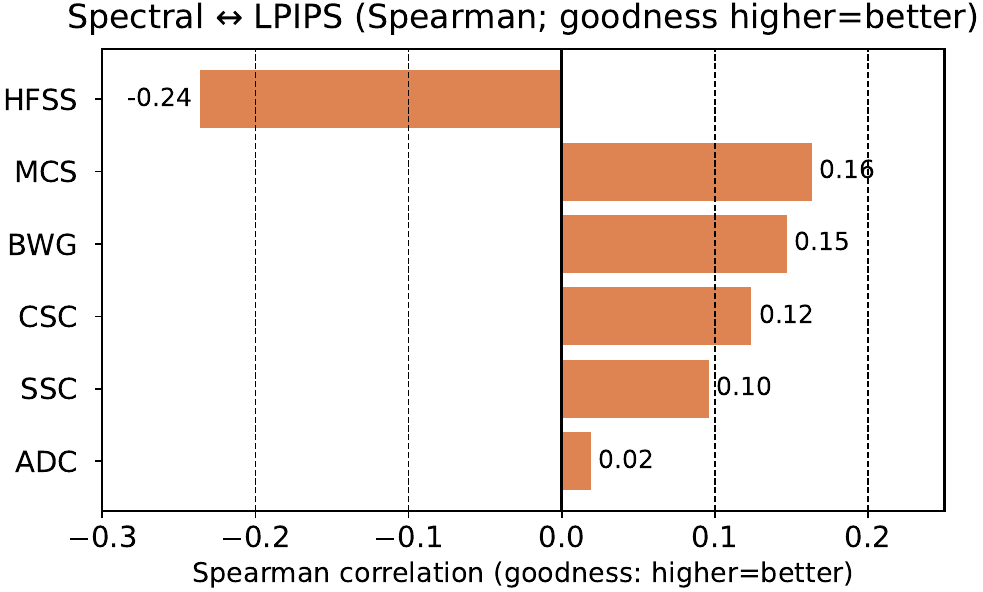}
  \vspace{-0.3em}
  \caption{Left (AG): Lanczos; Right (AT): Lanczos.}
\end{subfigure}
\vspace{0.6em}
% ===================== (d) =====================
\begin{subfigure}[t]{\linewidth}
  \centering
  \includegraphics[width=0.49\linewidth]{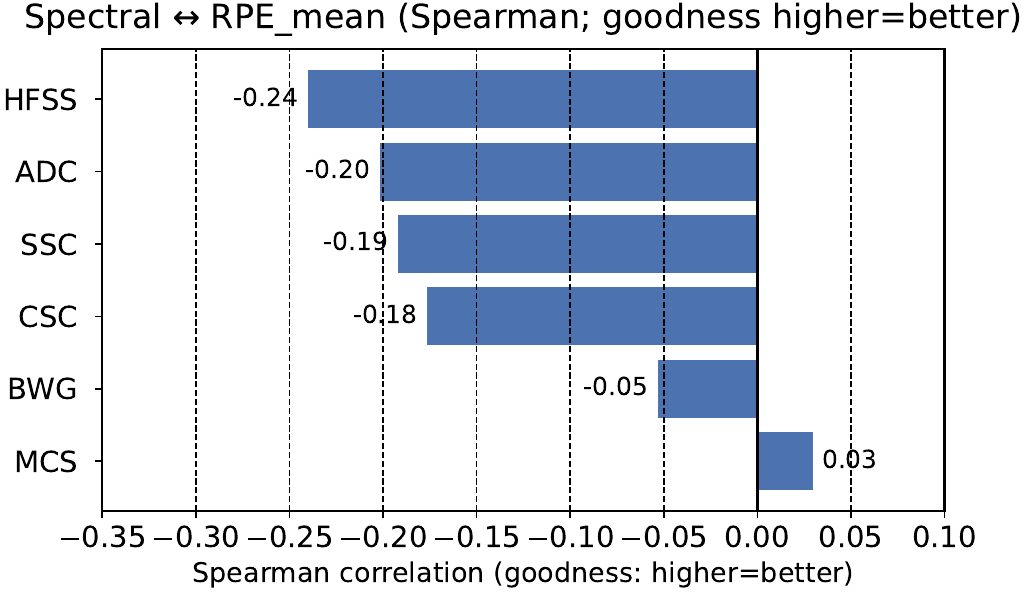}
  \hfill
  \includegraphics[width=0.49\linewidth]{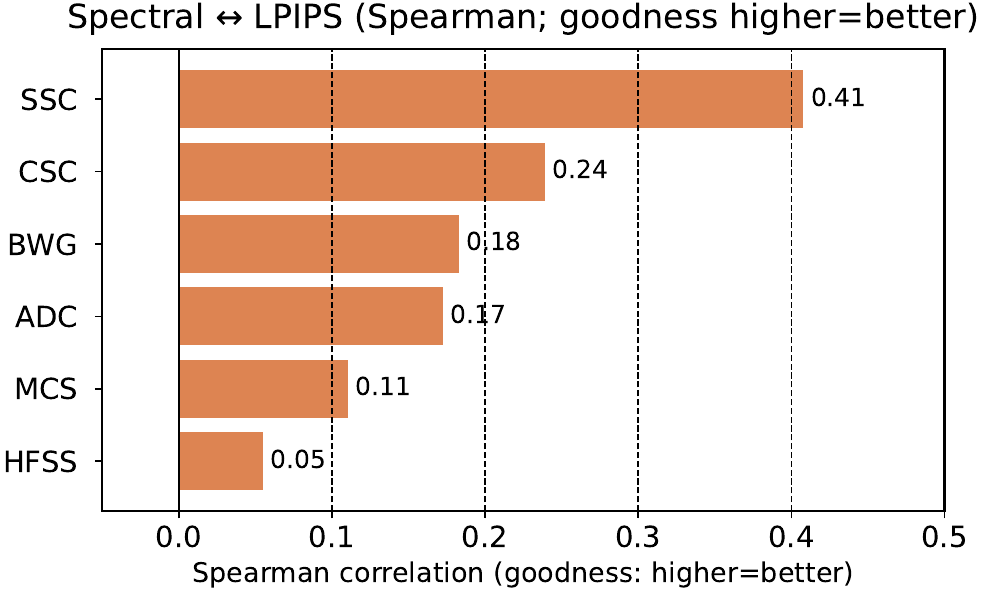}
  \vspace{-0.3em}
  \caption{Left (AG): JAFAR~\cite{couairon2025jafar}; Right (AT): JAFAR~\cite{couairon2025jafar}.}
\end{subfigure}
\caption{
Spectral diagnostics under geometry-only (AG) and texture-only (AT) settings (CLIP+DUSt3R~\cite{wang2024dust3r}). \textbf{Additional results are in the Supplementary Material.}
}
\label{fig:ag-at}
\vspace{0.02em}
% ===================== (a) =====================
\begin{subfigure}[t]{\linewidth}
  \centering
  \includegraphics[width=0.49\linewidth]{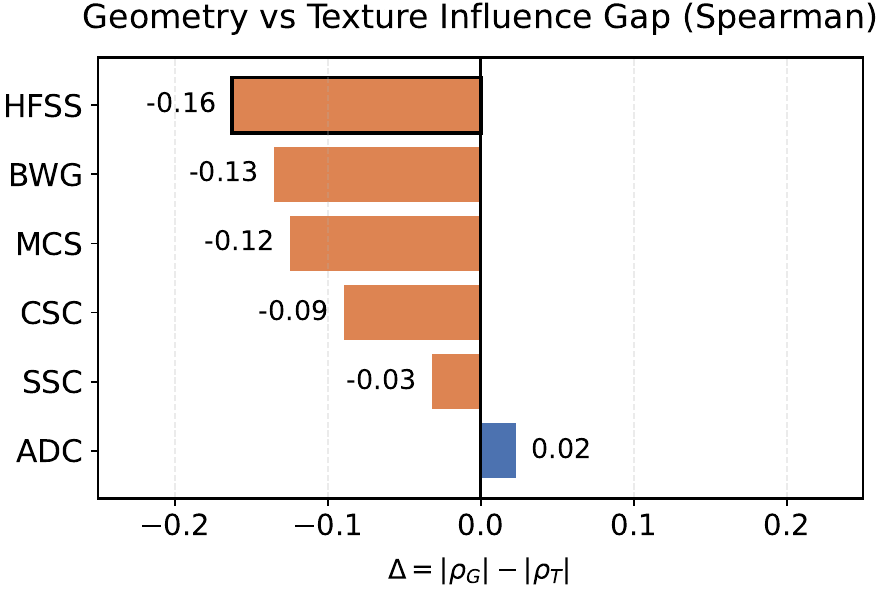}
  \hfill
   \includegraphics[width=0.49\linewidth]{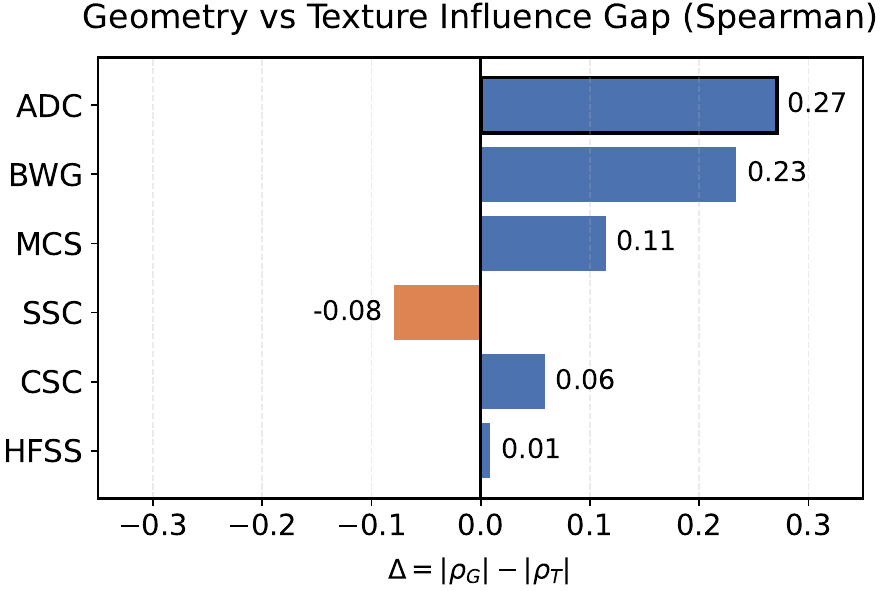}
  \vspace{-0.3em}
  \caption{Left: CLIP+Lanczos+DUSt3R~\cite{wang2024dust3r}; Right: CLIP+FeatUp~\cite{fu2024featup}+MASt3R~\cite{leroy2024grounding}.}
\end{subfigure}
\vspace{0.6em}
% ===================== (b) =====================
\begin{subfigure}[t]{\linewidth}
  \centering
  \includegraphics[width=0.49\linewidth]{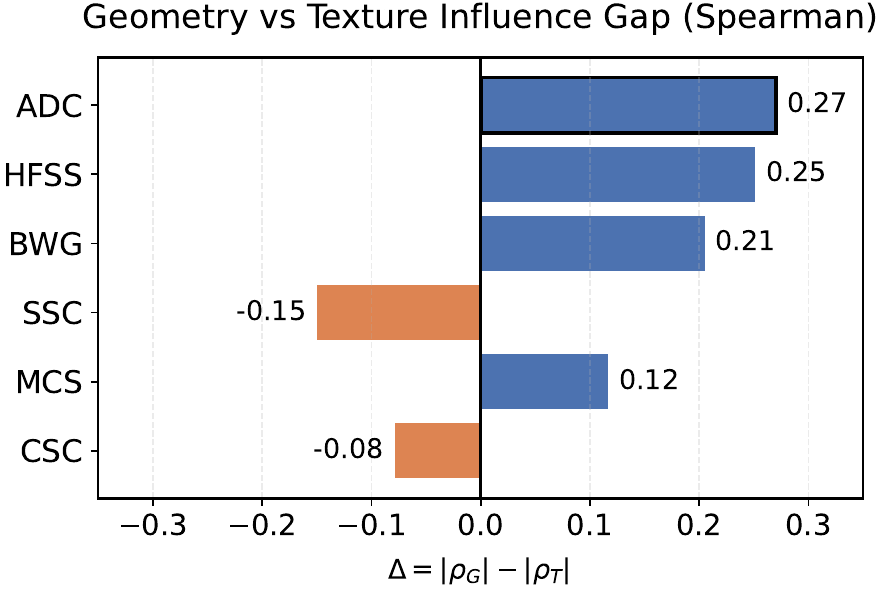}
  \hfill
  \includegraphics[width=0.49\linewidth]{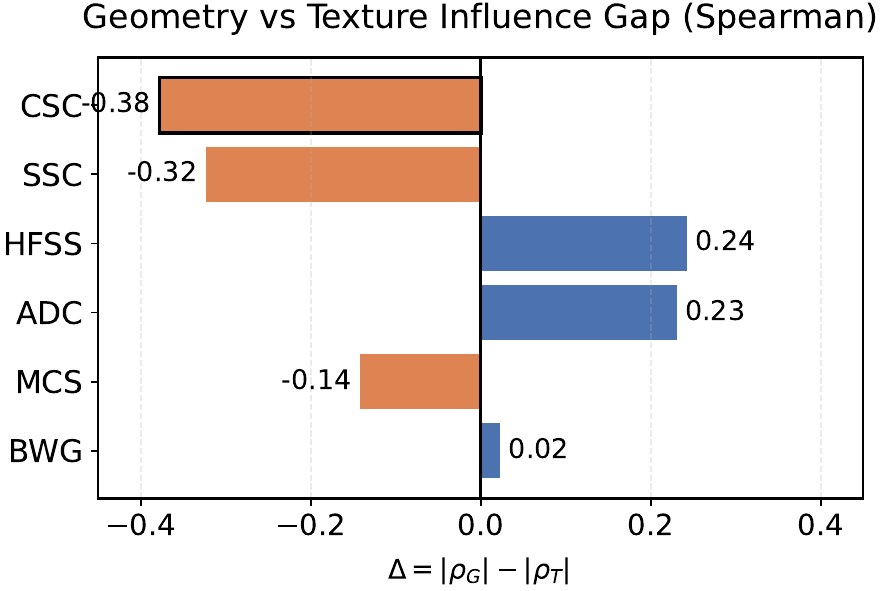}
  \vspace{-0.3em}
  \caption{Left: CLIP+AnyUp~\cite{wimmer2025anyup}+MASt3R~\cite{leroy2024grounding}, right: DINO+AnyUp~\cite{wimmer2025anyup}+MASt3R~\cite{leroy2024grounding}.}
\end{subfigure}
\caption{Geometry--texture influence gap.
\textbf{Results of other upsamplers are provided in the Supplementary Material.}
}
\label{fig:gap}
\end{figure}

\subsection{Experimental Results}
\noindent\textbf{Main Results.}
Tables~\ref{tab:dust3r-all} and~\ref{tab:mast3r-all} report cross-dataset average NVS performance using DUSt3R~\cite{wang2024dust3r} and MASt3R~\cite{leroy2024grounding} under three probing modes: All (geometry+texture), Geometry-only, and Texture-only. 
Under the joint All setting, classical interpolation methods such as bicubic and Lanczos achieve competitive performance compared with learnable upsampling approaches. 
Table~\ref{tab:dust3r-clip} further reports per-dataset results using the CLIP backbone. 
Despite dataset-specific variations, similar trends remain consistent across LLFF, DL3DV, Casual, MipNeRF360, MVImgNet, and T\&T. More experimental results are provided in the Supplementary Material. Fig.~\ref{fig:NVS} shows visualizations of images synthesized using different upsamplers.

\noindent\textbf{Spectral Diagnostics and Their Relationship with NVS Quality.}
We visualize the relationship between spectral diagnostics and NVS quality using three complementary analyses.
Fig.~\ref{fig:heatmap} shows cross-scene Spearman correlations between spectral diagnostics and NVS metrics under the joint All probing setting. More experimental results are provided in the Supplementary Material.
To further disentangle geometric and perceptual effects, we introduce two specialized probing modes:

\textbf{AG (geometry).} 
For each scene, we quantify geometric consistency by the mean relative pose error $\mathrm{RPE}_{\text{mean}}$ computed from the optimized camera poses.
Since $\mathrm{RPE}_{\text{mean}}$ is a lower-is-better quantity, we convert it to a unified higher-is-better geometry score as
$\mathrm{AG} \;=\; -\,\mathrm{RPE}_{\text{mean}}$ .

\textbf{AT (texture/appearance).}
We measure perceptual appearance using LPIPS from the NVS evaluation.
Because LPIPS is also lower-is-better, we similarly convert it to a unified higher-is-better appearance score:
$\mathrm{AT} \;=\; -\,\mathrm{LPIPS}$.

To quantify geometry–texture asymmetry, we further compute the influence gap
$\Delta = |\rho_G| - |\rho_T|$,
where $\rho_G$ and $\rho_T$ denote correlations under AG and AT probing, respectively (Fig.~\ref{fig:gap}).
Positive values indicate stronger geometry coupling, whereas negative values indicate stronger texture coupling. Overall, we have three key findings.

\noindent\textcolor{blue}{1) Structural spectral consistency predicts NVS quality, while HFSS shows negative correlations, challenging the common assumption that enhancing high-frequency details improves reconstruction.}
Across backbone–upsampler configurations, the structural diagnostics SSC and CSC consistently exhibit the strongest positive correlations with reconstruction quality under the joint All setting (Fig.~\ref{fig:heatmap}). 
This indicates that preserving global spectral structure and coherence is crucial for stable 2D-to-3D reconstruction. 
Although other factors such as magnitude redistribution (ADC/BWG/MCS) also contribute, NVS performance mainly depends on the joint preservation of spectral magnitude distribution and structural coherence. 
Notably, HFSS frequently shows negative correlations with overall reconstruction quality (All)(Fig.~\ref{fig:heatmap}), as well as with geometry (AG) and texture (AT) (Fig.~\ref{fig:ag-at}). 
This indicates that stronger high-frequency spectral drift is often associated with worse reconstruction performance. 
These observations suggest that simply emphasizing high-frequency details, a common objective of many learnable upsampling methods, does not necessarily improve 3D reconstruction.

\noindent\textcolor{blue}{2) Geometry and texture respond differently to spectral perturbations.} 
The AG and AT analyses reveal distinct sensitivities of geometry and appearance to spectral variations (Fig.~\ref{fig:gap}). 
In particular, the amplitude distribution metric ADC frequently exhibits positive influence gaps across multiple interpolation and learned upsampling methods, indicating stronger coupling with geometry-related reconstruction metrics. 
However, the magnitude of this effect varies depending on the backbone encoder and the 3D reconstructor. 
In contrast, the structural spectral consistency metrics (SSC and CSC) tend to influence texture fidelity slightly more than geometric accuracy but this trend can also be affected by the backbone encoder and the 3D reconstructor.

\noindent\textcolor{blue}{3) Classical interpolation remains competitive with learnable upsamplers, and upsampling effectiveness depends on the reconstruction model.}
Although learnable upsamplers reshape spectral statistics (Fig.~\ref{fig:heatmap}), their reconstruction improvements are often modest and backbone-dependent (Tables~\ref{tab:dust3r-all},~\ref{tab:mast3r-all}, and~\ref{tab:dust3r-clip}). Classical interpolation methods such as Lanczos and Bicubic frequently achieve comparable NVS performance with learnable upsamplers, highlighting that spectral–reconstruction interactions, rather than spatial detail enhancement alone, largely determine the final reconstruction quality.
Furthermore, with DUSt3R~\cite{wang2024dust3r} (Table~\ref{tab:dust3r-all}), the simple NSM baseline yields the weakest geometry-only performance. 
In contrast, with MASt3R~\cite{leroy2024grounding} (Table~\ref{tab:mast3r-all}), NSM becomes more competitive and can outperform several interpolation baselines, suggesting that the impact of upsampling depends on how the reconstructor utilizes feature representations.

%%%%%%%%%%%%%%%%%%%%%%%%%%%%%%%%%%%%%%%%%%

% in preamble
% \usepackage{graphicx}
% \usepackage{subcaption}
% \captionsetup{font=small}
% \captionsetup[subfigure]{font=small, labelfont=bf}

\section{Limitations and Future Work}
This work is primarily evaluation-oriented and does not explicitly incorporate these findings into the design of feature upsamplers. 
Future work will build on these insights to develop upsamplers that better preserve cross-view structural consistency. 
In particular, integrating spectral diagnostics into the training objectives of learnable upsamplers may guide frequency redistribution and phase alignment, improving the robustness of 2D-to-3D representations.

\section{Conclusion}
We investigate how feature upsampling influences 3D awareness in 2D-to-3D pipelines and examine whether the common assumption of existing state-of-the-art learnable upsamplers truly benefits 3D reconstruction. 
To analyze this effect, we introduce a spectral diagnostic framework that decomposes upsampling-induced changes into magnitude redistribution, phase alignment, and directional structure. 
We systematically evaluate classical interpolation and learnable upsampling methods under identical settings and analyze their correlations with NVS quality.
Our experiments reveal three key findings. 
First, structural spectral consistency (SSC/CSC) is the strongest predictor of NVS quality, whereas HFSS often shows negative correlations with reconstruction performance. This indicates that stronger high-frequency spectral drift tends to degrade reconstruction quality, suggesting that simply emphasizing high-frequency details does not necessarily benefit 3D reconstruction.
Second, the amplitude distribution metric ADC frequently exhibits stronger coupling with geometry-related reconstruction metrics, whereas structural spectral consistency metrics (SSC and CSC) tend to correlate more strongly with texture fidelity. These effects may vary depending on the backbone encoder and the 3D reconstructor. 
Third, learnable upsamplers do not consistently outperform classical interpolation, and the effectiveness of upsampling depends on the reconstruction model.

\title{Supplementary: Spectral Probing of Feature Upsamplers in 2D-to-3D Scene Reconstruction} 
\author{
Ling Xiao\inst{1}\orcidlink{0000-0002-4650-8841} \and
Yuliang Xiu\inst{2} \orcidlink{0000-0003-0165-5909} \and
Yue Chen\inst{2} \and
Guoming Wang\inst{3} \and
Toshihiko Yamasaki\inst{4} \orcidlink{0000-0002-1784-2314}
}

\authorrunning{L. Xiao et al.}

\institute{
Hokkaido University, Sapporo, Japan\\
\email{ling@ist.hokudai.ac.jp}
\and
Westlake University, Hangzhou, China\\
\email{\{xiuyuliang,faneggchen\}@westlake.edu.cn}
\and
Zhejiang University, Hangzhou, China\\
\email{nb21013@zju.edu.cn}
\and
The University of Tokyo, Tokyo, Japan\\
\email{yamasaki@cvm.t.u-tokyo.ac.jp}
}

\maketitle

\section{More Experimental Results}
\label{sec:exp}

\subsection{Detailed Results Across Different Datasets.}

More detailed per-dataset NVS evaluation results are reported in 
Tables~\ref{tab:dust3r-dino},~\ref{tab:mast-dino}, and~\ref{tab:mast-clip}. 
These tables present results on six multi-view datasets (LLFF, DL3DV, Casual, MipNeRF360, MVImgNet, and T\&T) under the three probing modes: Geometry-only (G), Texture-only (T), and All (A).

Table~\ref{tab:dust3r-dino} reports results using DUSt3R with the DINO backbone. 
Across datasets, classical interpolation methods such as Bilinear, Bicubic, and Lanczos generally provide stable performance, while the no-upsampling baseline (NSM) consistently yields weaker geometry results. 
This confirms that dense spatial interpolation is important for maintaining geometric consistency when DUSt3R is used as the 3D reconstructor.

Tables~\ref{tab:mast-dino} and~\ref{tab:mast-clip} present the corresponding results using MASt3R with DINO and CLIP backbones. 
Compared with DUSt3R, the performance differences between upsampling strategies become smaller, and the NSM baseline often remains competitive across datasets. 
This observation further supports that the effectiveness of feature upsampling is reconstructor-dependent.

Despite dataset-specific variations, several consistent trends can be observed. 
First, classical interpolation methods remain strong baselines and often achieve performance comparable to learned upsampling methods. 
Second, the relative ranking of upsampling strategies is broadly consistent across datasets.

%%% BEGIN AUTOGENERATED %%%
\setlength{\tabcolsep}{8pt}
\begin{table*}[t!]
\caption{Per-dataset NVS performance using DUSt3R~\cite{wang2024dust3r} with the DINO backbone. 
Results are reported on six multi-view datasets (LLFF, DL3DV, Casual, MipNeRF360, MVImgNet, and T\&T) under three probing modes: Geometry-only, Texture-only, and All (geometry + texture). 
We evaluate PSNR ($\uparrow$), SSIM ($\uparrow$), and LPIPS ($\downarrow$). 
NSM denotes the no-upsampling baseline. 
Best results within each dataset and probing mode are highlighted in bold.}
\label{tab:dust3r-dino}
\setlength{\tabcolsep}{4pt}
\centering
\resizebox{\textwidth}{!}{
\begin{tabular}{l|>{\raggedleft\arraybackslash}p{0.9cm}>{\raggedleft\arraybackslash}p{0.9cm}>{\raggedleft\arraybackslash}p{0.9cm}|>{\raggedleft\arraybackslash}p{0.9cm}>{\raggedleft\arraybackslash}p{0.9cm}>{\raggedleft\arraybackslash}p{0.9cm}|>{\raggedleft\arraybackslash}p{0.9cm}>{\raggedleft\arraybackslash}p{0.9cm}>{\raggedleft\arraybackslash}p{0.9cm}|>{\raggedleft\arraybackslash}p{0.9cm}>{\raggedleft\arraybackslash}p{0.9cm}>{\raggedleft\arraybackslash}p{0.9cm}|>{\raggedleft\arraybackslash}p{0.9cm}>{\raggedleft\arraybackslash}p{0.9cm}>{\raggedleft\arraybackslash}p{0.9cm}|>{\raggedleft\arraybackslash}p{0.9cm}>{\raggedleft\arraybackslash}p{0.9cm}>{\raggedleft\arraybackslash}p{0.9cm}}
\toprule
\multicolumn{1}{c|}{} & \multicolumn{9}{c|}{LLFF} & \multicolumn{9}{c}{DL3DV} \\
\midrule
\multicolumn{1}{c|}{} & \multicolumn{3}{c|}{\textbf{G}eometry} & \multicolumn{3}{c|}{\textbf{T}exture} & \multicolumn{3}{c|}{\textbf{A}ll} & \multicolumn{3}{c|}{\textbf{G}eometry} & \multicolumn{3}{c|}{\textbf{T}exture} & \multicolumn{3}{c}{\textbf{A}ll} \\
\midrule
Upsamplers & \fontsize{8.5pt}{9pt}\selectfont{PSNR$\uparrow$} & \fontsize{8.5pt}{9pt}\selectfont{SSIM$\uparrow$} & \fontsize{8.5pt}{9pt}\selectfont{LPIPS$\downarrow$} & \fontsize{8.5pt}{9pt}\selectfont{PSNR$\uparrow$} & \fontsize{8.5pt}{9pt}\selectfont{SSIM$\uparrow$} & \fontsize{8.5pt}{9pt}\selectfont{LPIPS$\downarrow$} & \fontsize{8.5pt}{9pt}\selectfont{PSNR$\uparrow$} & \fontsize{8.5pt}{9pt}\selectfont{SSIM$\uparrow$} & \fontsize{8.5pt}{9pt}\selectfont{LPIPS$\downarrow$} & \fontsize{8.5pt}{9pt}\selectfont{PSNR$\uparrow$} & \fontsize{8.5pt}{9pt}\selectfont{SSIM$\uparrow$} & \fontsize{8.5pt}{9pt}\selectfont{LPIPS$\downarrow$} & \fontsize{8.5pt}{9pt}\selectfont{PSNR$\uparrow$} & \fontsize{8.5pt}{9pt}\selectfont{SSIM$\uparrow$} & \fontsize{8.5pt}{9pt}\selectfont{LPIPS$\downarrow$} & \fontsize{8.5pt}{9pt}\selectfont{PSNR$\uparrow$} & \fontsize{8.5pt}{9pt}\selectfont{SSIM$\uparrow$} & \fontsize{8.5pt}{9pt}\selectfont{LPIPS$\downarrow$} \\
\midrule
NSM    &18.54 &.8298 &   .1655 &23.04 &.8804 &   .0949 &21.97 &.8657 &   .1230 &16.11 &.7658 &   .2303 &20.98 &\textbf{.8408} &   \textbf{.1498} &\underline{22.22} &\underline{.8628} &   \underline{.1575} \\  \midrule
Bilinear    &25.01 &\underline{.9061} &  \underline{.0696} &23.97 &.8879 &   .0804 &24.08 &.8791 &   .1052 &\textbf{22.62} &\textbf{.8681} &   \textbf{.1453} &20.92 &.8395 &   .1509 &22.16 &.8619 &   .1582 \\
 \midrule
NN   &22.63 &.8501 &   .1275 &24.00 &.8882 &   .0803 &24.02 &.8797 &   .1050 &20.93 &.8489 &   .1695 &20.91 &.8394 &   .1513 &21.72 &.8570 &   .1620 \\
  \midrule
 Lanczos  &25.04 &.9060 &   .0703 &24.12 &.8903 &   \underline{.0782} &25.04 &\underline{.8884} &   \underline{.0904} &22.58 &.8662 &   .1458 &20.95 &.8399 &   .1506 &22.00 &.8618 &   .1582  \\  \midrule
 Bicubic   &25.01 &\textbf{.9067} &   \textbf{.0695} &24.09 &.8899 &   .0783 &\textbf{25.14} &\textbf{.8898} &   .0910 &\underline{22.61} &\underline{.8674} &   \underline{.1456} &\textbf{21.00} &\underline{.8402} &   \underline{.1502} &\textbf{22.32} &\textbf{.8660} &   \textbf{.1549} \\
 \midrule
 LiFT~\cite{suri2024lift}  &25.03 &.8999 &   .0752 &\underline{24.18} &\underline{.8900} &   \underline{.0782} & 24.88 &.8854 &   \textbf{.0892} &22.41 &.8607 &   .1515 &20.56 &.8311 &   .1564 &22.12 &.8621 &   .1581 \\  \midrule
JAFAR~\cite{couairon2025jafar}   &\textbf{25.24} &.9041 &   .0751 &22.10 &.8745 &   .1035 &22.15 &.8703 &   .1207 &21.99 &.8538 &   .1596 &\underline{20.99} &\underline{.8402} &   .1504 &21.94 &.8595 &   .1588 \\ \midrule
 AnyUp~\cite{wimmer2025anyup}  &\underline{25.17} &.9026 &   .0758 &\textbf{24.36} &\textbf{.8914} &   \textbf{.0763} &\underline{25.06} &.8842 &   .0959 &21.89 &.8558 &   .1558 &20.62 &.8334 &   .1556 &21.83 &.8605 &   .1607 \\
\midrule
\multicolumn{1}{c|}{} & \multicolumn{9}{c|}{Casual} & \multicolumn{9}{c}{MipNeRF360} \\
\midrule
\multicolumn{1}{c|}{} & \multicolumn{3}{c|}{\textbf{G}eometry} & \multicolumn{3}{c|}{\textbf{T}exture} & \multicolumn{3}{c|}{\textbf{A}ll} & \multicolumn{3}{c|}{\textbf{G}eometry} & \multicolumn{3}{c|}{\textbf{T}exture} & \multicolumn{3}{c}{\textbf{A}ll} \\
\midrule
Upsamplers & \fontsize{8.5pt}{9pt}\selectfont{PSNR$\uparrow$} & \fontsize{8.5pt}{9pt}\selectfont{SSIM$\uparrow$} & \fontsize{8.5pt}{9pt}\selectfont{LPIPS$\downarrow$} & \fontsize{8.5pt}{9pt}\selectfont{PSNR$\uparrow$} & \fontsize{8.5pt}{9pt}\selectfont{SSIM$\uparrow$} & \fontsize{8.5pt}{9pt}\selectfont{LPIPS$\downarrow$} & \fontsize{8.5pt}{9pt}\selectfont{PSNR$\uparrow$} & \fontsize{8.5pt}{9pt}\selectfont{SSIM$\uparrow$} & \fontsize{8.5pt}{9pt}\selectfont{LPIPS$\downarrow$} & \fontsize{8.5pt}{9pt}\selectfont{PSNR$\uparrow$} & \fontsize{8.5pt}{9pt}\selectfont{SSIM$\uparrow$} & \fontsize{8.5pt}{9pt}\selectfont{LPIPS$\downarrow$} & \fontsize{8.5pt}{9pt}\selectfont{PSNR$\uparrow$} & \fontsize{8.5pt}{9pt}\selectfont{SSIM$\uparrow$} & \fontsize{8.5pt}{9pt}\selectfont{LPIPS$\downarrow$} & \fontsize{8.5pt}{9pt}\selectfont{PSNR$\uparrow$} & \fontsize{8.5pt}{9pt}\selectfont{SSIM$\uparrow$} & \fontsize{8.5pt}{9pt}\selectfont{LPIPS$\downarrow$} \\
\midrule
NSM   &16.88 &.7773 &   .2446 &\underline{23.16} &.7981 &   \underline{.1732} &24.09 &.8439 &   .1854 &18.59 &.6927 &   .2926 &\underline{23.25} &\underline{.7310} &   .1727 &24.75 &.7527 &   .2164 \\  \midrule
Bilinear   &\underline{25.10} &\textbf{.8452} &   .1643 &23.09 &.7970 &   .1763 &\underline{24.84} &\underline{.8448} &   .1848 &\textbf{24.91} &\textbf{.7621} &   \underline{.1711} &23.14 &.7292 &   .1742 &24.70 &.7516 &   .2168 \\
 \midrule
NN    &22.96 &.8265 &   .2000 &23.14 &.7968 &   .1748 &24.48 &.8445 &   .1847 &23.22 &.7316 &   .2453 &23.15 &.7291 &   .1740 &24.73 &.7526 &   .2175 \\
  \midrule
 Lanczos &24.87 &.8439 &   \textbf{.1627} &23.15 &.7980 &   .1735 &24.44 &.8447 &   \textbf{.1841} &24.86 &\underline{.7613} &   \textbf{.1700} &23.16 &.7282 &   .1738 &\textbf{24.81} &\underline{.7529} &   .2156 \\  \midrule
 Bicubic  &25.08 &.8440 &   \underline{.1633} &23.12 &\textbf{.7987} &   \underline{.1732} &24.44 &.8436 &   .1873 &\underline{24.90} &.7610 &   .1716 &\underline{23.25} &.7309 &   .1726 &\underline{24.80} &\underline{.7529} &   .2139 \\
 \midrule
 LiFT~\cite{suri2024lift}    &\textbf{25.18} &.8423 &   .1716 &23.08 &.7962 &   .1737 &\textbf{25.07} &\textbf{.8467} &  \textbf{.1841} &24.59 &.7482 &   .1799 &\textbf{23.37} &\textbf{.7313} &   \textbf{.1714} &24.33 &.7430 &   \textbf{.2111} \\ \midrule
JAFAR~\cite{couairon2025jafar}    &24.98 &.8436 &   .1695 &20.67 &.7780 &   .2032 &24.63 &.8433 &   \underline{.1842} &24.61 &.7516 &   .1802 &21.35 &.7087 &   .2032 &24.66 &\textbf{.7530} &   \underline{.2143} \\   \midrule
 AnyUp~\cite{wimmer2025anyup}  &24.67 &\underline{.8448} &   .1677 &\textbf{23.33} &\underline{.7986} &   \textbf{.1715} &24.48 &.8426 &   .1883 &24.78 &.7564 &   .1757 &23.22 &.7306 &   \underline{.1724} &24.63 &.7504 &   .2220 \\
\midrule
\multicolumn{1}{c|}{} & \multicolumn{9}{c|}{MVImgNet} & \multicolumn{9}{c}{T\&T} \\
\midrule
\multicolumn{1}{c|}{} & \multicolumn{3}{c|}{\textbf{G}eometry} & \multicolumn{3}{c|}{\textbf{T}exture} & \multicolumn{3}{c|}{\textbf{A}ll} & \multicolumn{3}{c|}{\textbf{G}eometry} & \multicolumn{3}{c|}{\textbf{T}exture} & \multicolumn{3}{c}{\textbf{A}ll} \\
\midrule
Upsamplers & \fontsize{8.5pt}{9pt}\selectfont{PSNR$\uparrow$} & \fontsize{8.5pt}{9pt}\selectfont{SSIM$\uparrow$} & \fontsize{8.5pt}{9pt}\selectfont{LPIPS$\downarrow$} & \fontsize{8.5pt}{9pt}\selectfont{PSNR$\uparrow$} & \fontsize{8.5pt}{9pt}\selectfont{SSIM$\uparrow$} & \fontsize{8.5pt}{9pt}\selectfont{LPIPS$\downarrow$} & \fontsize{8.5pt}{9pt}\selectfont{PSNR$\uparrow$} & \fontsize{8.5pt}{9pt}\selectfont{SSIM$\uparrow$} & \fontsize{8.5pt}{9pt}\selectfont{LPIPS$\downarrow$} & \fontsize{8.5pt}{9pt}\selectfont{PSNR$\uparrow$} & \fontsize{8.5pt}{9pt}\selectfont{SSIM$\uparrow$} & \fontsize{8.5pt}{9pt}\selectfont{LPIPS$\downarrow$} & \fontsize{8.5pt}{9pt}\selectfont{PSNR$\uparrow$} & \fontsize{8.5pt}{9pt}\selectfont{SSIM$\uparrow$} & \fontsize{8.5pt}{9pt}\selectfont{LPIPS$\downarrow$} & \fontsize{8.5pt}{9pt}\selectfont{PSNR$\uparrow$} & \fontsize{8.5pt}{9pt}\selectfont{SSIM$\uparrow$} & \fontsize{8.5pt}{9pt}\selectfont{LPIPS$\downarrow$} \\
\midrule
NSM   &18.69 &.7734 &   .2132 &21.22 &.7949 &   .1460 &\textbf{24.07} &\textbf{.8167} &   .1630 &19.97 &.8474 &   .1809 &23.74 &.8773 &   .1125 &24.61 &\underline{.8882} &  \textbf{.1313} \\  \midrule
Bilinear   &23.72 &.8271 &   .1406 &21.05 &.7904 &   .1497 &23.92 &.8142 &   .1635 &\textbf{24.57} &\textbf{.8812} &   .1298 &23.30 &.8717 &   .1161 &\textbf{24.78} &\underline{.8882} &   .1322 \\
 \midrule
NN    &22.32 &.7948 &   .1908 &21.04 &.7905 &   .1504 &23.95 &.8155 &   .1624 &23.19 &.8766 &   .1412 &\underline{24.00} &\underline{.8824} &   \underline{.1047} &\underline{24.75} &\textbf{.8883} &   \underline{.1320} \\
 \midrule
 Lanczos  &\textbf{24.23} &\textbf{.8301} &   \textbf{.1249} &\underline{21.24} &.7953 &   .1460 &\underline{24.04} &\textbf{.8167} &   .1628 &24.14 &.8778 &   .1314 &23.66 &.8771 &   .1129 &24.10 &.8805 &   .1381 \\ \midrule
 Bicubic  &\underline{24.13} &\underline{.8275} &   \underline{.1263} &21.17 &\underline{.7954} &   \underline{.1453} &23.99 &.8157 &   \textbf{.1616} &24.42 &.8799 &   \underline{.1293} &\textbf{24.36} & \textbf{.8873} &   \textbf{.1022} &24.22 &.8814 &   .1376 \\
\midrule
 LiFT~\cite{suri2024lift}    &23.77 &.8131 &   .1376 &\textbf{21.29} &\textbf{.7957} &   \textbf{.1452} &23.74 &.8128 &   \underline{.1619} &23.97 &.8759 &   .1337 &23.84 &.8765 &   .1136 &24.13 &.8835 &   .1379 \\  \midrule
JAFAR~\cite{couairon2025jafar}    &23.79 &.8163 &   .1362 &20.03 &.7681 &   .1713 &24.01 &\underline{.8166} &   .1624 &24.32 &.8803 &   .1313 &23.95 &.8810 &   .1090 &24.26 &.8823 &   .1377 \\  \midrule
 AnyUp~\cite{wimmer2025anyup}   &23.84 &.8179 &   .1366 &21.18 &.7937 &   .1455 &\underline{24.04} &.8141 &   .1691 &\underline{24.44} &\underline{.8806} &   \textbf{.1287} &23.73 &.8779 &   .1118 &24.23 &.8817 &   .1392 \\
\bottomrule
\end{tabular}
}
\end{table*}
%%% END AUTOGENERATED %%%

%%% BEGIN AUTOGENERATED %%%
\setlength{\tabcolsep}{8pt}
\begin{table*}[t!]
\caption{Per-dataset NVS performance using MASt3R~\cite{leroy2024grounding} with the DINO backbone. 
Results are reported on six multi-view datasets (LLFF, DL3DV, Casual, MipNeRF360, MVImgNet, and T\&T) under three probing modes: Geometry-only, Texture-only, and All (geometry + texture). 
We evaluate PSNR ($\uparrow$), SSIM ($\uparrow$), and LPIPS ($\downarrow$). 
NSM denotes the no-upsampling baseline. 
Best results within each dataset and probing mode are highlighted in bold.}
\label{tab:mast-dino}
\setlength{\tabcolsep}{4pt}
\centering
\resizebox{\textwidth}{!}{
\begin{tabular}{l|>{\raggedleft\arraybackslash}p{0.9cm}>{\raggedleft\arraybackslash}p{0.9cm}>{\raggedleft\arraybackslash}p{0.9cm}|>{\raggedleft\arraybackslash}p{0.9cm}>{\raggedleft\arraybackslash}p{0.9cm}>{\raggedleft\arraybackslash}p{0.9cm}|>{\raggedleft\arraybackslash}p{0.9cm}>{\raggedleft\arraybackslash}p{0.9cm}>{\raggedleft\arraybackslash}p{0.9cm}|>{\raggedleft\arraybackslash}p{0.9cm}>{\raggedleft\arraybackslash}p{0.9cm}>{\raggedleft\arraybackslash}p{0.9cm}|>{\raggedleft\arraybackslash}p{0.9cm}>{\raggedleft\arraybackslash}p{0.9cm}>{\raggedleft\arraybackslash}p{0.9cm}|>{\raggedleft\arraybackslash}p{0.9cm}>{\raggedleft\arraybackslash}p{0.9cm}>{\raggedleft\arraybackslash}p{0.9cm}}
\toprule
\multicolumn{1}{c|}{} & \multicolumn{9}{c|}{LLFF} & \multicolumn{9}{c}{DL3DV} \\
\midrule
\multicolumn{1}{c|}{} & \multicolumn{3}{c|}{\textbf{G}eometry} & \multicolumn{3}{c|}{\textbf{T}exture} & \multicolumn{3}{c|}{\textbf{A}ll} & \multicolumn{3}{c|}{\textbf{G}eometry} & \multicolumn{3}{c|}{\textbf{T}exture} & \multicolumn{3}{c}{\textbf{A}ll} \\
\midrule
Upsamplers & \fontsize{8.5pt}{9pt}\selectfont{PSNR$\uparrow$} & \fontsize{8.5pt}{9pt}\selectfont{SSIM$\uparrow$} & \fontsize{8.5pt}{9pt}\selectfont{LPIPS$\downarrow$} & \fontsize{8.5pt}{9pt}\selectfont{PSNR$\uparrow$} & \fontsize{8.5pt}{9pt}\selectfont{SSIM$\uparrow$} & \fontsize{8.5pt}{9pt}\selectfont{LPIPS$\downarrow$} & \fontsize{8.5pt}{9pt}\selectfont{PSNR$\uparrow$} & \fontsize{8.5pt}{9pt}\selectfont{SSIM$\uparrow$} & \fontsize{8.5pt}{9pt}\selectfont{LPIPS$\downarrow$} & \fontsize{8.5pt}{9pt}\selectfont{PSNR$\uparrow$} & \fontsize{8.5pt}{9pt}\selectfont{SSIM$\uparrow$} & \fontsize{8.5pt}{9pt}\selectfont{LPIPS$\downarrow$} & \fontsize{8.5pt}{9pt}\selectfont{PSNR$\uparrow$} & \fontsize{8.5pt}{9pt}\selectfont{SSIM$\uparrow$} & \fontsize{8.5pt}{9pt}\selectfont{LPIPS$\downarrow$} & \fontsize{8.5pt}{9pt}\selectfont{PSNR$\uparrow$} & \fontsize{8.5pt}{9pt}\selectfont{SSIM$\uparrow$} & \fontsize{8.5pt}{9pt}\selectfont{LPIPS$\downarrow$} \\
\midrule
 NSM   &\textbf{17.60} &\textbf{.8187} &   \textbf{.1717} &15.13 &\underline{.7452} &    \underline{.1924} &17.48 &\underline{.8140} &   \underline{.1707} &15.83 &\textbf{.7822} &   \textbf{.2176} &13.75 &.6983 &   .2611 &16.04 &\underline{.7866} &   \underline{.2243} \\
 \midrule
 Bilinear   &14.06 &.7144 &   .2571 &13.87 &.7051 &   .2647 &16.03 &.7493 &   .2252 &14.35 &.7114 &   .2658 &13.13 &.6677 &   .2897 &\textbf{17.53} &.7830 &   .2266 \\  \midrule
 NN    & \underline{16.85} &\underline{.7931} &   \underline{.1808} &15.12 &\textbf{.7463} &   \textbf{.1902} &17.31 &.8060 &   .1802 &\underline{16.19} &\underline{.7817} &   \underline{.2213} &13.67 &.6961 &   .2633 &15.81 &.7755 &   .2259 \\  \midrule
 Lanczos    &15.17 &.7256 &   .2290 &\underline{15.19} &.7250 &   .2523 &\underline{17.97} &.8126 &   .1723 &15.87 &.7641 &   .2317 &13.70 &.6976 &   .2620 &15.44 &.7838 &   .2251 \\
 \midrule
 Bicubic   &15.21 &.7397 &   .2091 &14.74 &.7228 &   .2251 &15.60 &.7492 &   .2036 &15.77 &.7587 &   .2323 &\underline{14.96} &\underline{.7321} &   \underline{.2470} &16.06 &.7689 &   .2319 \\ \midrule
 LiFT~\cite{suri2024lift}    &15.55 &.7431 &   .2201 &14.33 &.7115 &   .2479 &16.12 &.7586 &   .2272 &15.97 &.7413 &   .2293 &13.74 &.6984 &   .2702 &\underline{16.63} &\textbf{.7903} &   \textbf{.2237} \\ \midrule
 JAFAR~\cite{couairon2025jafar}   &15.75 &.7380 &   .2272 &15.11 &.7062 &   .2557 &15.69 &.7527 &   .1998 &\textbf{16.93} &.7611 &   .2317 &\textbf{16.87} &\textbf{.7537} &   \textbf{.2329} &15.56 &.7622 &   .2351 \\  \midrule
 AnyUp~\cite{wimmer2025anyup}     &13.58 &.7004 &   .2585 &\textbf{15.27} &.7264 &   .2507 &\textbf{18.22} &\textbf{.8227} &   \textbf{.1701} &14.68 &.7220 &   .2564 &13.73 &.6980 &   .2614 & 16.13 &.7800 &   .2295 \\
\midrule
\multicolumn{1}{c|}{} & \multicolumn{9}{c|}{Casual} & \multicolumn{9}{c}{MipNeRF360} \\
\midrule
\multicolumn{1}{c|}{} & \multicolumn{3}{c|}{\textbf{G}eometry} & \multicolumn{3}{c|}{\textbf{T}exture} & \multicolumn{3}{c|}{\textbf{A}ll} & \multicolumn{3}{c|}{\textbf{G}eometry} & \multicolumn{3}{c|}{\textbf{T}exture} & \multicolumn{3}{c}{\textbf{A}ll} \\
\midrule
Upsamplers & \fontsize{8.5pt}{9pt}\selectfont{PSNR$\uparrow$} & \fontsize{8.5pt}{9pt}\selectfont{SSIM$\uparrow$} & \fontsize{8.5pt}{9pt}\selectfont{LPIPS$\downarrow$} & \fontsize{8.5pt}{9pt}\selectfont{PSNR$\uparrow$} & \fontsize{8.5pt}{9pt}\selectfont{SSIM$\uparrow$} & \fontsize{8.5pt}{9pt}\selectfont{LPIPS$\downarrow$} & \fontsize{8.5pt}{9pt}\selectfont{PSNR$\uparrow$} & \fontsize{8.5pt}{9pt}\selectfont{SSIM$\uparrow$} & \fontsize{8.5pt}{9pt}\selectfont{LPIPS$\downarrow$} & \fontsize{8.5pt}{9pt}\selectfont{PSNR$\uparrow$} & \fontsize{8.5pt}{9pt}\selectfont{SSIM$\uparrow$} & \fontsize{8.5pt}{9pt}\selectfont{LPIPS$\downarrow$} & \fontsize{8.5pt}{9pt}\selectfont{PSNR$\uparrow$} & \fontsize{8.5pt}{9pt}\selectfont{SSIM$\uparrow$} & \fontsize{8.5pt}{9pt}\selectfont{LPIPS$\downarrow$} & \fontsize{8.5pt}{9pt}\selectfont{PSNR$\uparrow$} & \fontsize{8.5pt}{9pt}\selectfont{SSIM$\uparrow$} & \fontsize{8.5pt}{9pt}\selectfont{LPIPS$\downarrow$} \\
\midrule
 NSM   &18.94 &\textbf{.7976} &   .2321 &16.73 &.7059 &   .2594 &17.27 &.7768 &   .2497 &17.92 &\underline{.6721} &   .3068 &17.27 &.6145 &   .2770 &18.54 &\textbf{.6873} &   .2970 \\ \midrule
 Bilinear   &15.97 &.7238 &   .2713 &15.59 &.6950 &   .2800 &17.99 &.7715 &   .2462 &18.11 &.6367 &   .2834 &16.81 &.5969 &   .2885 &\textbf{20.01} &\underline{.6856} &   \textbf{.2677} \\  \midrule
 NN &16.70 &.7546 &   .2594 &16.70 &.7041 &   .2604 &17.62 &.7802 &   .2428 &19.71 &\textbf{.6842} &   .2779 &17.24 &.6138 &   .2771 &16.92 &.6314 &   .3524 \\  \midrule
 Lanczos  &\underline{19.14} &.7796 &   .2262 &17.39 &\underline{.7476} &   \underline{.2299} &17.18 &.7599 &   .2615 &18.48 &.6497 &   .2634 &\underline{18.09} &.6215 &   \underline{.2517} &18.54 &.6836 &   .2986 \\
 \midrule
 Bicubic  &\textbf{19.52} &\underline{.7851} &   \underline{.2218} &\textbf{18.29} &.7448 &   .2358 & \underline{18.94} & \underline{.7869} &   \textbf{.2370} &18.08 &.6318 &   .2794 &17.15 &.5998 &   .2780 &18.27 &.6490 &   .2962 \\  \midrule
 LiFT~\cite{suri2024lift}      &18.77 &.7803 &   .2219 &15.47 &.7017 &   .2667 &16.75 &.7597 &   .2544 &18.06 &.6395 &   .2736 &17.76 &.6097 &   .2653 &\underline{19.41} &.6823 &   \underline{.2866} \\ \midrule
 JAFAR~\cite{couairon2025jafar}  &17.71 &.7565 &   .2450 &16.00 &.7129 &   .2607 &\textbf{19.14} &\textbf{.7879} &   \underline{.2402} &\textbf{20.06} &.6720 &   \textbf{.2430} &18.06 &\textbf{.6227} &   .2598 &18.35 &.6509 &   .2928 \\   \midrule
 AnyUp~\cite{wimmer2025anyup}   &18.74 &.7849 &   \textbf{.2199} &\underline{17.41} & \textbf{.7496} &   \textbf{.2293} &16.68 &.7548 &   .2649 &\underline{19.91} &.6609 &   \underline{.2432} &\textbf{18.16} &\underline{.6222} &   \textbf{.2500} &18.49 &.6809 &   .2923 \\
\midrule
\multicolumn{1}{c|}{} & \multicolumn{9}{c|}{MVImgNet} & \multicolumn{9}{c}{T\&T} \\
\midrule
\multicolumn{1}{c|}{} & \multicolumn{3}{c|}{\textbf{G}eometry} & \multicolumn{3}{c|}{\textbf{T}exture} & \multicolumn{3}{c|}{\textbf{A}ll} & \multicolumn{3}{c|}{\textbf{G}eometry} & \multicolumn{3}{c|}{\textbf{T}exture} & \multicolumn{3}{c}{\textbf{A}ll} \\
\midrule
Upsamplers & \fontsize{8.5pt}{9pt}\selectfont{PSNR$\uparrow$} & \fontsize{8.5pt}{9pt}\selectfont{SSIM$\uparrow$} & \fontsize{8.5pt}{9pt}\selectfont{LPIPS$\downarrow$} & \fontsize{8.5pt}{9pt}\selectfont{PSNR$\uparrow$} & \fontsize{8.5pt}{9pt}\selectfont{SSIM$\uparrow$} & \fontsize{8.5pt}{9pt}\selectfont{LPIPS$\downarrow$} & \fontsize{8.5pt}{9pt}\selectfont{PSNR$\uparrow$} & \fontsize{8.5pt}{9pt}\selectfont{SSIM$\uparrow$} & \fontsize{8.5pt}{9pt}\selectfont{LPIPS$\downarrow$} & \fontsize{8.5pt}{9pt}\selectfont{PSNR$\uparrow$} & \fontsize{8.5pt}{9pt}\selectfont{SSIM$\uparrow$} & \fontsize{8.5pt}{9pt}\selectfont{LPIPS$\downarrow$} & \fontsize{8.5pt}{9pt}\selectfont{PSNR$\uparrow$} & \fontsize{8.5pt}{9pt}\selectfont{SSIM$\uparrow$} & \fontsize{8.5pt}{9pt}\selectfont{LPIPS$\downarrow$} & \fontsize{8.5pt}{9pt}\selectfont{PSNR$\uparrow$} & \fontsize{8.5pt}{9pt}\selectfont{SSIM$\uparrow$} & \fontsize{8.5pt}{9pt}\selectfont{LPIPS$\downarrow$} \\
\midrule
 NSM  &18.50 &\textbf{.7717} &   .2196 &\underline{17.26} &\underline{.7089} &   \underline{.2074} &18.24 &\textbf{.7693} &   .2130 &18.43 &.8086 &   .2056 &17.38 &\underline{.7993} &   .1920 &\textbf{19.99} &\textbf{.8570} &   \textbf{.1727} \\
 \midrule
 Bilinear  &18.60 &.7442 &   .1933 &16.93 &.7048 &   .2137 &\textbf{19.32} &.7623 &   \textbf{.2013} &19.00 &\underline{.8329} &   .1807 &17.00 &.7873 &   .2061 &18.71 &.8253 &   .1969 \\  \midrule
 NN    &18.21 &\underline{.7605} &   .2200 &\textbf{17.31} &\textbf{.7097} &   \textbf{.2070} &15.40 &.7097 &   .2527 &\underline{19.58} &\textbf{.8389} &   \underline{.1757} &17.45 &\textbf{.7994} &   .1921 &\underline{19.86} &\underline{.8516} &   \underline{.1766} \\ \midrule
 Lanczos  &18.63 &.7397 &   .1926 &16.70 &.7053 &   .2113 &18.04 & \underline{.7672} &   .2177 &\textbf{20.39} &\textbf{.8389} &   \textbf{.1744} &\underline{17.94} &.7939 &   .1933 &19.50 &.8448 &   .1816 \\
  \midrule
 Bicubic &18.58 &.7381 &   .1931 &16.66 &.7006 &   .2144 &18.93 &.7538 &   .2097 &18.98 &.8273 &   .1885 &\textbf{17.95} &.7955 &   .1929 &19.51 &.8402 &   .1828 \\ \midrule
 LiFT~\cite{suri2024lift}    &\underline{19.02} &.7427 &   .1910 &15.98 &.6910 &   .2185 &18.18 &.7483 &   .2287 &18.35 &.8173 &   .1938 &17.64 &.7922 &   \underline{.1916} &18.96 &.8376 &   .1776 \\  \midrule
 JAFAR~\cite{couairon2025jafar}   &18.96 &.7440 &   \underline{.1881} &16.61 &.6938 &   .2140 & \underline{19.02} &.7550 &   \underline{.2072} &18.44 &.8136 &   .1963 &17.32 &.7816 &   .2096 &18.88 &.8350 &   .1866 \\  \midrule
 AnyUp~\cite{wimmer2025anyup}    &\textbf{19.52} &.7522 &   \textbf{.1820} &17.21 &.7081 &   .2094 &18.75 &.7702 &   .2093 &18.30 &.8099 &   .2105 &17.93 &.7933 &   \textbf{.1903} &19.42 &.8351 &   .1839 \\
\bottomrule
\end{tabular}
}
\end{table*}
%%% END AUTOGENERATED %%%

%%% BEGIN AUTOGENERATED %%%
\setlength{\tabcolsep}{8pt}
\begin{table*}[t!]
\caption{Per-dataset NVS performance using MASt3R~\cite{leroy2024grounding} with the CLIP backbone. 
Results are reported on six multi-view datasets (LLFF, DL3DV, Casual, MipNeRF360, MVImgNet, and T\&T) under three probing modes: Geometry-only, Texture-only, and All (geometry + texture). 
We evaluate PSNR ($\uparrow$), SSIM ($\uparrow$), and LPIPS ($\downarrow$). 
NSM denotes the no-upsampling baseline. 
Best results within each dataset and probing mode are highlighted in bold.
}
\label{tab:mast-clip}
\setlength{\tabcolsep}{4pt}
\centering
\resizebox{\textwidth}{!}{
\begin{tabular}{l|>{\raggedleft\arraybackslash}p{0.9cm}>{\raggedleft\arraybackslash}p{0.9cm}>{\raggedleft\arraybackslash}p{0.9cm}|>{\raggedleft\arraybackslash}p{0.9cm}>{\raggedleft\arraybackslash}p{0.9cm}>{\raggedleft\arraybackslash}p{0.9cm}|>{\raggedleft\arraybackslash}p{0.9cm}>{\raggedleft\arraybackslash}p{0.9cm}>{\raggedleft\arraybackslash}p{0.9cm}|>{\raggedleft\arraybackslash}p{0.9cm}>{\raggedleft\arraybackslash}p{0.9cm}>{\raggedleft\arraybackslash}p{0.9cm}|>{\raggedleft\arraybackslash}p{0.9cm}>{\raggedleft\arraybackslash}p{0.9cm}>{\raggedleft\arraybackslash}p{0.9cm}|>{\raggedleft\arraybackslash}p{0.9cm}>{\raggedleft\arraybackslash}p{0.9cm}>{\raggedleft\arraybackslash}p{0.9cm}}
\toprule
\multicolumn{1}{c|}{} & \multicolumn{9}{c|}{LLFF} & \multicolumn{9}{c}{DL3DV} \\
\midrule
\multicolumn{1}{c|}{} & \multicolumn{3}{c|}{\textbf{G}eometry} & \multicolumn{3}{c|}{\textbf{T}exture} & \multicolumn{3}{c|}{\textbf{A}ll} & \multicolumn{3}{c|}{\textbf{G}eometry} & \multicolumn{3}{c|}{\textbf{T}exture} & \multicolumn{3}{c}{\textbf{A}ll} \\
\midrule
Upsamplers & \fontsize{8.5pt}{9pt}\selectfont{PSNR$\uparrow$} & \fontsize{8.5pt}{9pt}\selectfont{SSIM$\uparrow$} & \fontsize{8.5pt}{9pt}\selectfont{LPIPS$\downarrow$} & \fontsize{8.5pt}{9pt}\selectfont{PSNR$\uparrow$} & \fontsize{8.5pt}{9pt}\selectfont{SSIM$\uparrow$} & \fontsize{8.5pt}{9pt}\selectfont{LPIPS$\downarrow$} & \fontsize{8.5pt}{9pt}\selectfont{PSNR$\uparrow$} & \fontsize{8.5pt}{9pt}\selectfont{SSIM$\uparrow$} & \fontsize{8.5pt}{9pt}\selectfont{LPIPS$\downarrow$} & \fontsize{8.5pt}{9pt}\selectfont{PSNR$\uparrow$} & \fontsize{8.5pt}{9pt}\selectfont{SSIM$\uparrow$} & \fontsize{8.5pt}{9pt}\selectfont{LPIPS$\downarrow$} & \fontsize{8.5pt}{9pt}\selectfont{PSNR$\uparrow$} & \fontsize{8.5pt}{9pt}\selectfont{SSIM$\uparrow$} & \fontsize{8.5pt}{9pt}\selectfont{LPIPS$\downarrow$} & \fontsize{8.5pt}{9pt}\selectfont{PSNR$\uparrow$} & \fontsize{8.5pt}{9pt}\selectfont{SSIM$\uparrow$} & \fontsize{8.5pt}{9pt}\selectfont{LPIPS$\downarrow$} \\
\midrule
NSM  &\textbf{17.46} &\textbf{.8081} &   \textbf{.1773} &13.75 &.7039 &   .2465 &16.79 &\textbf{.7851} &   \textbf{.1899} &16.26 &\textbf{.7996} &   \textbf{.2065} &13.94 &.7168 &   .2494 &16.24 &\underline{.7976} &   \underline{.2136} \\
 \midrule
Bilinear &15.59 &.7424 &   .2066 &15.11 &.7220 &   .2308 &15.03 &.7064 &   .2687 &14.53 &.7368 &   .2476 &13.50 &.6992 &   .2647 &\underline{16.50} &.7489 &   .2376 \\ \midrule
NN   &\underline{16.52} &\underline{7701} &   .2153 &\underline{15.18} &\underline{.7259} &   .2491 &\underline{16.84} &\underline{.7833} &   .1919 &\textbf{16.82} &\underline{.7947} &   \underline{.2127} &13.88 &.7146 &   .2523 &16.17 &\textbf{.7987} &   \textbf{.2115} \\ \midrule
Lanczos  &14.19 &.7220 &   .2489 &14.38 &.7190 &   .2206 &16.19 &.7781 &   .2049 &15.36 &.7411 &   .2489 &14.06 &.7157 &   .2538 &\textbf{17.04} &.7968 &   .2181 \\
  \midrule
Bicubic  &14.71 &.7246 &   .2354 &14.36 &.7111 &   .2467 &16.19 &.7567 &   .2300 &14.39 &.7139 &   .2592 &13.37 &.6793 &   .2740 &14.50 &.7232 &   .2559 \\ \midrule
LoftUp~\cite{huang2025loftup}   &14.20 &.7157 &   .2359 &13.79 &.7015 &   .2550 &15.98 &.7495 &   .2078 &16.41 &.7497 &   .2383 &\textbf{15.80} &\underline{.7348} &   \underline{.2390} &14.73 &.7493 &   .2422 \\  \midrule
FeatUp~\cite{fu2024featup}    &16.19 &.7602 &   \underline{.1955} &\textbf{17.26} &\textbf{.7514} &   \textbf{.2154} &\textbf{18.30} &.7757 &   \underline{.1913} &14.84 &.7072 &   .2706 &15.16 &\textbf{.7431} &   \textbf{.2330} &15.95 &.7726 &   .2226 \\ \midrule
JAFAR~\cite{couairon2025jafar}   &14.54 &.6931 &   .2753 &14.17 &.6834 &   .2869 &15.07 &.7397 &   .2164 &\underline{16.50} &.7423 &   .2374 &\underline{15.40} &.7147 &   .2521 &14.23 &.7190 &   .2591 \\ \midrule
AnyUp~\cite{wimmer2025anyup} &14.82 &.7344 &   .2021 &14.31 &.7182 &   \underline{.2203} &16.48 &.7817 &   .1977 &15.12 &.7468 &   .2388 &14.13 &.7172 &   .2527 &16.27 &.7859 &   .2242 \\
\midrule
\multicolumn{1}{c|}{} & \multicolumn{9}{c|}{Casual} & \multicolumn{9}{c}{MipNeRF360} \\
\midrule
\multicolumn{1}{c|}{} & \multicolumn{3}{c|}{\textbf{G}eometry} & \multicolumn{3}{c|}{\textbf{T}exture} & \multicolumn{3}{c|}{\textbf{A}ll} & \multicolumn{3}{c|}{\textbf{G}eometry} & \multicolumn{3}{c|}{\textbf{T}exture} & \multicolumn{3}{c}{\textbf{A}ll} \\
\midrule
Upsamplers & \fontsize{8.5pt}{9pt}\selectfont{PSNR$\uparrow$} & \fontsize{8.5pt}{9pt}\selectfont{SSIM$\uparrow$} & \fontsize{8.5pt}{9pt}\selectfont{LPIPS$\downarrow$} & \fontsize{8.5pt}{9pt}\selectfont{PSNR$\uparrow$} & \fontsize{8.5pt}{9pt}\selectfont{SSIM$\uparrow$} & \fontsize{8.5pt}{9pt}\selectfont{LPIPS$\downarrow$} & \fontsize{8.5pt}{9pt}\selectfont{PSNR$\uparrow$} & \fontsize{8.5pt}{9pt}\selectfont{SSIM$\uparrow$} & \fontsize{8.5pt}{9pt}\selectfont{LPIPS$\downarrow$} & \fontsize{8.5pt}{9pt}\selectfont{PSNR$\uparrow$} & \fontsize{8.5pt}{9pt}\selectfont{SSIM$\uparrow$} & \fontsize{8.5pt}{9pt}\selectfont{LPIPS$\downarrow$} & \fontsize{8.5pt}{9pt}\selectfont{PSNR$\uparrow$} & \fontsize{8.5pt}{9pt}\selectfont{SSIM$\uparrow$} & \fontsize{8.5pt}{9pt}\selectfont{LPIPS$\downarrow$} & \fontsize{8.5pt}{9pt}\selectfont{PSNR$\uparrow$} & \fontsize{8.5pt}{9pt}\selectfont{SSIM$\uparrow$} & \fontsize{8.5pt}{9pt}\selectfont{LPIPS$\downarrow$} \\
\midrule
NSM    &18.80 &\underline{.7906} &   .2350 &15.92 &.6886 &   .2800 &16.50 &.7693 &   .2562 &18.78 &\textbf{.6874} &   .2897 &\underline{18.02} &\textbf{.6226} &   .2651 &19.51 &\underline{.6919} &   .2887  \\ \midrule
Bilinear  &17.47 &.7414 &   .2497 &15.75 &.6997 &   .2640 &17.40 &.7698 &   .2511 &19.03 &.6453 &   .2659 &17.35 &.6068 &   .2718 &18.74 &.6671 &   .2795 \\ \midrule
NN   &17.81 &.7609 &   .2507 &16.04 &.7006 &   .2685 &17.58 &\textbf{.7821} &   \underline{.2469} &\textbf{19.77} &\underline{.6847} &   .2810 &\textbf{18.42} &\underline{.6202} &   \textbf{.2565} &18.80 &.6909 &   .2931 \\ \midrule
Lanczos  &\underline{19.23} &.7866 &   \underline{.2206} &\underline{17.87} &\underline{.7470} &   \textbf{.2352} &17.62 &.7638 &   .2526 &18.69 &.6515 &   .2663 &17.05 &.6021 &   .2792 &\underline{19.74} &.6873 &   .2845 \\
\midrule
Bicubic    &16.37 &.7303 &   .2677 &15.58 &.7022 &   .2747 &\textbf{17.81} &\underline{.7744} &   \textbf{.2457} &\underline{19.25} &.6615 &   \textbf{.2471} &17.61 &.6192 &   \underline{.2578} &19.67 &.6788 &   .2851 \\ \midrule
LoftUp~\cite{huang2025loftup}   &16.44 &.7441 &   .2461 &15.44 &.7073 &   .2585 &17.50 &.7576 &   .2562 &19.17 &.6580 &   \underline{.2531} &17.31 &.6054 &   .2745 &19.31 &.6652 &   .2831 \\  \midrule
FeatUp~\cite{fu2024featup}  &\textbf{19.68} &\textbf{.7939} &   \textbf{.2080}  &16.40 &.7045 &   .2574 &17.31 &.7556 &   .2585 &18.68 &.6445 &   .2644 &17.51 &.6061 &   .2691 &18.90 &.6538 &   \underline{.2730} \\ \midrule
JAFAR~\cite{couairon2025jafar}    &17.15 &.7521 &   .2467 &16.26 &.7104 &   .2565 &16.59 &.7527 &   .2703 &18.24 &.6400 &   .2660 &17.05 &.6002 &   .2791 &19.38 &.6772 &   \textbf{.2646} \\ \midrule
AnyUp~\cite{wimmer2025anyup}    &18.91 &.7777 &   .2323 &\textbf{17.91} &\textbf{.7471} &   \underline{.2364} &\underline{17.66} &.7652 &   .2504 &18.76 &.6476 &   .2620 &17.03 &.6018 &   .2797 &\textbf{20.13} &\textbf{.6932} &   .2814 \\
\midrule
\multicolumn{1}{c|}{} & \multicolumn{9}{c|}{MVImgNet} & \multicolumn{9}{c}{T\&T} \\
\midrule
\multicolumn{1}{c|}{} & \multicolumn{3}{c|}{\textbf{G}eometry} & \multicolumn{3}{c|}{\textbf{T}exture} & \multicolumn{3}{c|}{\textbf{A}ll} & \multicolumn{3}{c|}{\textbf{G}eometry} & \multicolumn{3}{c|}{\textbf{T}exture} & \multicolumn{3}{c}{\textbf{A}ll} \\
\midrule
Upsamplers & \fontsize{8.5pt}{9pt}\selectfont{PSNR$\uparrow$} & \fontsize{8.5pt}{9pt}\selectfont{SSIM$\uparrow$} & \fontsize{8.5pt}{9pt}\selectfont{LPIPS$\downarrow$} & \fontsize{8.5pt}{9pt}\selectfont{PSNR$\uparrow$} & \fontsize{8.5pt}{9pt}\selectfont{SSIM$\uparrow$} & \fontsize{8.5pt}{9pt}\selectfont{LPIPS$\downarrow$} & \fontsize{8.5pt}{9pt}\selectfont{PSNR$\uparrow$} & \fontsize{8.5pt}{9pt}\selectfont{SSIM$\uparrow$} & \fontsize{8.5pt}{9pt}\selectfont{LPIPS$\downarrow$} & \fontsize{8.5pt}{9pt}\selectfont{PSNR$\uparrow$} & \fontsize{8.5pt}{9pt}\selectfont{SSIM$\uparrow$} & \fontsize{8.5pt}{9pt}\selectfont{LPIPS$\downarrow$} & \fontsize{8.5pt}{9pt}\selectfont{PSNR$\uparrow$} & \fontsize{8.5pt}{9pt}\selectfont{SSIM$\uparrow$} & \fontsize{8.5pt}{9pt}\selectfont{LPIPS$\downarrow$} & \fontsize{8.5pt}{9pt}\selectfont{PSNR$\uparrow$} & \fontsize{8.5pt}{9pt}\selectfont{SSIM$\uparrow$} & \fontsize{8.5pt}{9pt}\selectfont{LPIPS$\downarrow$} \\
\midrule
NSM   &18.12 &\textbf{.7663} &   .2133 &15.81 &.6773 &   .2280 &18.35 &.7593 &   .2156 &\textbf{20.08} &\textbf{.8544} &   \textbf{.1671} &\textbf{18.44} &\textbf{.7997} &   \textbf{.1844} &\textbf{20.71} &\textbf{.8614} &   \textbf{.1665} \\ \midrule
Bilinear  &18.94 &.7471 &   .1893 &16.81 &.7025 &   .2103 &\textbf{20.59} &\textbf{.7754} &   \textbf{.1894} &18.70 &.8155 &   .1948 &17.62 &.7830 &   .2161 &19.14 &.8315 &   .2005 \\ \midrule
NN   &18.56 &.7558 &   .2176 &16.99 &.6920 &   .2142 &18.71 &.7655 &   .2098 &18.61 &.8185 &   .2033 &\underline{17.74} &.7824 &   .2162 &\underline{20.46} &\underline{.8572} &   \underline{.1705} \\ \midrule
Lanczos   &18.47 &.7436 &   .1899 &\textbf{18.07} &\textbf{.7224} &   \textbf{.1925} &19.01 &.7647 &   .2109 &18.59 &.8170 &   \underline{.1837} &17.56 &.7842 &   .2164 &19.31 &.8372 &   .1815 \\
  \midrule
Bicubic  &18.94 &.7453 &   .1864 &16.93 &.7013 &   .2066 &17.29 &.7276 &   .2285 &18.06 &.8033 &   .2122 &17.33 &.7751 &   .2188 &18.20 &.8110 &   .2172 \\ \midrule
LoftUp~\cite{huang2025loftup}    &18.97 &.7471 &   .1854 &16.81 &.6999 &   .2088 &\underline{19.49} &.7659 &   .2020 &18.24 &.8052 &   .2068 &17.11 &.7760 &   .2143 &18.62 &.8239 &   .1991 \\  \midrule
FeatUp~\cite{fu2024featup}   &19.11 &.7416 &   .1897 &16.75 &.7084 &   .2040 &19.03 &.7589 &   \underline{.1985} & \underline{19.28} & \underline{.8246} &   .1867 &17.65 &\underline{.7875} &   .2133 &19.45 &.8313 &   .1881 \\ \midrule
JAFAR~\cite{couairon2025jafar}  &\textbf{20.25} &.7579 &   \textbf{.1708} &17.93 &.7118 &   .1963 &19.25 &.7591 &   .2006 &19.19 &.8234 &   .1963 &17.47 &.7868 &   \underline{.2054} &18.22 &.8149 &   .2133 \\ \midrule
AnyUp~\cite{wimmer2025anyup}    &\underline{20.22} &\underline{.7643} &   \underline{.1747} &\underline{18.04} &\underline{.7223} &   \underline{.1932} &19.47 &\underline{.7700} &   .2116 &18.57 &.8052 &   .2115 &17.71 &.7853 &   .2165 &18.51 &.8275 &   .1952 \\
\bottomrule
\end{tabular}
}

\end{table*}

\subsection{Correlation Heatmaps}
\begin{figure}[t]
\centering
\captionsetup[subfigure]{font=small, labelfont=bf}
% ===================== (a) =====================
\begin{subfigure}[t]{\linewidth}
    \centering
    \includegraphics[width=0.49\linewidth]{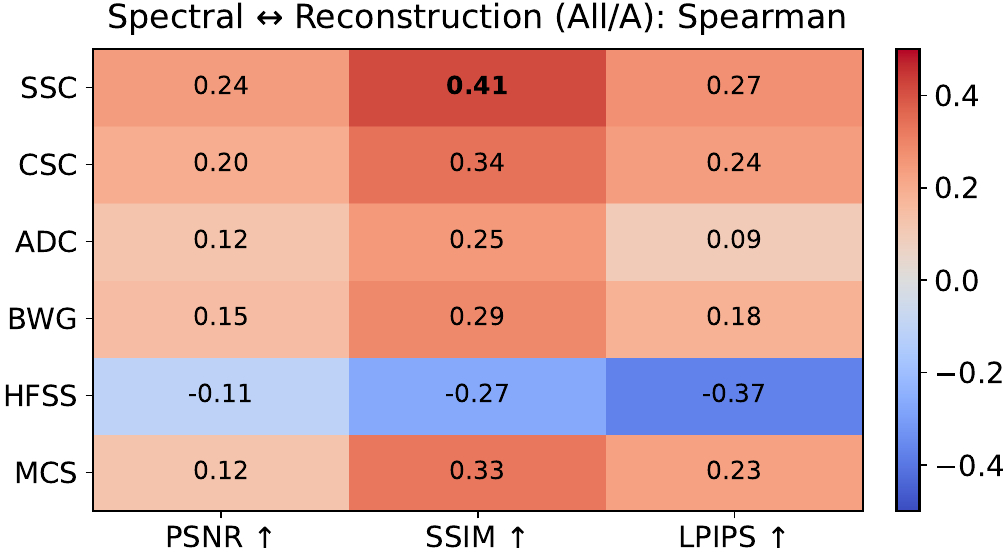}
    \hfill
   \includegraphics[width=0.49\linewidth]{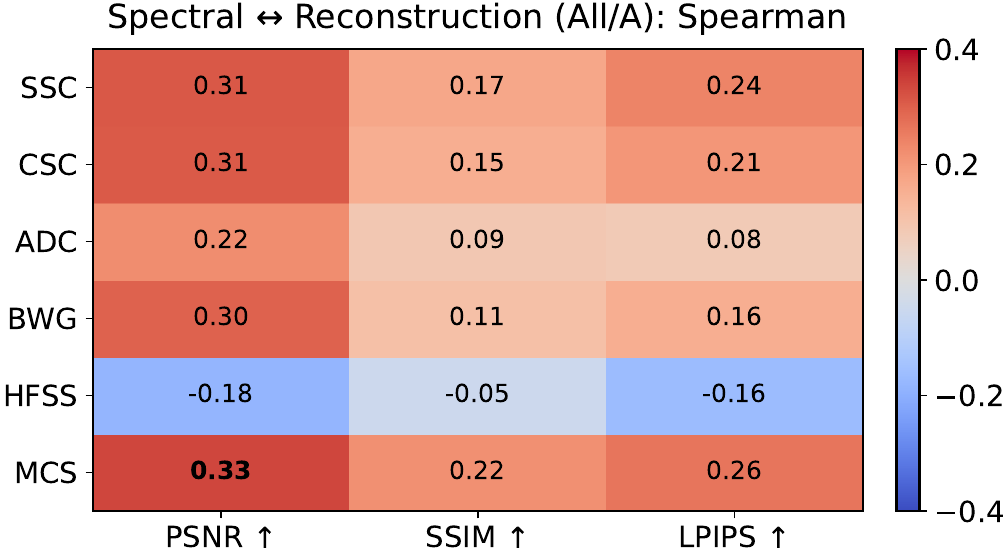}
    \vspace{-0.3em}
    \caption{Left: CLIP+Bilinear+DUSt3R~\cite{wang2024dust3r}, right: DINO+Bilinear+DUSt3R~\cite{wang2024dust3r}.}
\end{subfigure}
\vspace{0.6em}
% ===================== (b) =====================
\begin{subfigure}[t]{\linewidth}
    \centering
    \includegraphics[width=0.49\linewidth]{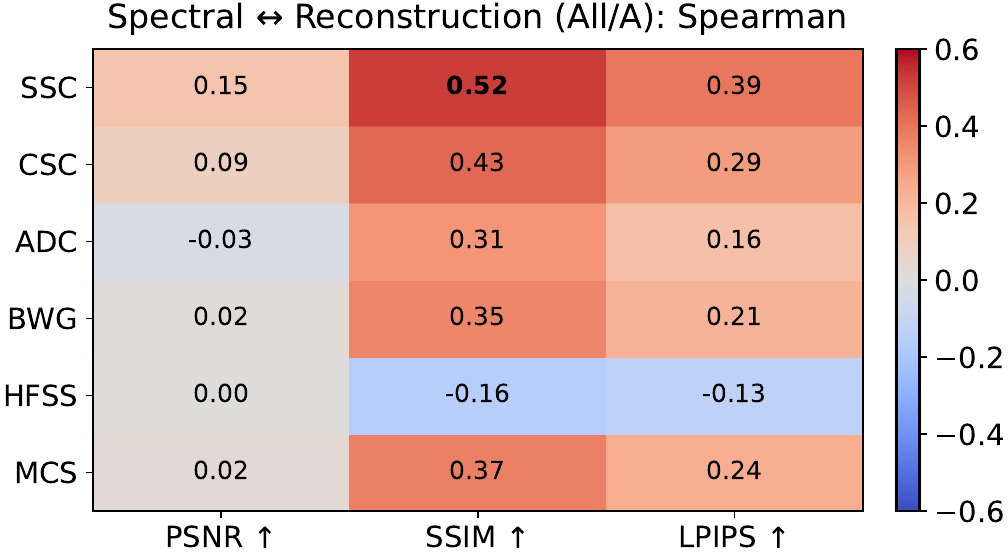}
    \hfill
    \includegraphics[width=0.49\linewidth]{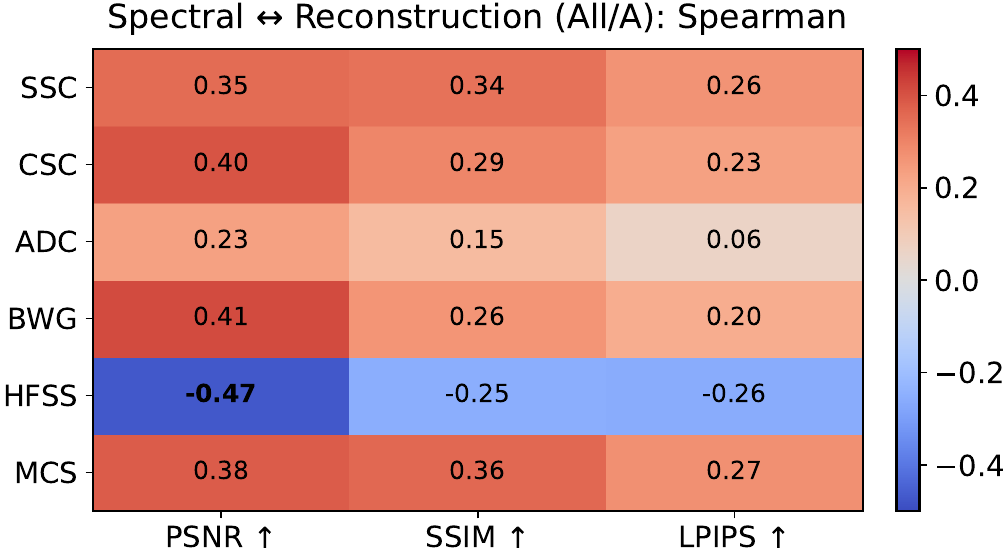}
    \vspace{-0.3em}
    \caption{Left: CLIP+Bilinear+MASt3R~\cite{leroy2024grounding}, right: DINO+Bilinear+MASt3R~\cite{leroy2024grounding}.}
\end{subfigure}
\vspace{0.6em}
\begin{subfigure}[t]{\linewidth}
    \centering
    \includegraphics[width=0.49\linewidth]{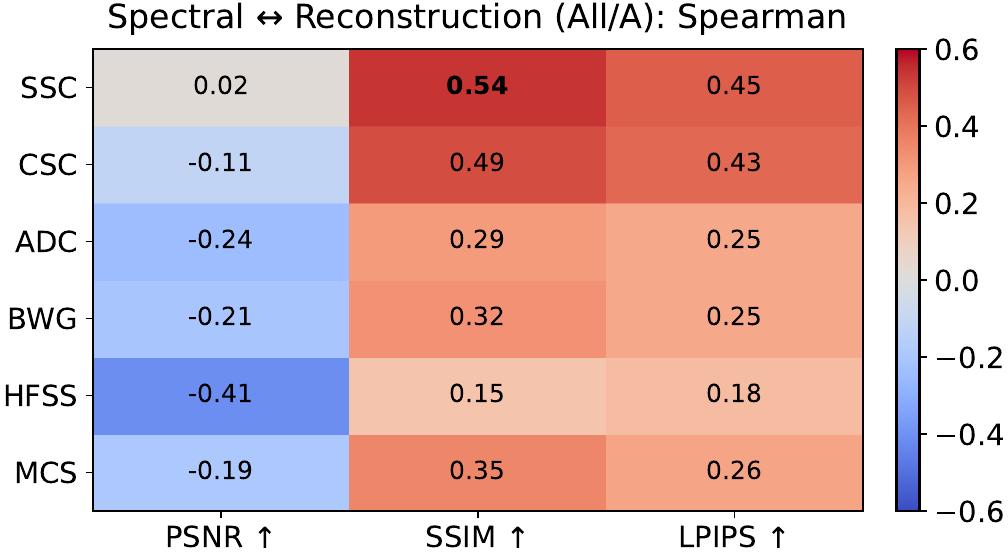}
    \hfill
    \includegraphics[width=0.49\linewidth]{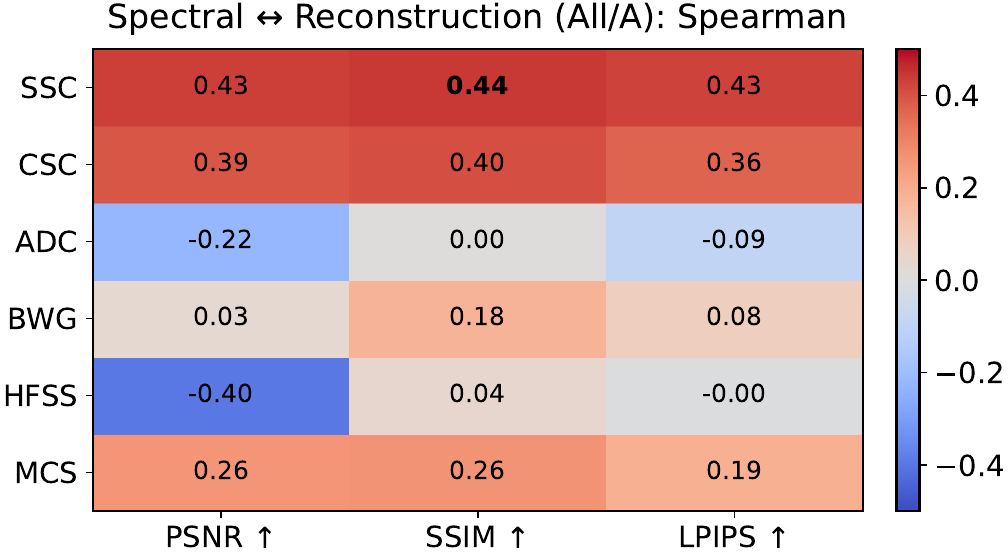}
    \vspace{-0.3em}
    \caption{Left: CLIP+AnyUp~\cite{wimmer2025anyup}+MASt3R~\cite{leroy2024grounding}, right: DINO+AnyUp~\cite{wimmer2025anyup}+MASt3R~\cite{leroy2024grounding}.}
\end{subfigure}
\vspace{0.6em}
\begin{subfigure}[t]{\linewidth}
  \centering
  \includegraphics[width=0.49\linewidth]{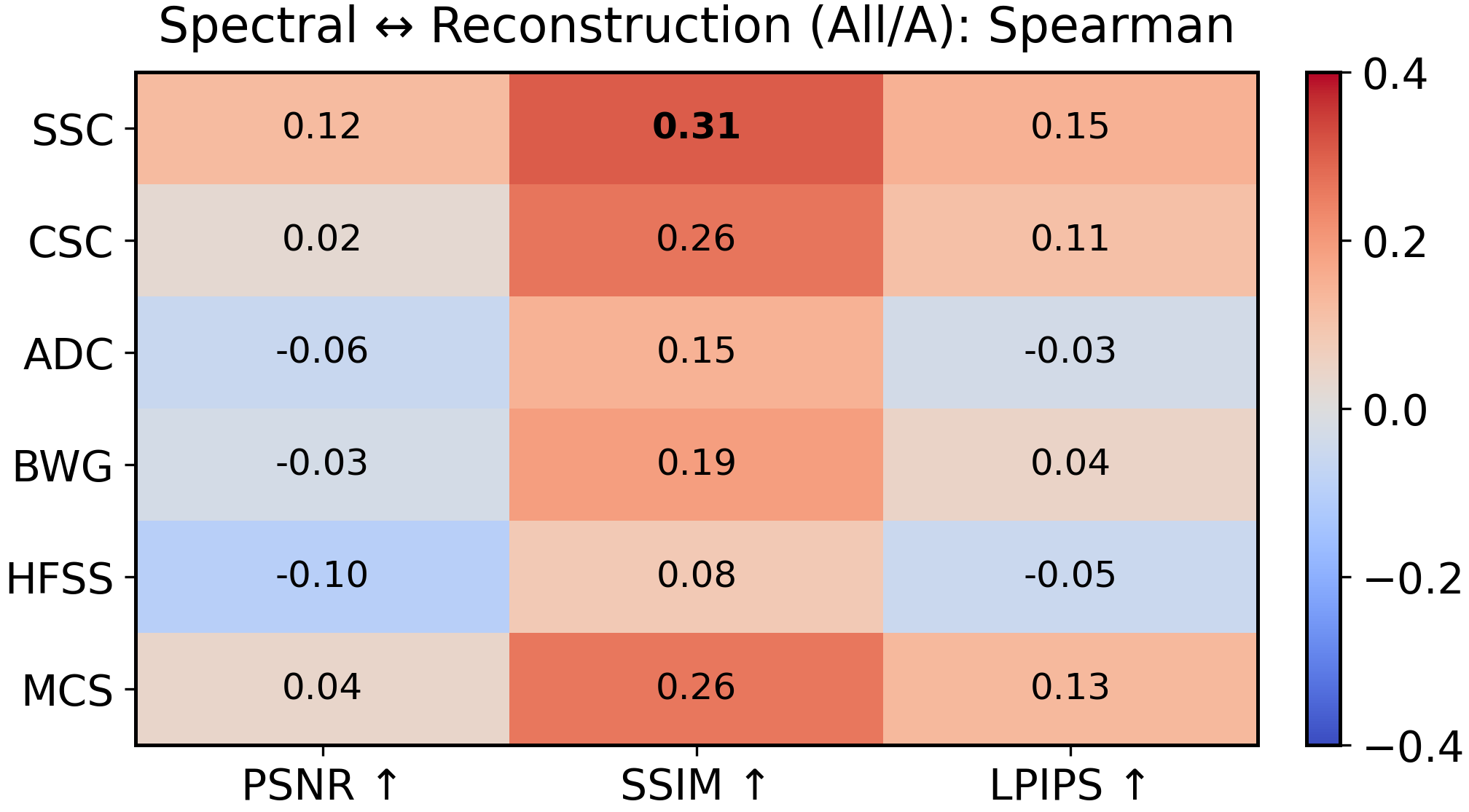}
  \hfill
  \includegraphics[width=0.49\linewidth]{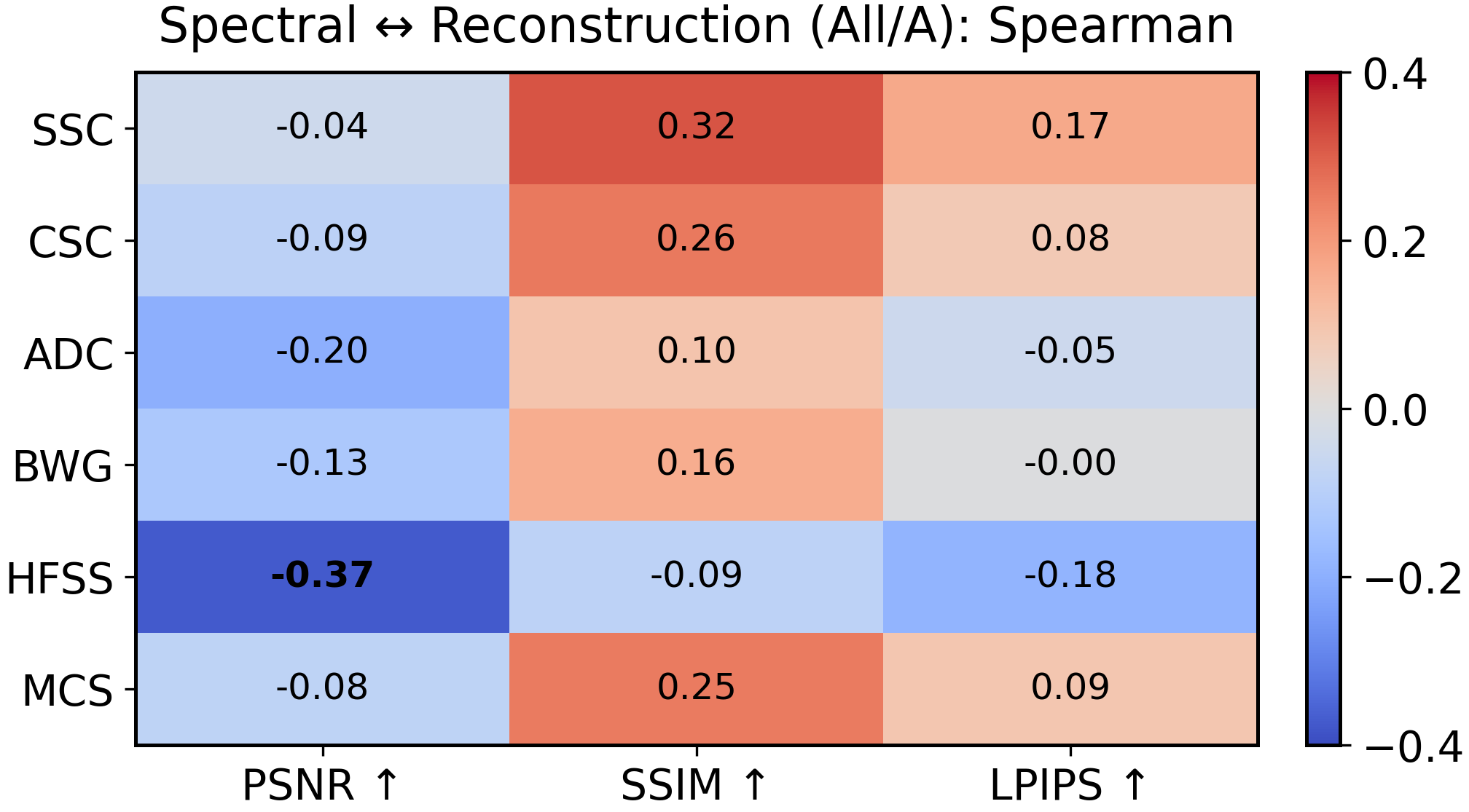}
  \vspace{-0.3em}
  \caption{Left: CLIP+FeatUp~\cite{fu2024featup}+DUSt3R~\cite{wang2024dust3r}, right: CLIP+FeatUp~\cite{fu2024featup}+MASt3R~\cite{leroy2024grounding}.}
\end{subfigure}

\caption{
\textbf{Spectral–reconstruction correlations.}
Each row shows the Spearman correlation heatmap.
}
\label{fig:heatmap-bilinear-nsm}
\end{figure}

\begin{figure}[t]
\centering
\captionsetup[subfigure]{font=small, labelfont=bf}
% ===================== (a) =====================
\begin{subfigure}[t]{\linewidth}
    \centering
    \includegraphics[width=0.49\linewidth]{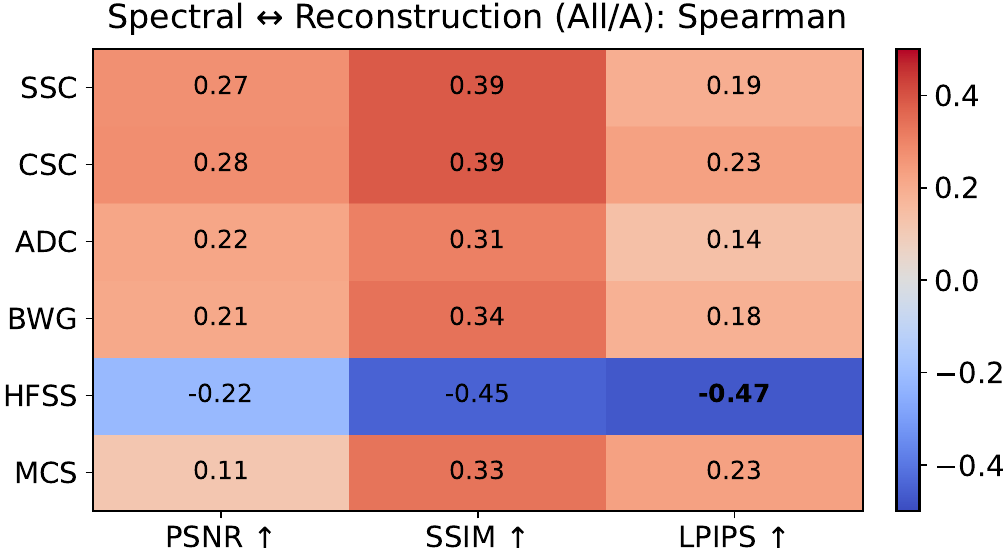}
    \hfill
   \includegraphics[width=0.49\linewidth]{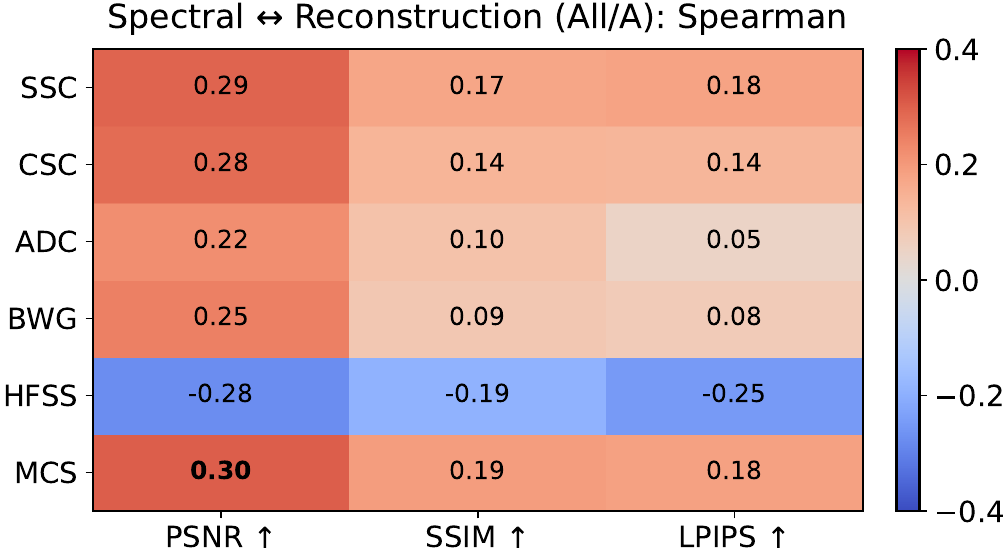}
    \vspace{-0.3em}
    \caption{Left: CLIP+Bicubic+DUSt3R~\cite{wang2024dust3r}, right: DINO+Bicubic+DUSt3R~\cite{wang2024dust3r}.}
\end{subfigure}
\vspace{0.6em}
% ===================== (b) =====================
\begin{subfigure}[t]{\linewidth}
    \centering
    \includegraphics[width=0.49\linewidth]{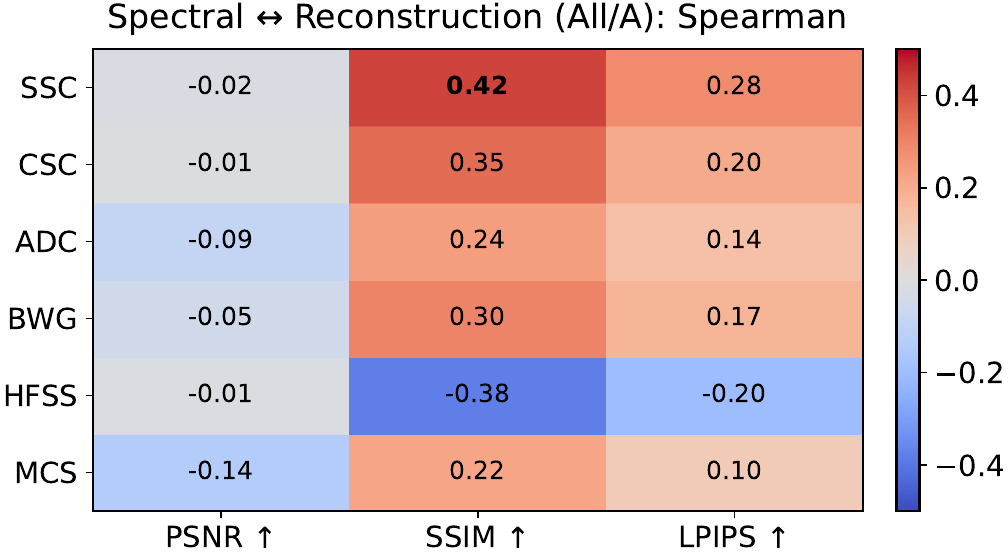}
    \hfill
    \includegraphics[width=0.49\linewidth]{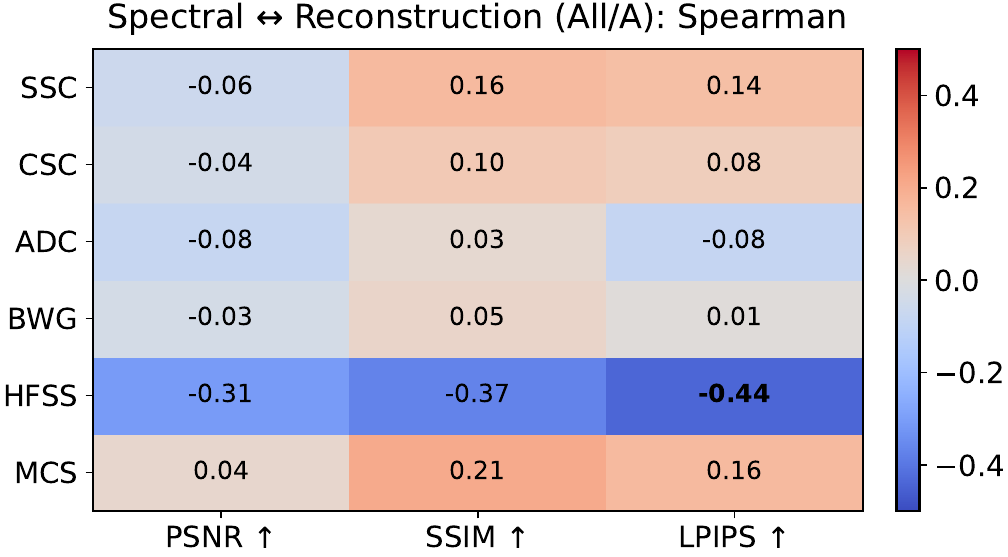}
    \vspace{-0.3em}
    \caption{Left: CLIP+Bicubic+MASt3R~\cite{leroy2024grounding}, right: DINO+Bicubic+MASt3R~\cite{leroy2024grounding}.}
\end{subfigure}

\vspace{0.6em}
\begin{subfigure}[t]{\linewidth}
  \centering
  \includegraphics[width=0.49\linewidth]{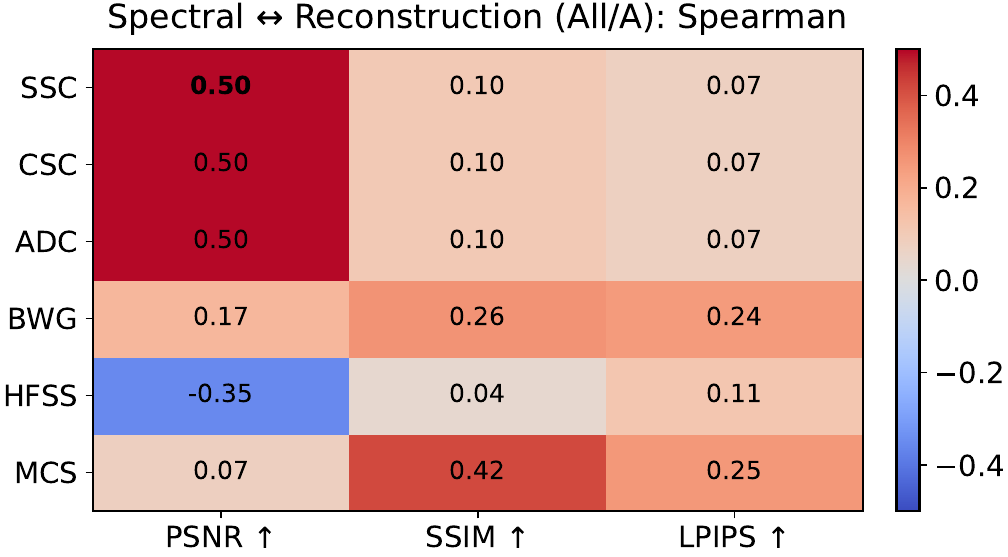}
  \hfill
  \includegraphics[width=0.49\linewidth]{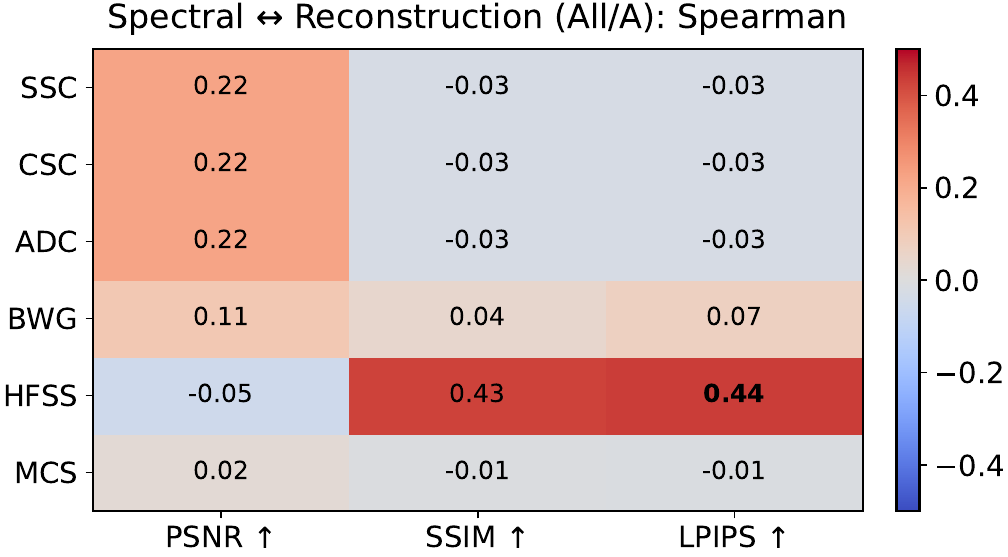}
  \vspace{-0.3em}
  \caption{Left: CLIP+NN+MASt3R~\cite{leroy2024grounding}, right: DINO+NN+MASt3R~\cite{leroy2024grounding}.}
\end{subfigure}
\caption{
\textbf{Spectral–reconstruction correlations.}
Each row shows the Spearman correlation heatmap.
}
\label{fig:heatmap-bicubic-nn}
\end{figure}

Additional Spearman correlation heatmaps under the joint (All, A) setting for other upsampling methods are shown in Figs.~\ref{fig:heatmap-bilinear-nsm}, and~\ref{fig:heatmap-bicubic-nn}. 
Each heatmap reports cross-scene correlations between spectral diagnostics (goodness-aligned) and NVS quality metrics (PSNR, SSIM, LPIPS), where positive values indicate that improved spectral behavior is associated with better reconstruction quality.

Across most interpolation and learned upsampling strategies, structural metrics (SSC and CSC) frequently exhibit positive correlations with reconstruction quality. 
In contrast, the high-frequency slope stability metric (HFSS) often shows negative correlations, indicating that stronger high-frequency spectral drift is associated with degraded reconstruction fidelity.

\subsection{AG and AT Correlations}
\begin{figure}[t]
\centering
\captionsetup[subfigure]{font=small, labelfont=bf}
% ===================== (a) =====================
\begin{subfigure}[t]{\linewidth}
  \centering
  \includegraphics[width=0.49\linewidth]{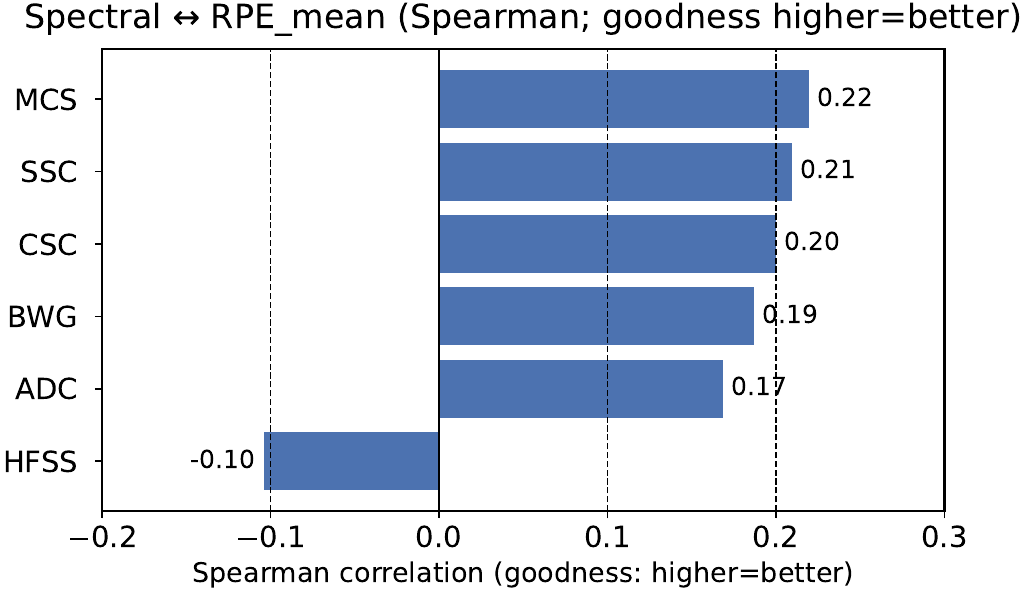}
  \hfill
  \includegraphics[width=0.49\linewidth]{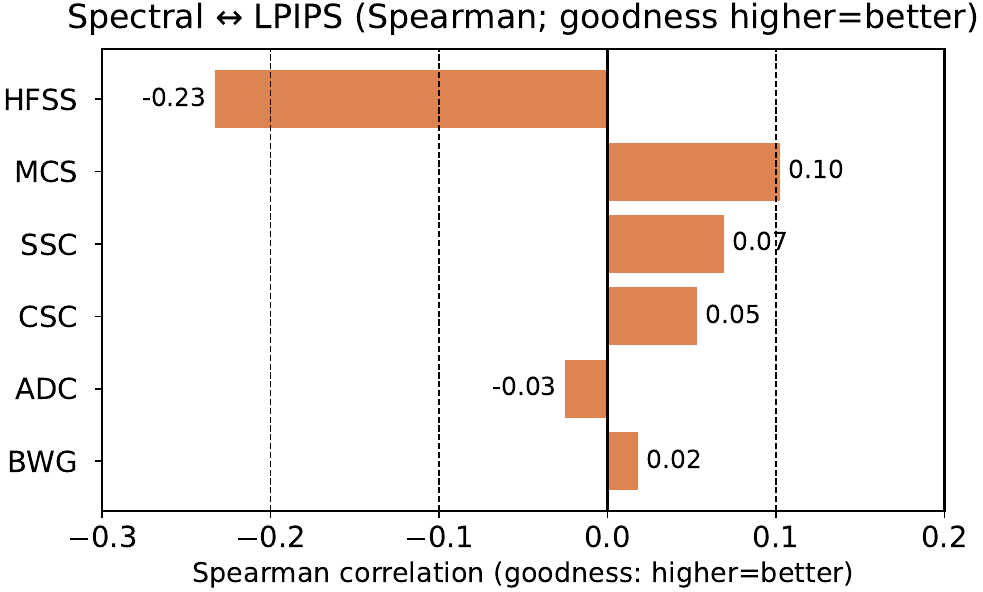}
  \vspace{-0.3em}
  \caption{Left (AG): Lanczos+DUSt3R~\cite{wang2024dust3r}, right (AT): Lanczos+DUSt3R~\cite{wang2024dust3r}.}
\end{subfigure}
\vspace{0.6em}
% ===================== (b) =====================
\begin{subfigure}[t]{\linewidth}
  \centering
  \includegraphics[width=0.49\linewidth]{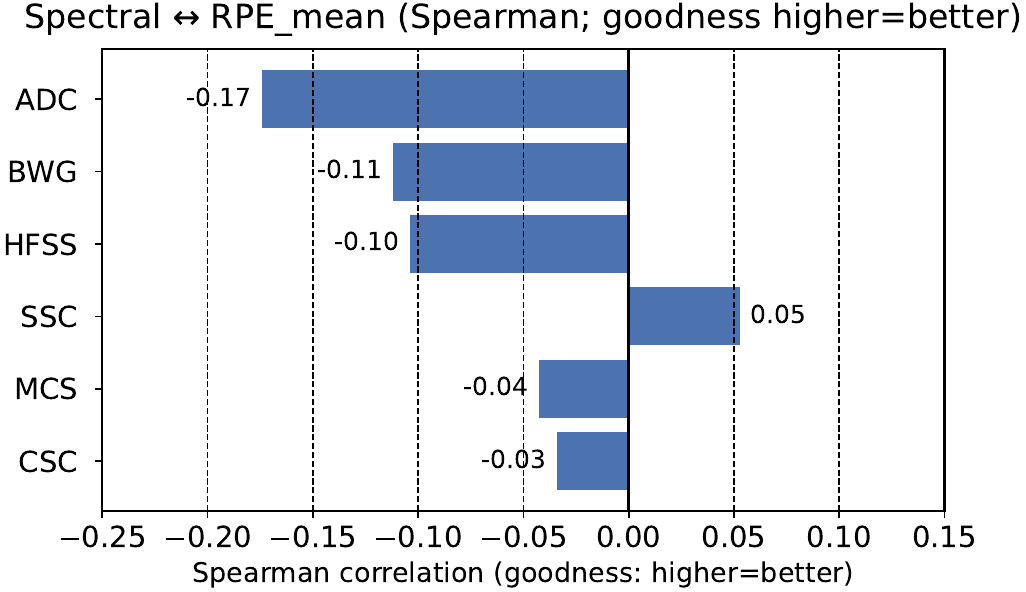}
  \hfill
  \includegraphics[width=0.49\linewidth]{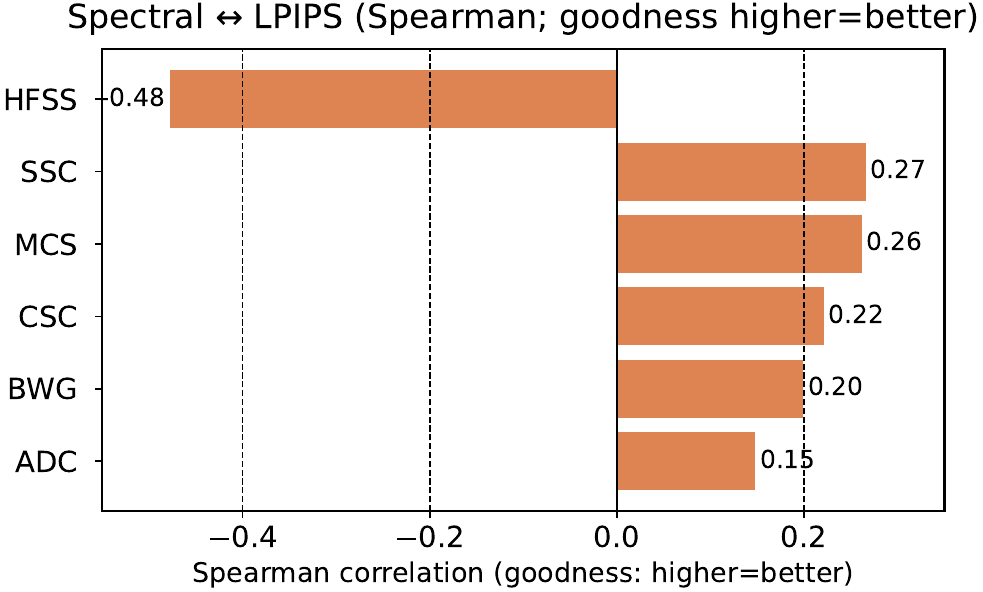}
  \vspace{-0.3em}
  \caption{Left (AG): Lanczos+MASt3R~\cite{leroy2024grounding}, right (AT): Lanczos+MASt3R~\cite{leroy2024grounding}.}
\end{subfigure}
\vspace{0.6em}
% ===================== (c) =====================
\begin{subfigure}[t]{\linewidth}
  \centering
  \includegraphics[width=0.49\linewidth]{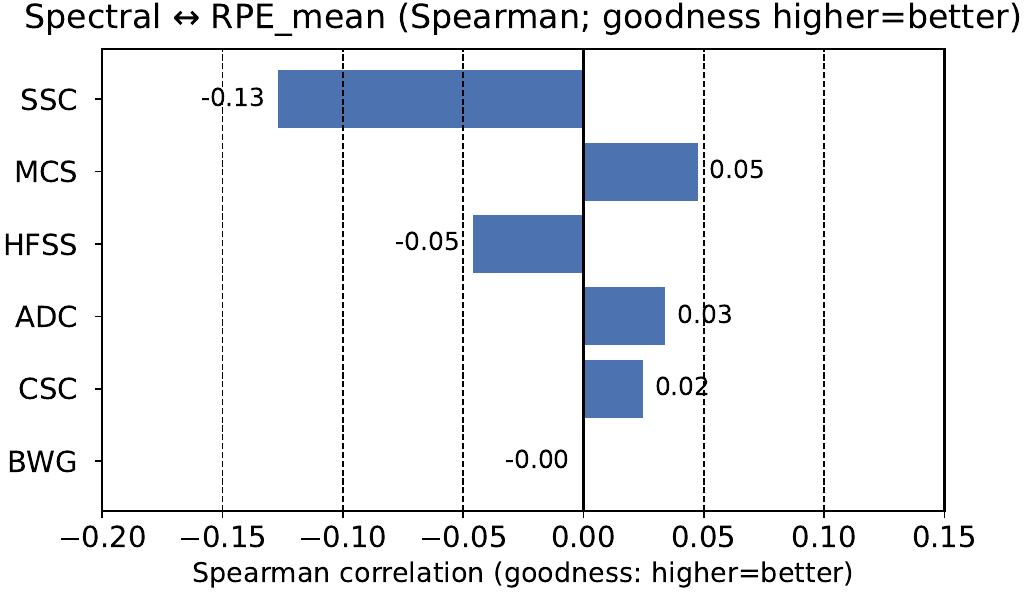}
  \hfill
  \includegraphics[width=0.49\linewidth]{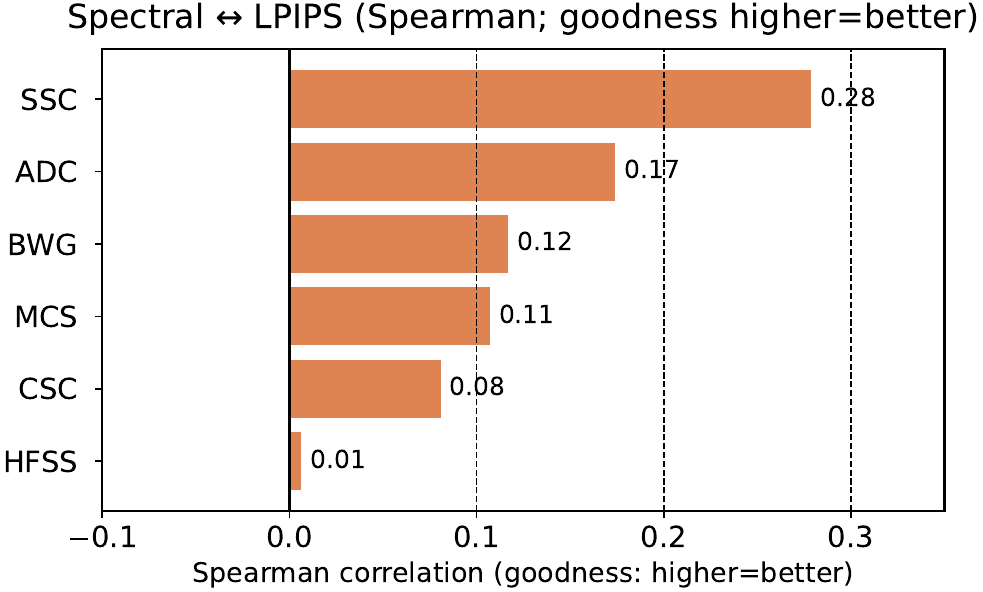}
  \vspace{-0.3em}
  \caption{Left (AG): JAFAR~\cite{couairon2025jafar}+DUSt3R~\cite{wang2024dust3r}, right (AT): JAFAR~\cite{couairon2025jafar}+DUSt3R~\cite{wang2024dust3r}.}
\end{subfigure}
\vspace{0.6em}
% ===================== (c) =====================
\begin{subfigure}[t]{\linewidth}
  \centering
  \includegraphics[width=0.49\linewidth]{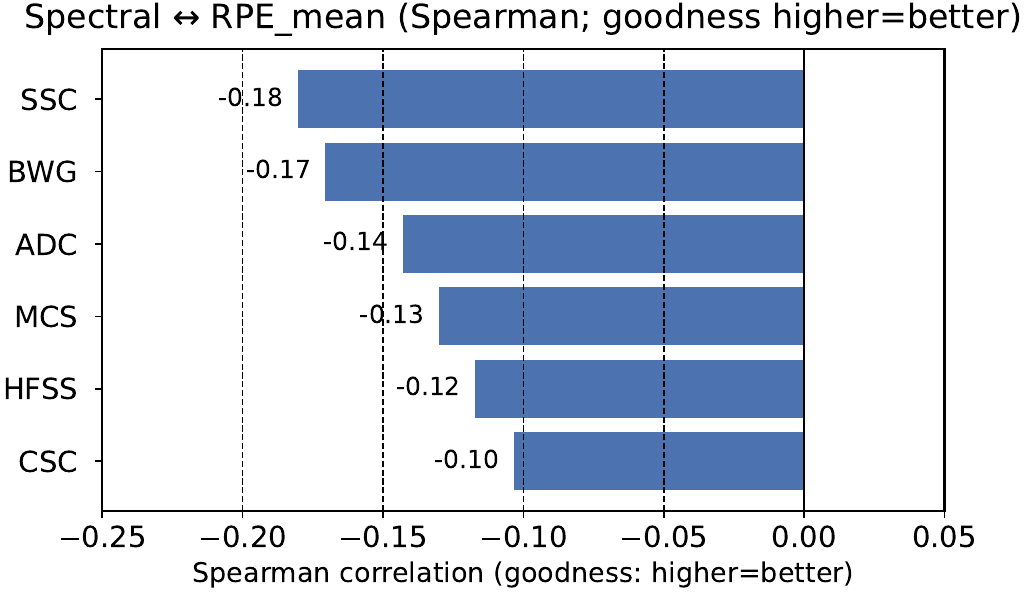}
  \hfill
  \includegraphics[width=0.49\linewidth]{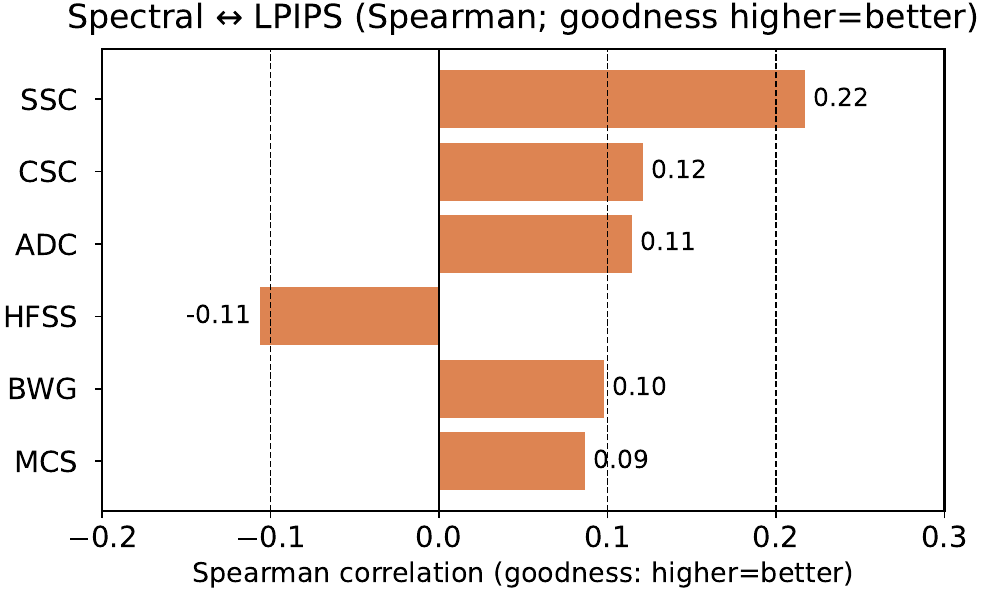}
  \vspace{-0.3em}
  \caption{Left (AG): JAFAR~\cite{couairon2025jafar}+MASt3R~\cite{leroy2024grounding}, right (AT): JAFAR~\cite{couairon2025jafar}+MASt3R~\cite{leroy2024grounding}.}
\end{subfigure}
\caption{
\textbf{Spearman correlations between spectral diagnostics and reconstruction quality under geometry-only (AG) and texture-only (AT) probing settings (DINO). Left panels show correlations with $-\mathrm{RPE}_{mean}$ (geometry quality), and right panels show correlations with $-\mathrm{LPIPS}$ (texture quality).}
}
\label{fig:ag-at-lanczos-jafar}
\end{figure}

\begin{figure}[t]
\centering
\captionsetup[subfigure]{font=small, labelfont=bf}
% ===================== (a) =====================
\begin{subfigure}[t]{\linewidth}
  \centering
  \includegraphics[width=0.49\linewidth]{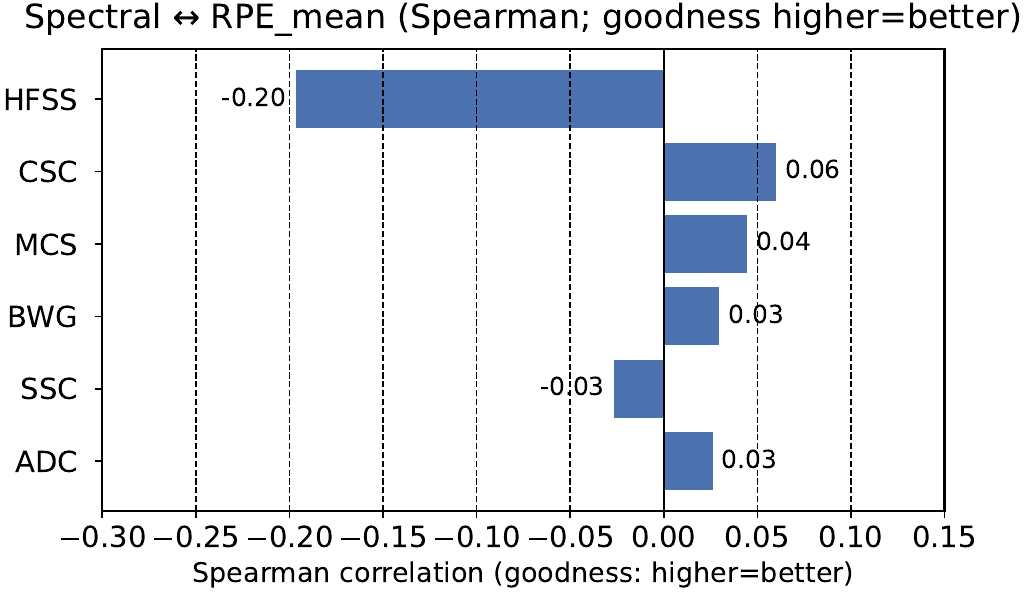}
  \hfill
  \includegraphics[width=0.49\linewidth]{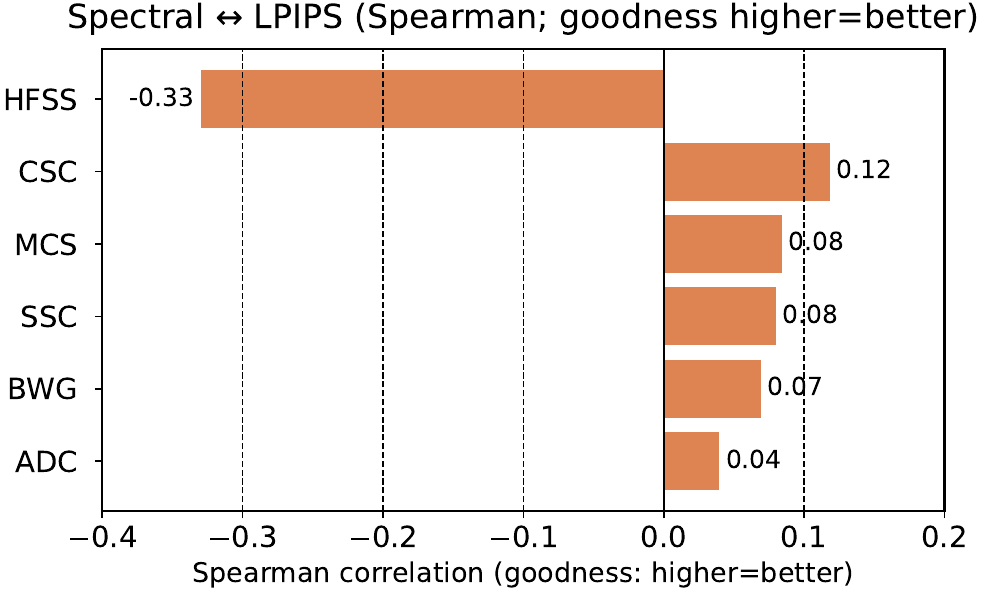}
  \vspace{-0.3em}
  \caption{Left (AG): CLIP+Bicubic+DUSt3R~\cite{wang2024dust3r}, right (AT): CLIP+Bicubic+DUSt3R~\cite{wang2024dust3r}.}
\end{subfigure}
\vspace{0.6em}
% ===================== (b) =====================
\begin{subfigure}[t]{\linewidth}
  \centering
  \includegraphics[width=0.49\linewidth]{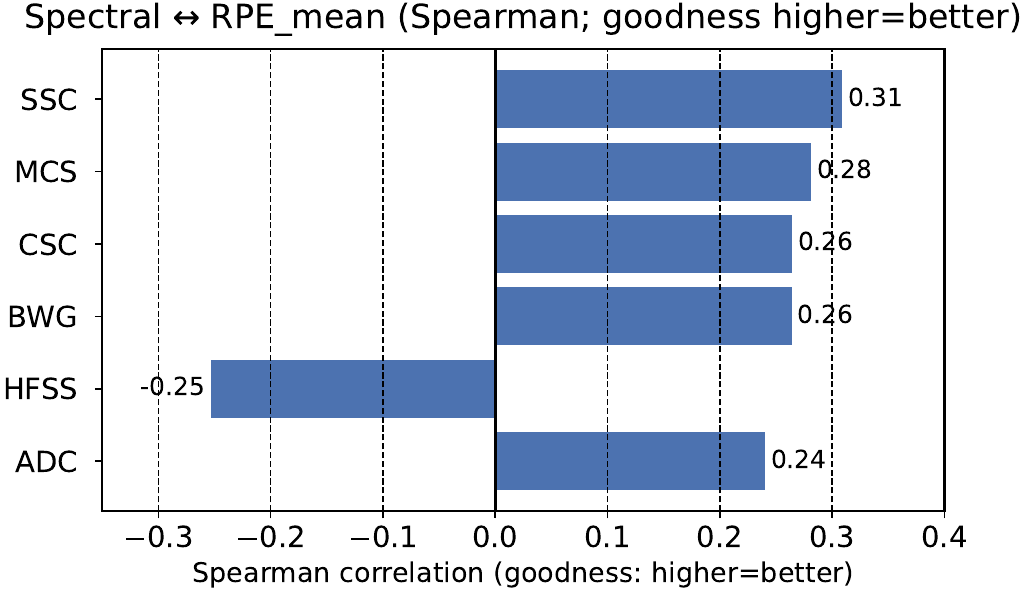}
  \hfill
  \includegraphics[width=0.49\linewidth]{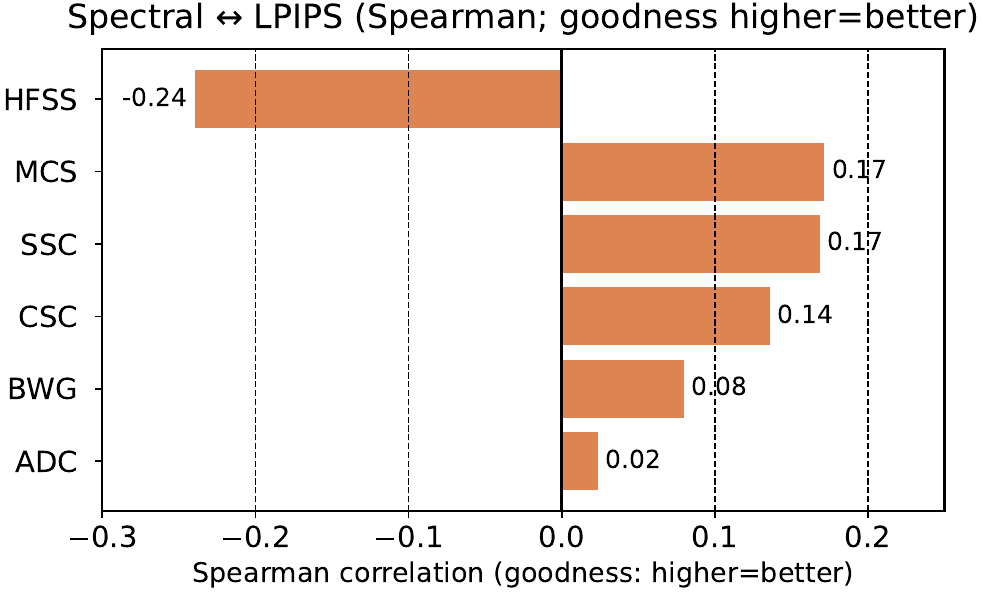}
  \vspace{-0.3em}
  \caption{Left (AG): DINO+Bicubic+DUSt3R~\cite{wang2024dust3r}, right (AT): DINO+Bicubic+DUSt3R~\cite{wang2024dust3r}.}
\end{subfigure}
\vspace{0.6em}
% ===================== (c) =====================
\begin{subfigure}[t]{\linewidth}
  \centering
  \includegraphics[width=0.49\linewidth]{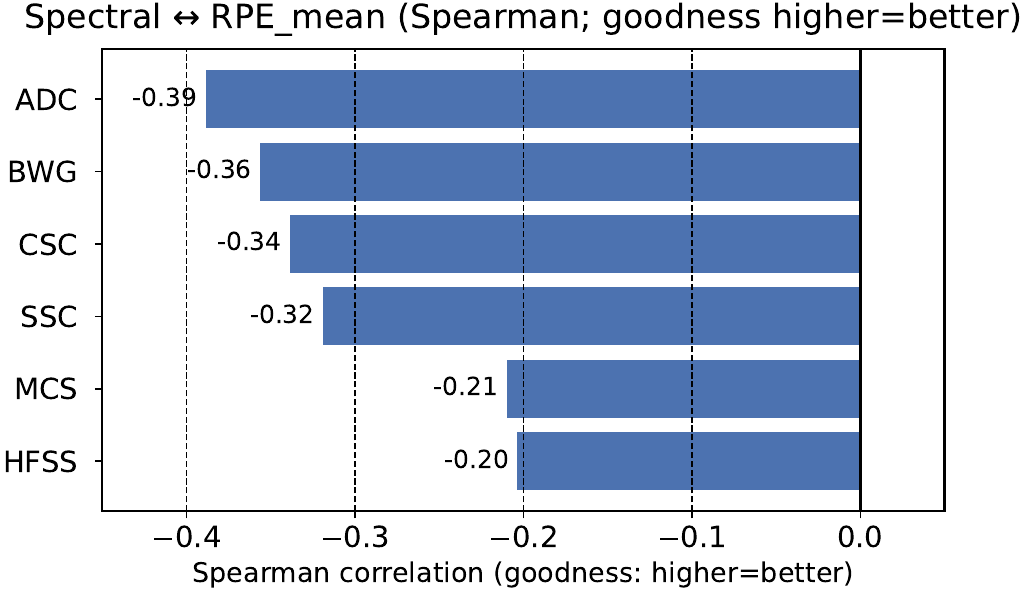}
  \hfill
  \includegraphics[width=0.49\linewidth]{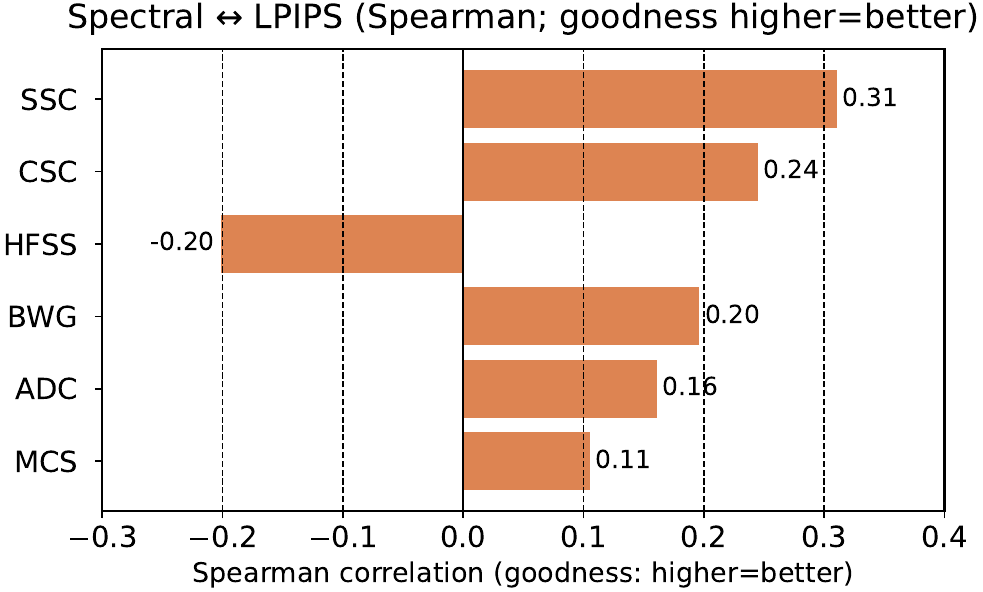}
  \vspace{-0.3em}
  \caption{Left (AG): CLIP+Bicubic+MASt3R~\cite{leroy2024grounding}, right (AT): CLIP+Bicubic+MASt3R~\cite{leroy2024grounding}.}
\end{subfigure}
\vspace{0.6em}
% ===================== (c) =====================
\begin{subfigure}[t]{\linewidth}
  \centering
  \includegraphics[width=0.49\linewidth]{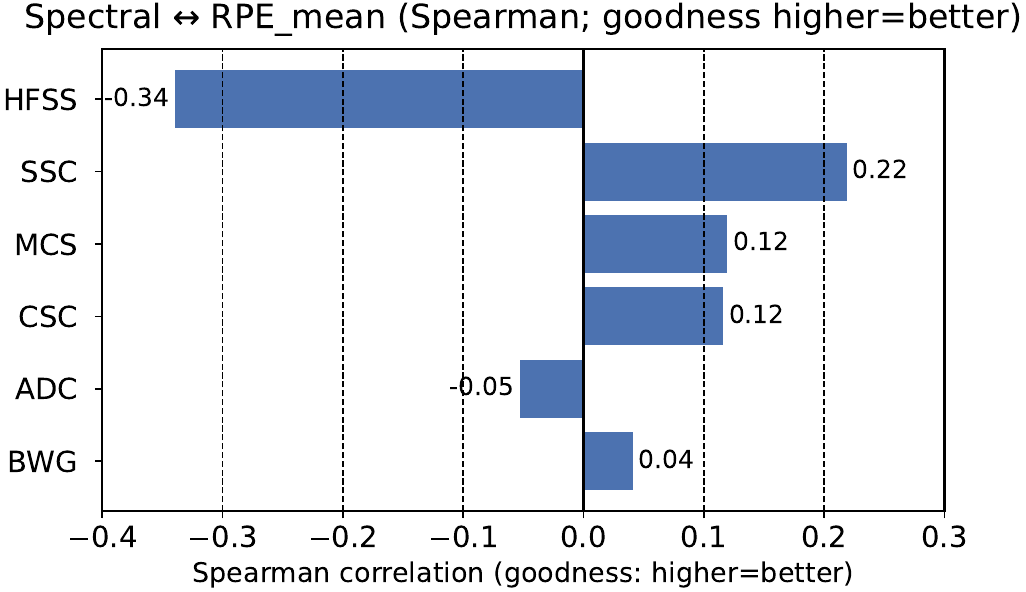}
  \hfill
  \includegraphics[width=0.49\linewidth]{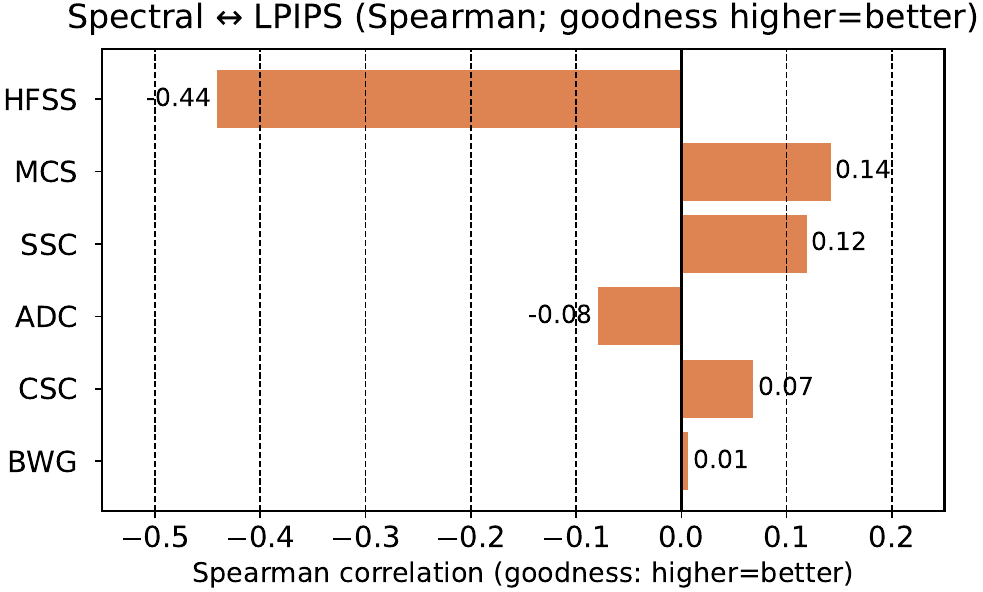}
  \vspace{-0.3em}
  \caption{Left (AG): DINO+Bicubic+MASt3R~\cite{leroy2024grounding}, right (AT): DINO+Bicubic+MASt3R~\cite{leroy2024grounding}.}
\end{subfigure}
\caption{
\textbf{Spearman correlations between spectral diagnostics and reconstruction quality under geometry-only (AG) and texture-only (AT) probing settings. Left panels show correlations with $-\mathrm{RPE}_{mean}$ (geometry quality), and right panels show correlations with $-\mathrm{LPIPS}$ (texture quality).}
}
\label{fig:ag-at-bicubic}
\end{figure}

\begin{figure}[t]
\centering
\captionsetup[subfigure]{font=small, labelfont=bf}
% ===================== (a) =====================
\begin{subfigure}[t]{\linewidth}
  \centering
  \includegraphics[width=0.49\linewidth]{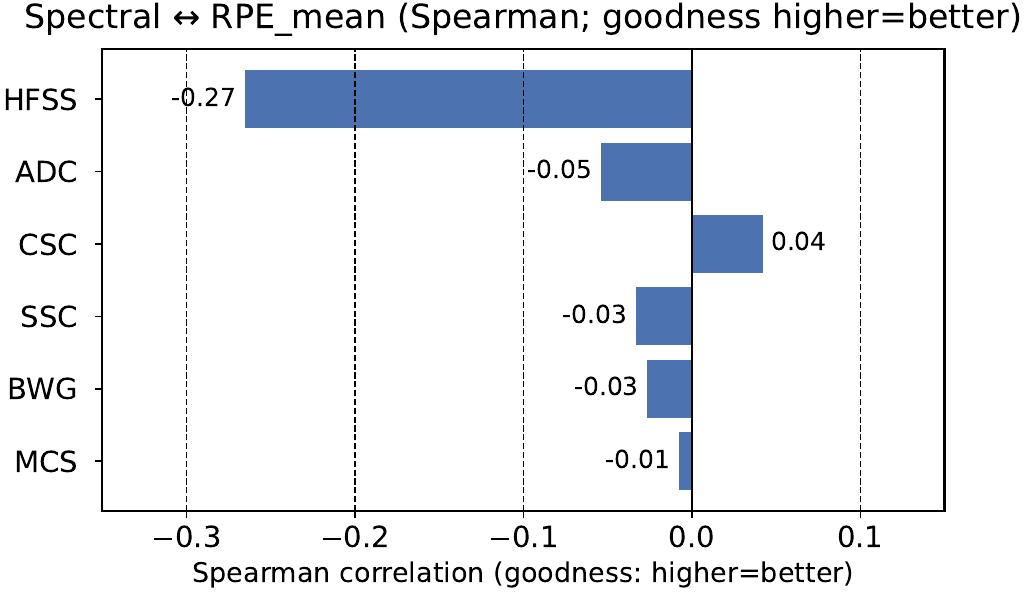}
  \hfill
  \includegraphics[width=0.49\linewidth]{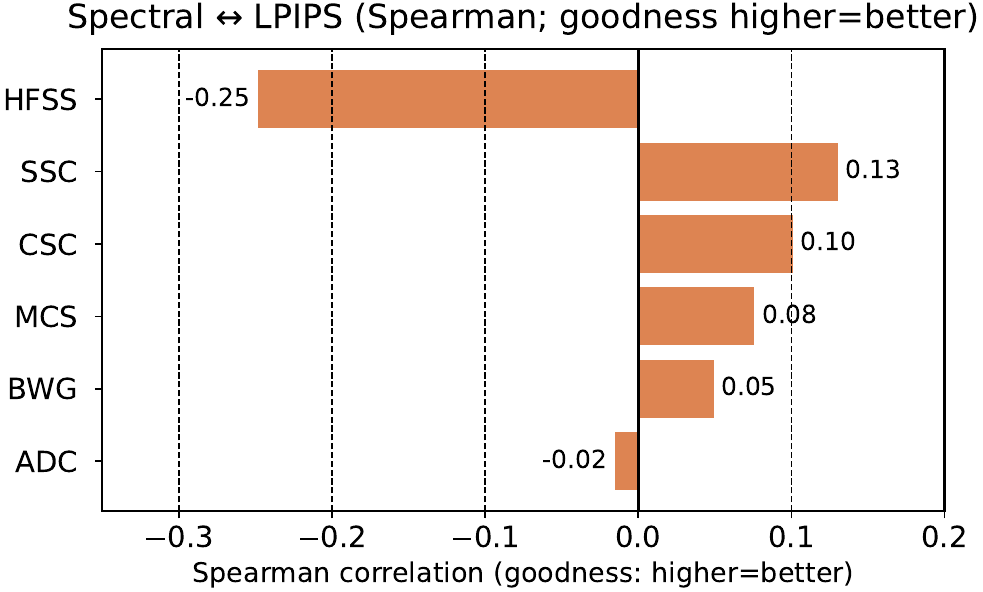}
  \vspace{-0.3em}
  \caption{Left (AG): CLIP+Bilinear+DUSt3R~\cite{wang2024dust3r}, right (AT): CLIP+Bilinear+DUSt3R~\cite{wang2024dust3r}.}
\end{subfigure}
\vspace{0.6em}
% ===================== (b) =====================
\begin{subfigure}[t]{\linewidth}
  \centering
  \includegraphics[width=0.49\linewidth]{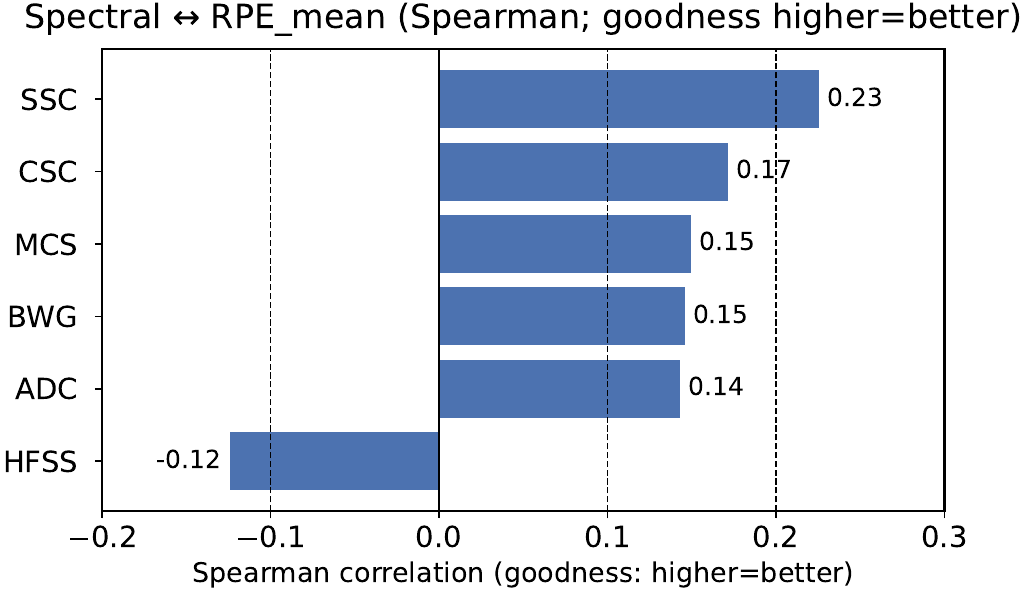}
  \hfill
  \includegraphics[width=0.49\linewidth]{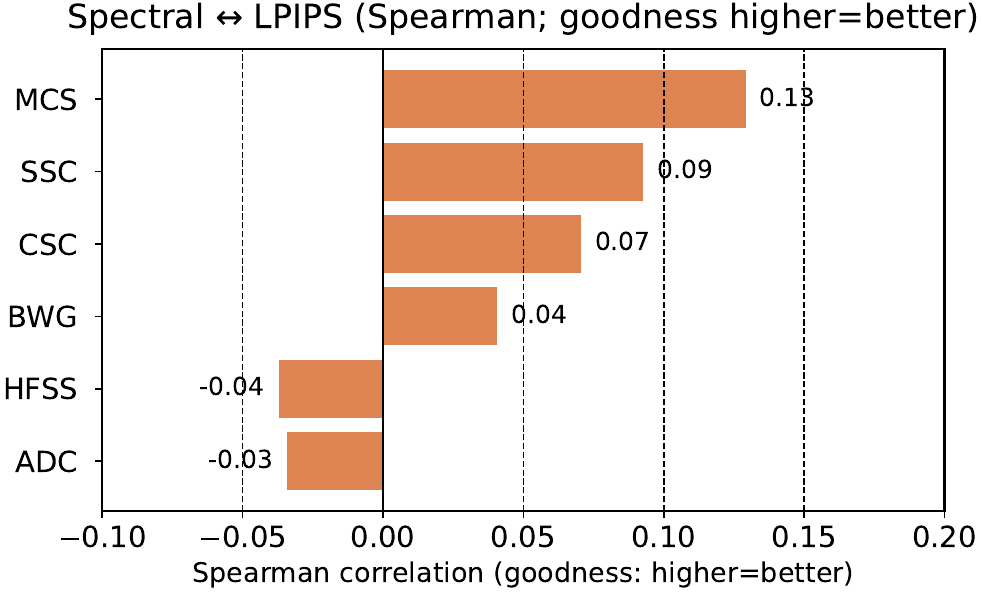}
  \vspace{-0.3em}
  \caption{Left (AG): DINO+Bilinear+DUSt3R~\cite{wang2024dust3r}, right (AT): DINO+Bilinear+DUSt3R~\cite{wang2024dust3r}.}
\end{subfigure}
\vspace{0.6em}
% ===================== (c) =====================
\begin{subfigure}[t]{\linewidth}
  \centering
  \includegraphics[width=0.49\linewidth]{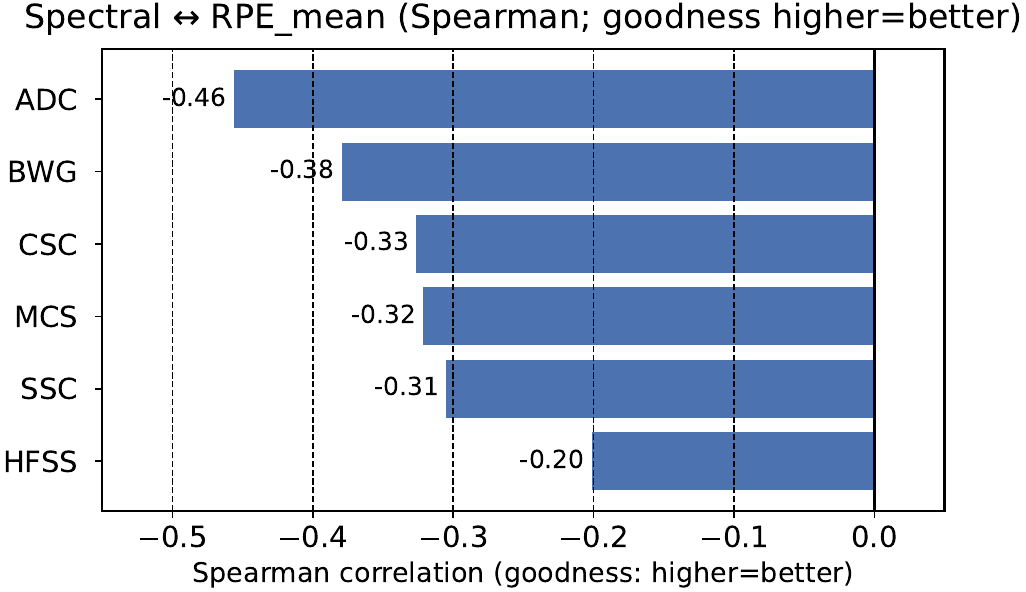}
  \hfill
  \includegraphics[width=0.49\linewidth]{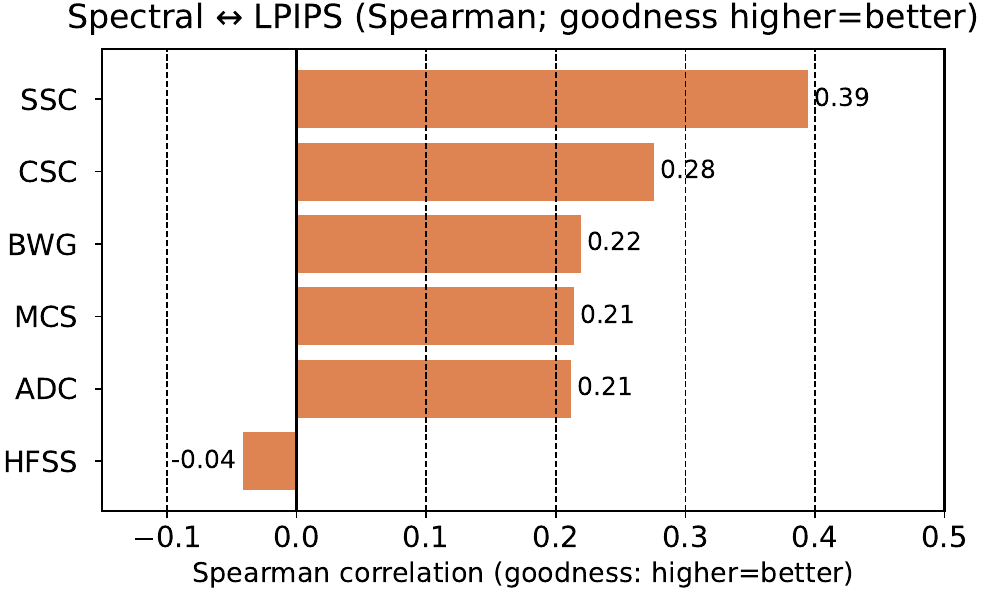}
  \vspace{-0.3em}
  \caption{Left (AG): CLIP+Bilinear+MASt3R~\cite{leroy2024grounding}, right (AT): CLIP+Bilinear+MASt3R~\cite{leroy2024grounding}.}
\end{subfigure}
\vspace{0.6em}
% ===================== (c) =====================
\begin{subfigure}[t]{\linewidth}
  \centering
  \includegraphics[width=0.49\linewidth]{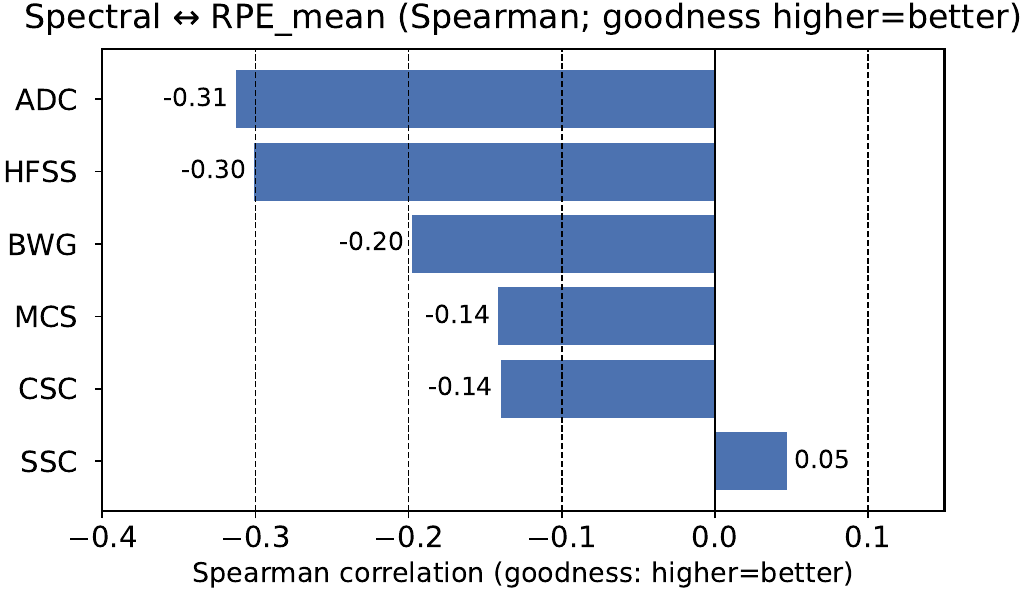}
  \hfill
  \includegraphics[width=0.49\linewidth]{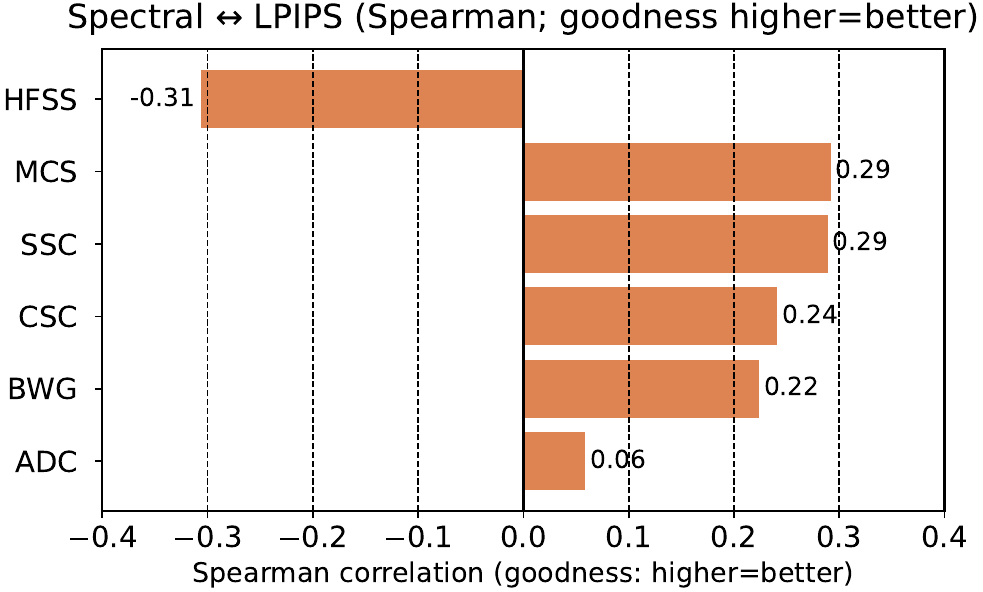}
  \vspace{-0.3em}
  \caption{Left (AG): DINO+Bilinear+MASt3R~\cite{leroy2024grounding}, right (AT): DINO+Bilinear+MASt3R~\cite{leroy2024grounding}.}
\end{subfigure}
\caption{
\textbf{Spearman correlations between spectral diagnostics and reconstruction quality under geometry-only (AG) and texture-only (AT) probing settings. Left panels show correlations with $-\mathrm{RPE}_{mean}$ (geometry quality), and right panels show correlations with $-\mathrm{LPIPS}$ (texture quality).}
}
\label{fig:ag-at-bilinear}
\end{figure}

\begin{figure}[t]
\centering
\captionsetup[subfigure]{font=small, labelfont=bf}

% ===================== (b) =====================
\begin{subfigure}[t]{\linewidth}
  \centering
  \includegraphics[width=0.49\linewidth]{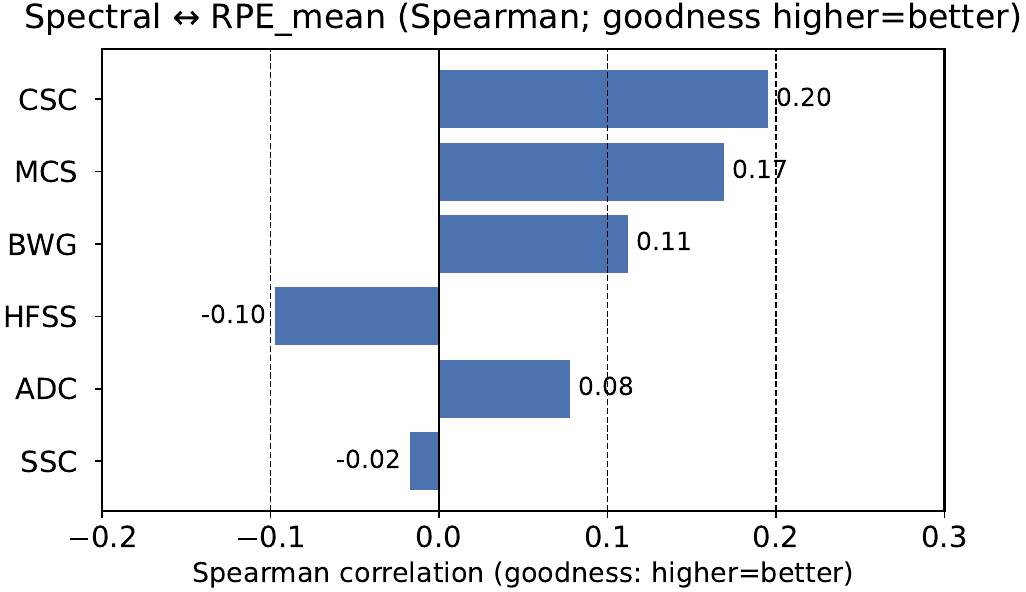}
  \hfill
  \includegraphics[width=0.49\linewidth]{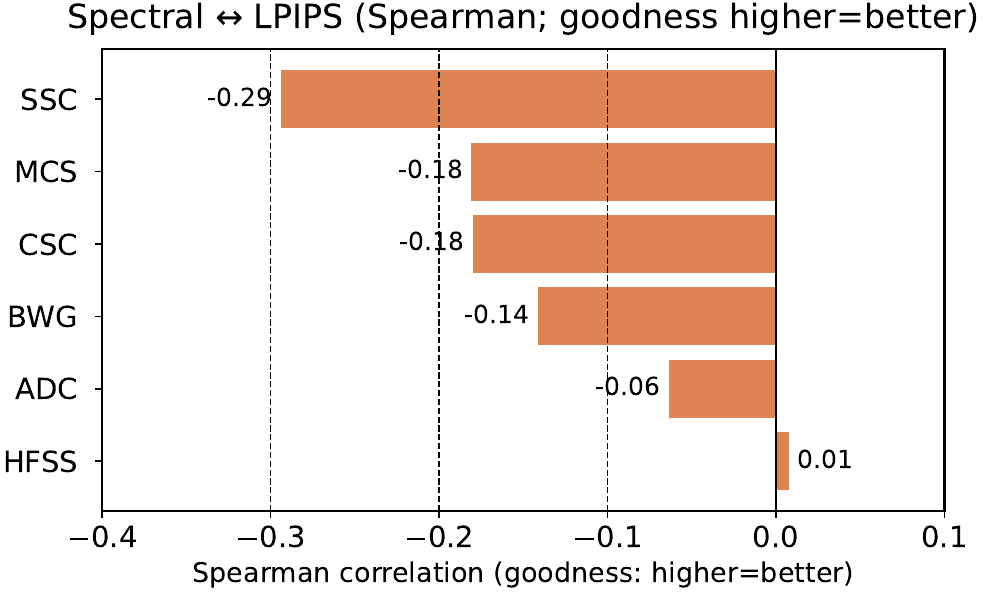}
  \vspace{-0.3em}
  \caption{Left (AG): DINO+NSM+DUSt3R~\cite{wang2024dust3r}, right (AT): DINO+NSM+DUSt3R~\cite{wang2024dust3r}.}
\end{subfigure}
\vspace{0.6em}
% ===================== (c) =====================

% ===================== (c) =====================
\begin{subfigure}[t]{\linewidth}
  \centering
  \includegraphics[width=0.49\linewidth]{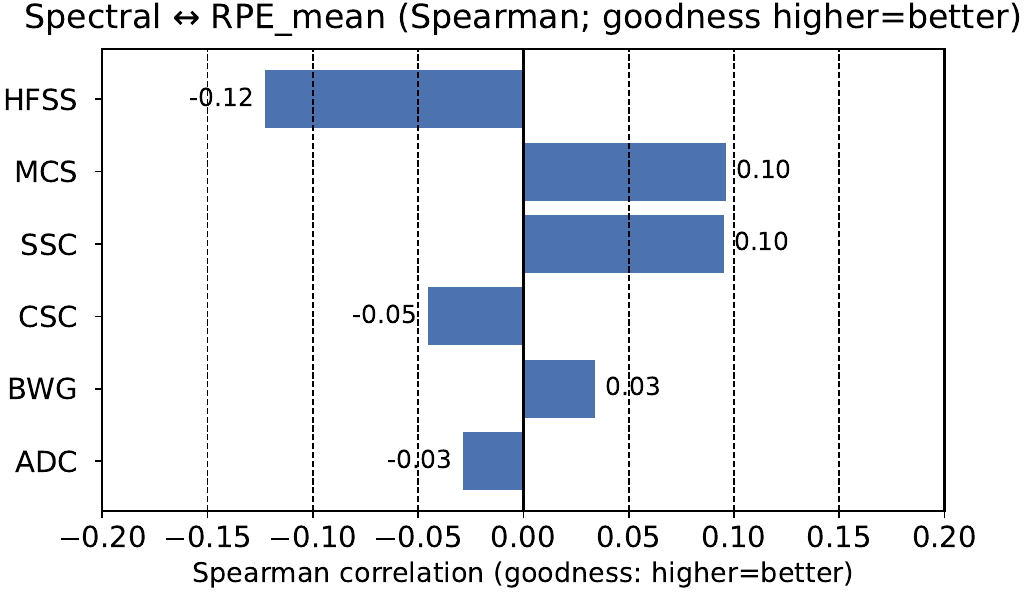}
  \hfill
  \includegraphics[width=0.49\linewidth]{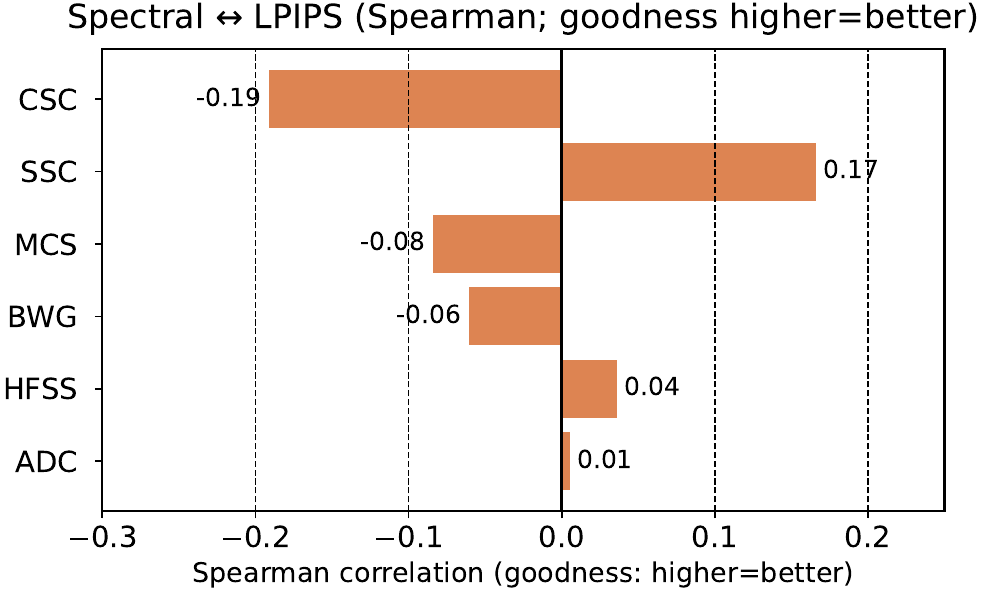}
  \vspace{-0.3em}
  \caption{Left (AG): DINO+NSM+MASt3R~\cite{leroy2024grounding}, right (AT): DINO+NSM+MASt3R~\cite{leroy2024grounding}.}
\end{subfigure}
\vspace{0.6em}
\begin{subfigure}[t]{\linewidth}
  \centering
  \includegraphics[width=0.49\linewidth]{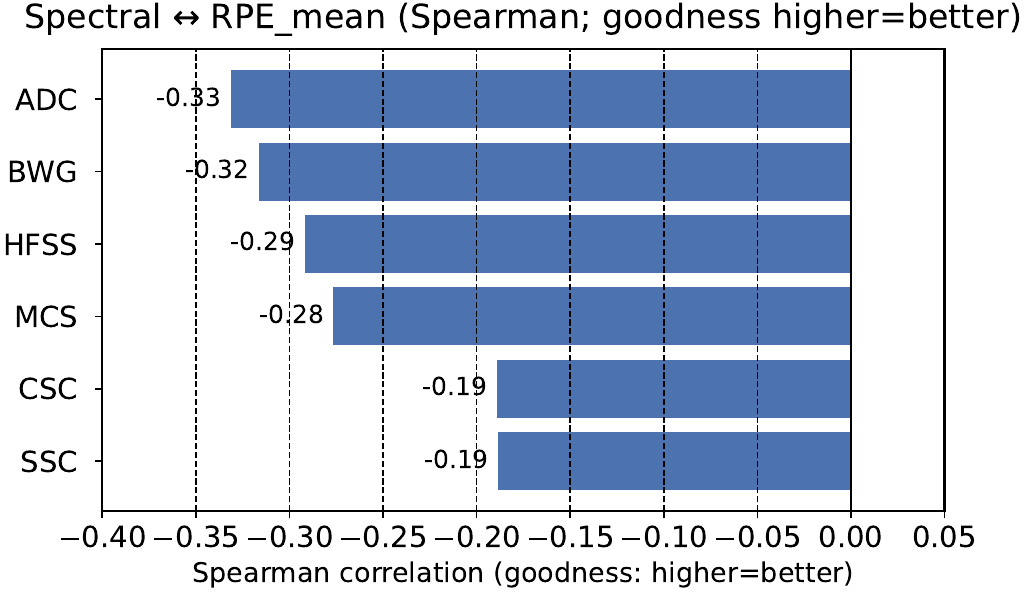}
  \hfill
  \includegraphics[width=0.49\linewidth]{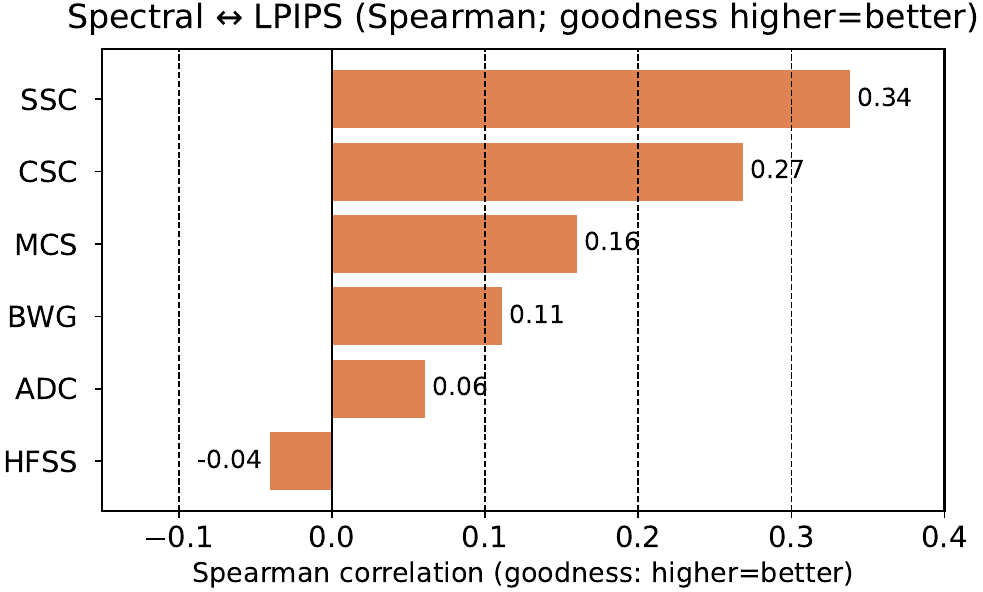}
  \vspace{-0.3em}
  \caption{Left (AG): CLIP+AnyUp~\cite{wimmer2025anyup}+MASt3R~\cite{leroy2024grounding}, right (AT): CLIP+AnyUp~\cite{wimmer2025anyup}+MASt3R~\cite{leroy2024grounding}.}
\end{subfigure}
\vspace{0.6em}
% ===================== (c) =====================
\begin{subfigure}[t]{\linewidth}
  \centering
  \includegraphics[width=0.49\linewidth]{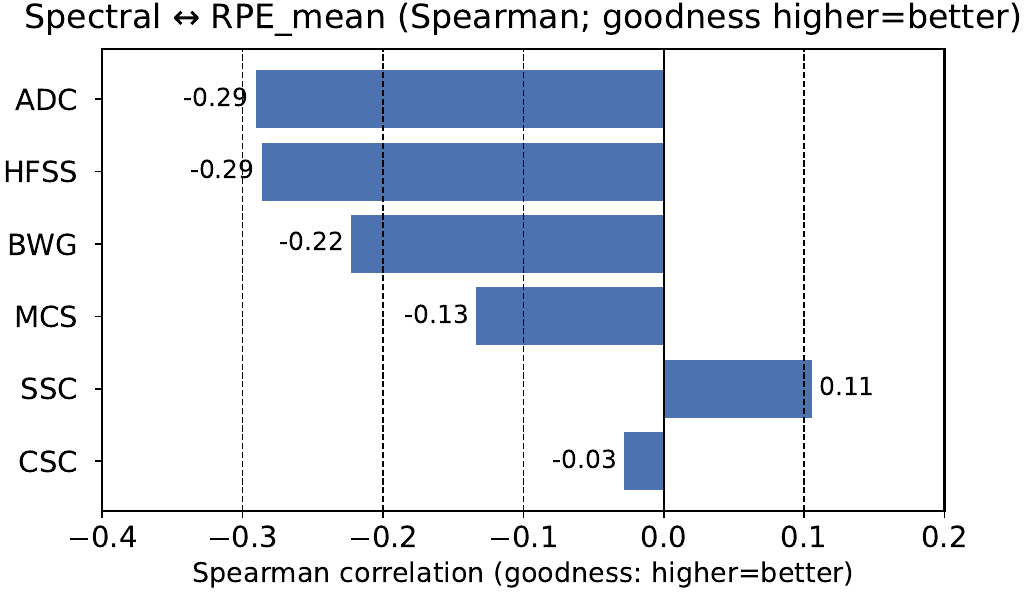}
  \hfill
  \includegraphics[width=0.49\linewidth]{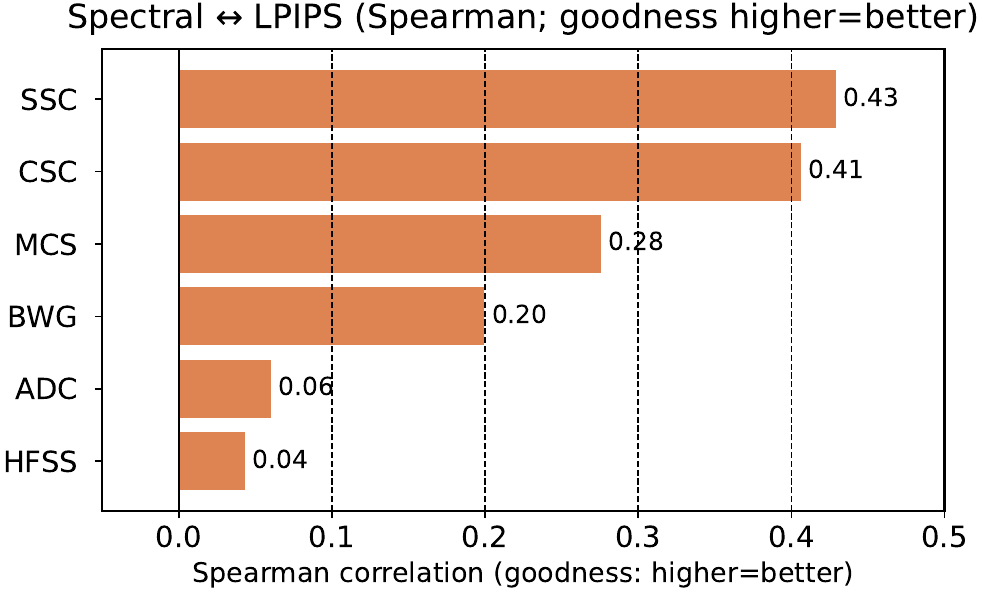}
  \vspace{-0.3em}
  \caption{Left (AG): DINO+AnyUp~\cite{wimmer2025anyup}+MASt3R~\cite{leroy2024grounding}, right (AT): DINO+AnyUp~\cite{wimmer2025anyup}+MASt3R~\cite{leroy2024grounding}.}
\end{subfigure}

\caption{
\textbf{Spearman correlations between spectral diagnostics and reconstruction quality under geometry-only (AG) and texture-only (AT) probing settings. Left panels show correlations with $-\mathrm{RPE}_{mean}$ (geometry quality), and right panels show correlations with $-\mathrm{LPIPS}$ (texture quality).}
}
\label{fig:ag-at-nsm}
\end{figure}

\begin{figure}[t]
\centering
\captionsetup[subfigure]{font=small, labelfont=bf}
% ===================== (a) =====================
\begin{subfigure}[t]{\linewidth}
  \centering
  \includegraphics[width=0.49\linewidth]{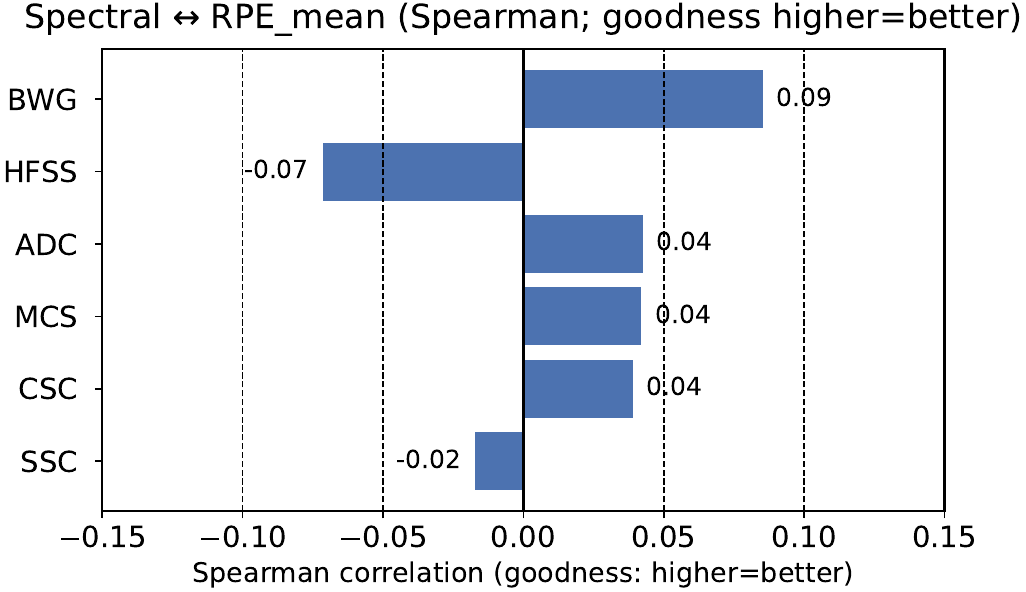}
  \hfill
  \includegraphics[width=0.49\linewidth]{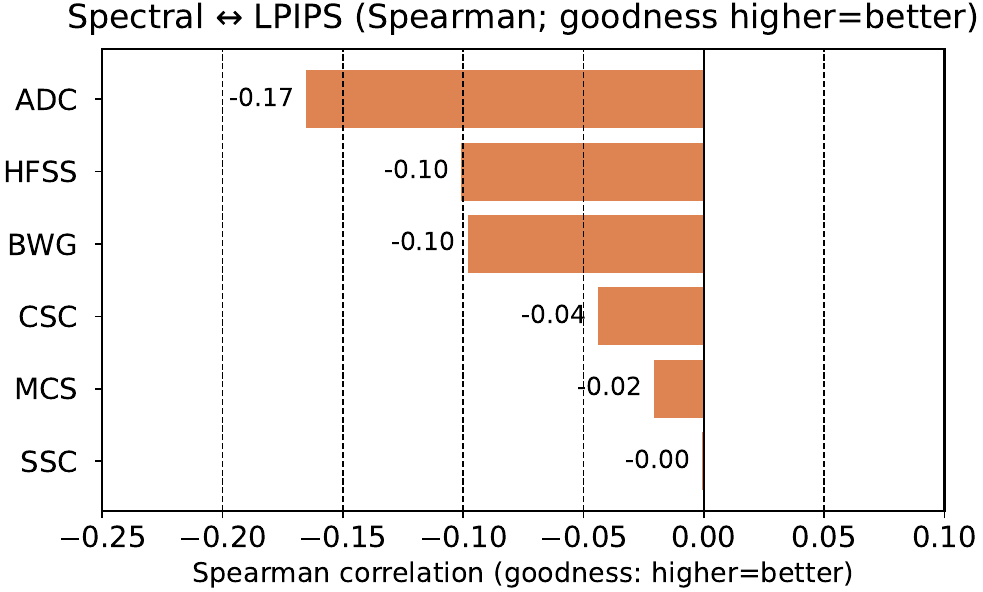}
  \vspace{-0.3em}
  \caption{Left (AG): CLIP+FeatUp~\cite{fu2024featup}+DUSt3R~\cite{wang2024dust3r}, right (AT): CLIP+FeatUp~\cite{fu2024featup}+DUSt3R~\cite{wang2024dust3r}.}
\end{subfigure}
\vspace{0.6em}
% ===================== (b) =====================
\begin{subfigure}[t]{\linewidth}
  \centering
  \includegraphics[width=0.49\linewidth]{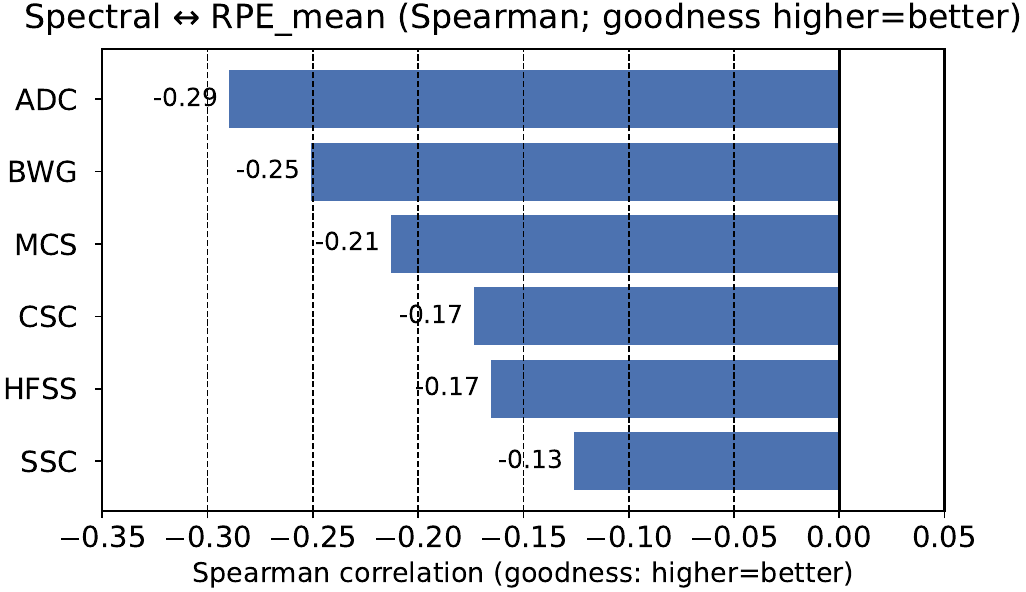}
  \hfill
  \includegraphics[width=0.49\linewidth]{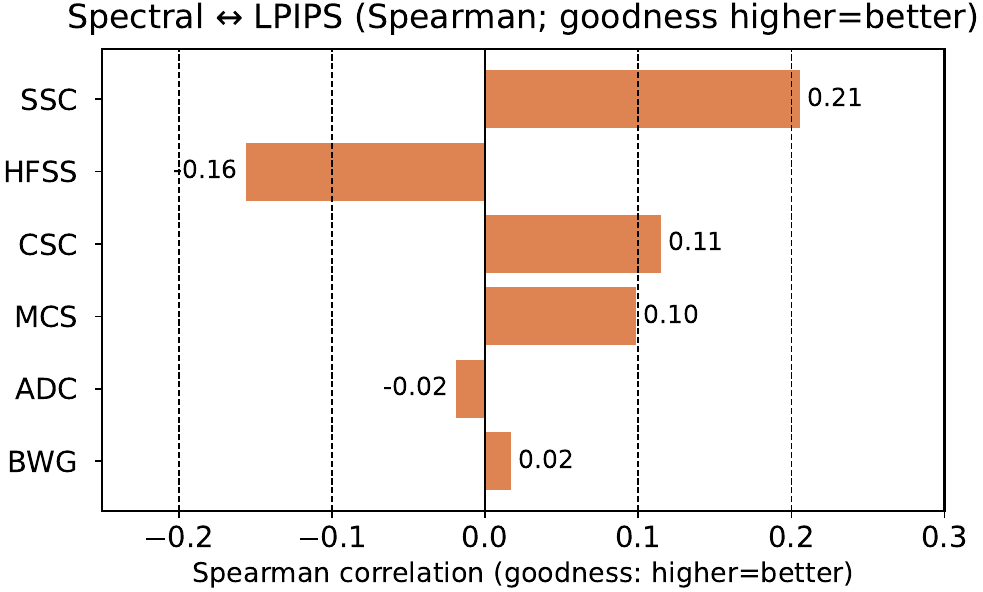}
  \vspace{-0.3em}
  \caption{Left (AG): CLIP+FeatUp~\cite{fu2024featup}+MASt3R~\cite{leroy2024grounding}, right (AT): CLIP+FeatUp~\cite{fu2024featup}+MASt3R~\cite{leroy2024grounding}.}
\end{subfigure}
\vspace{0.6em}

\caption{
\textbf{Spearman correlations between spectral diagnostics and reconstruction quality under geometry-only (AG) and texture-only (AT) probing settings. Left panels show correlations with $-\mathrm{RPE}_{mean}$ (geometry quality), and right panels show correlations with $-\mathrm{LPIPS}$ (texture quality).}
}
\label{fig:ag-at-featup-loftup}
\end{figure}

Additional geometry-only (AG) and texture-only (AT) correlation results are provided in 
Figs.~\ref{fig:ag-at-lanczos-jafar},~\ref{fig:ag-at-bilinear},~\ref{fig:ag-at-bicubic},~\ref{fig:ag-at-nsm}, and~\ref{fig:ag-at-featup-loftup}. 
These plots report Spearman correlations between spectral diagnostics and geometry-related ($-\mathrm{RPE}_{mean}$) or texture-related ($-\mathrm{LPIPS}$) reconstruction metrics across scenes. 
The results show consistent patterns with the main paper, 
the amplitude distribution metric ADC frequently exhibits positive influence gaps across multiple interpolation and learned upsampling methods, indicating stronger coupling with geometry-related reconstruction metrics. 
However, the magnitude of this effect varies depending on the backbone encoder and the 3D reconstructor. 
In contrast, the structural spectral consistency metrics (SSC and CSC) tend to influence texture fidelity slightly more than geometric accuracy but this trend can also be affected by the backbone encoder and the 3D reconstructor.

\subsection{Geometry--Texture Gap}
\begin{figure}[t]
\centering
\captionsetup[subfigure]{font=small, labelfont=bf}
% ===================== (a) =====================
\begin{subfigure}[t]{\linewidth}
  \centering
  \includegraphics[width=0.49\linewidth]{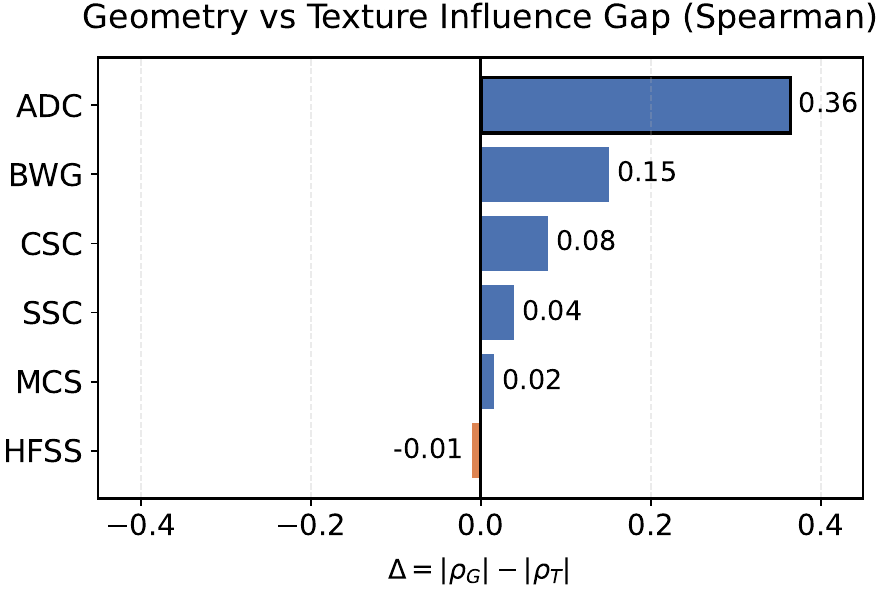}
  \hfill
  \includegraphics[width=0.49\linewidth]{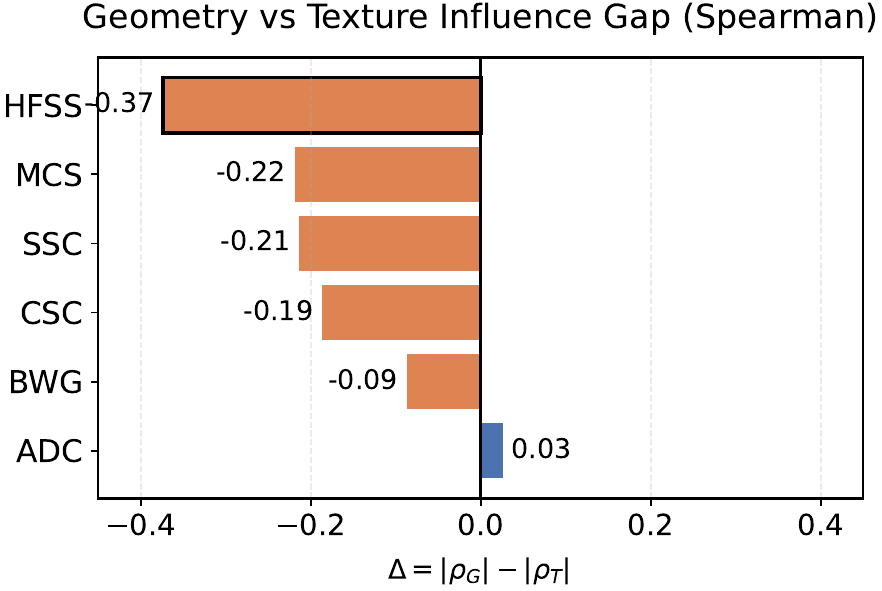}
  \vspace{-0.3em}
  \caption{Left: CLIP+Lanczos+MASt3R~\cite{leroy2024grounding}, right: DINO+Lanczos+MASt3R~\cite{leroy2024grounding}.}
\end{subfigure}
\vspace{0.6em}
% ===================== (b) =====================
\begin{subfigure}[t]{\linewidth}
  \centering
  \includegraphics[width=0.49\linewidth]{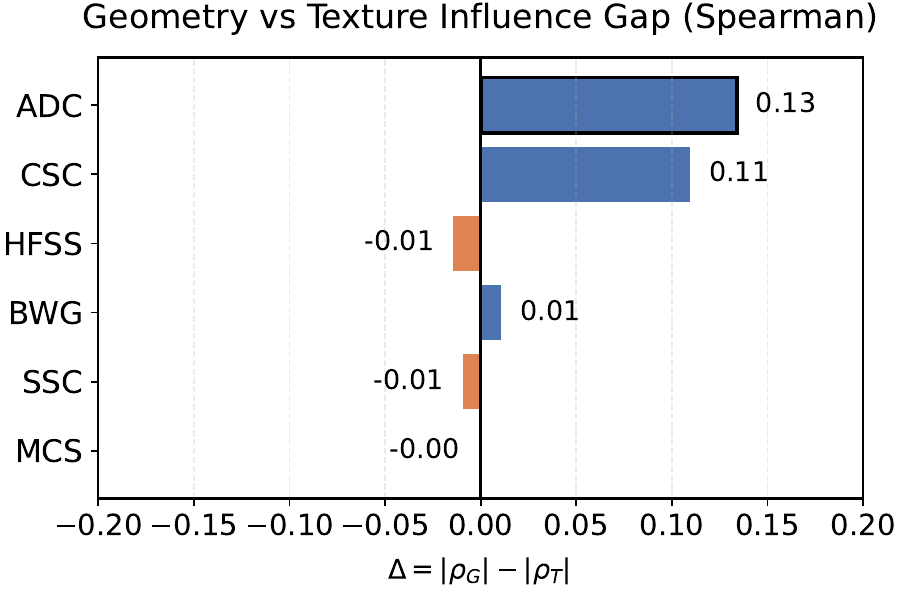}
  \hfill
  \includegraphics[width=0.49\linewidth]{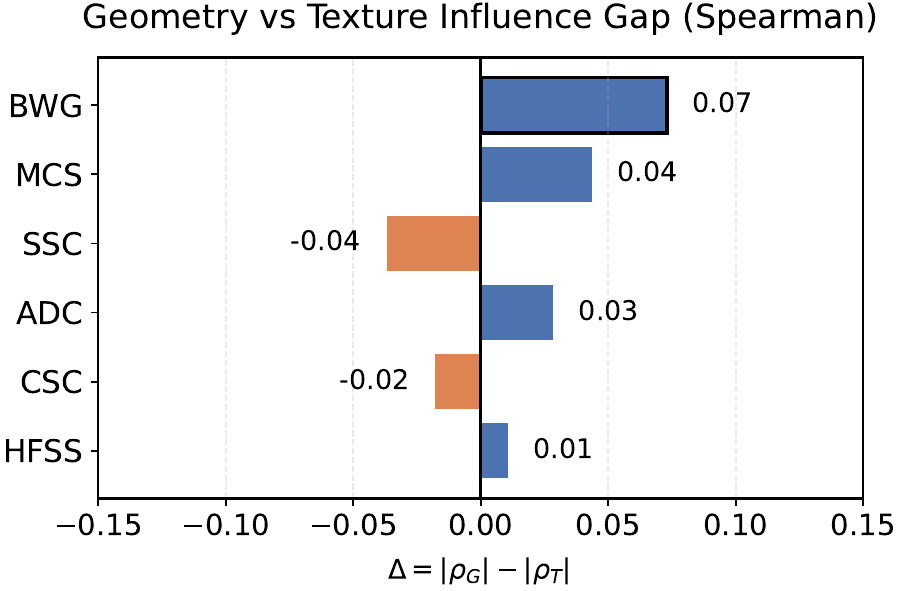}
  \vspace{-0.3em}
  \caption{Left: CLIP+JAFAR~\cite{couairon2025jafar}+MASt3R~\cite{leroy2024grounding} (AG), right: DINO+JAFAR~\cite{couairon2025jafar}+MASt3R~\cite{leroy2024grounding}.}
\end{subfigure}
\vspace{0.6em}
\begin{subfigure}[t]{\linewidth}
  \centering
  \includegraphics[width=0.49\linewidth]{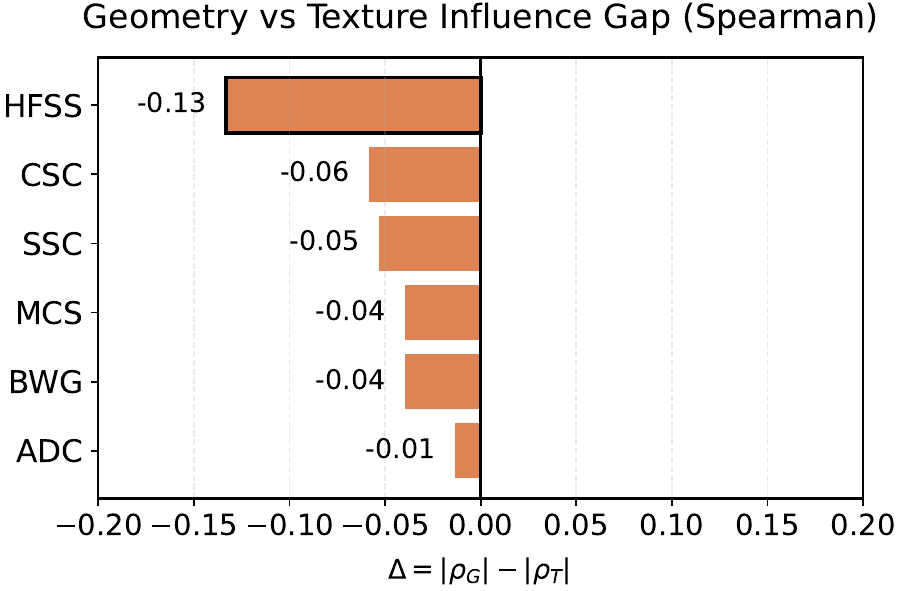}
  \hfill
  \includegraphics[width=0.49\linewidth]{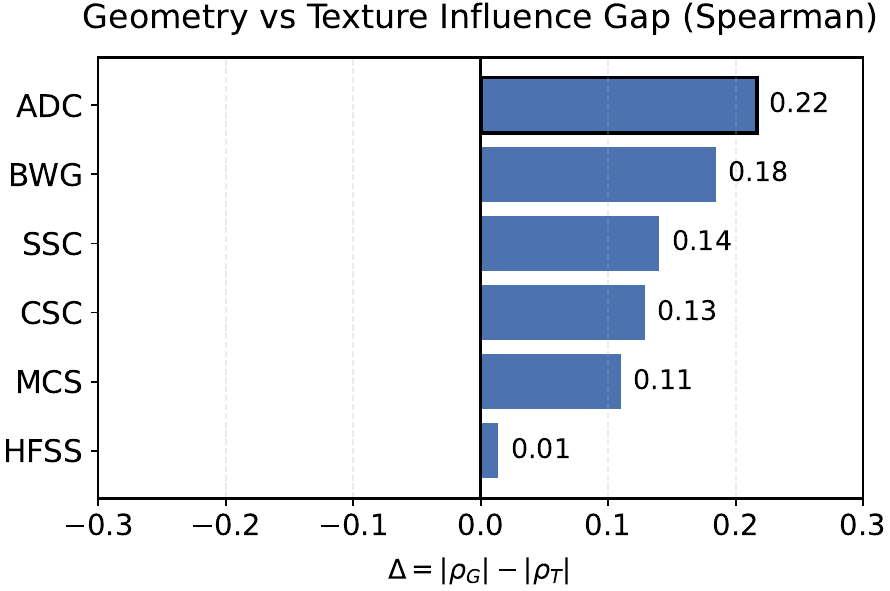}
  \vspace{-0.3em}
  \caption{Left: CLIP+Bicubic+DUSt3R~\cite{wang2024dust3r}, right: DINO+Bicubic+DUSt3R~\cite{wang2024dust3r}.}
\end{subfigure}
\vspace{0.6em}
% ===================== (b) =====================
\begin{subfigure}[t]{\linewidth}
  \centering
  \includegraphics[width=0.49\linewidth]{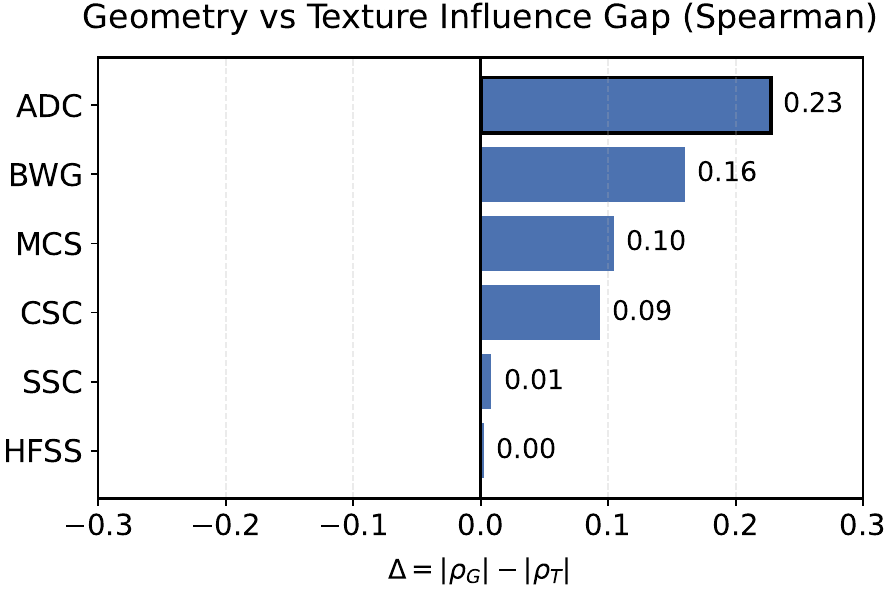}
  \hfill
  \includegraphics[width=0.49\linewidth]{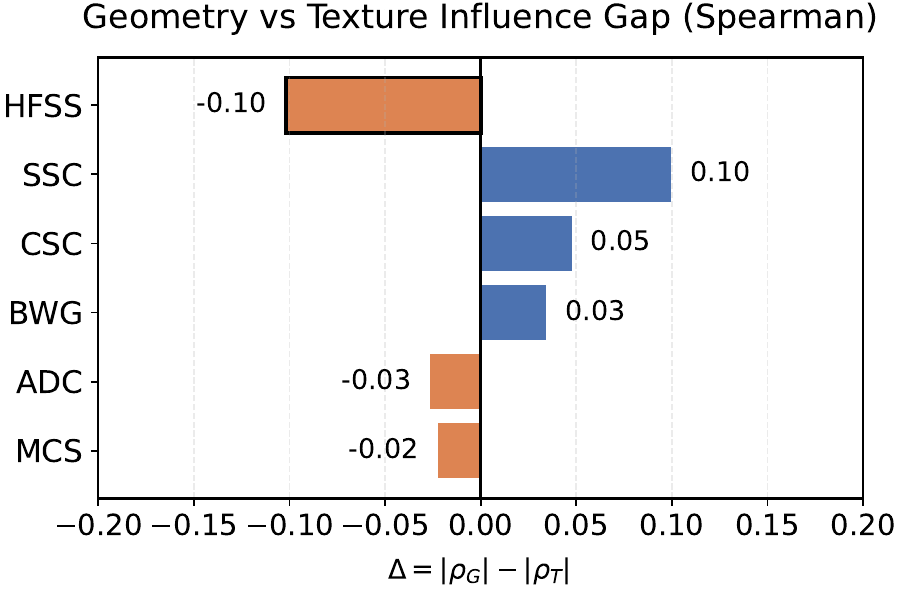}
  \vspace{-0.3em}
  \caption{Left: CLIP+Bicubic+MASt3R~\cite{leroy2024grounding}, right: DINO+Bicubic+MASt3R~\cite{leroy2024grounding}.}
\end{subfigure}
\vspace{0.6em}

\caption{
\textbf{Geometry--texture influence gap.}
}
\label{fig:gap-lanczos-jafar-mast3r}
\end{figure}

\begin{figure}[t]
\centering
\captionsetup[subfigure]{font=small, labelfont=bf}
% ===================== (a) =====================
\begin{subfigure}[t]{\linewidth}
  \centering
  \includegraphics[width=0.49\linewidth]{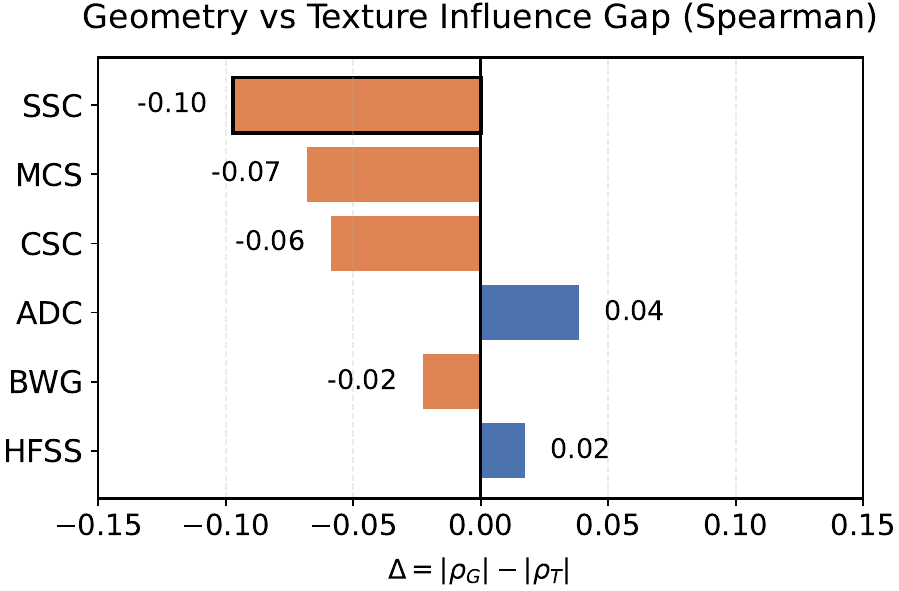}
  \hfill
  \includegraphics[width=0.49\linewidth]{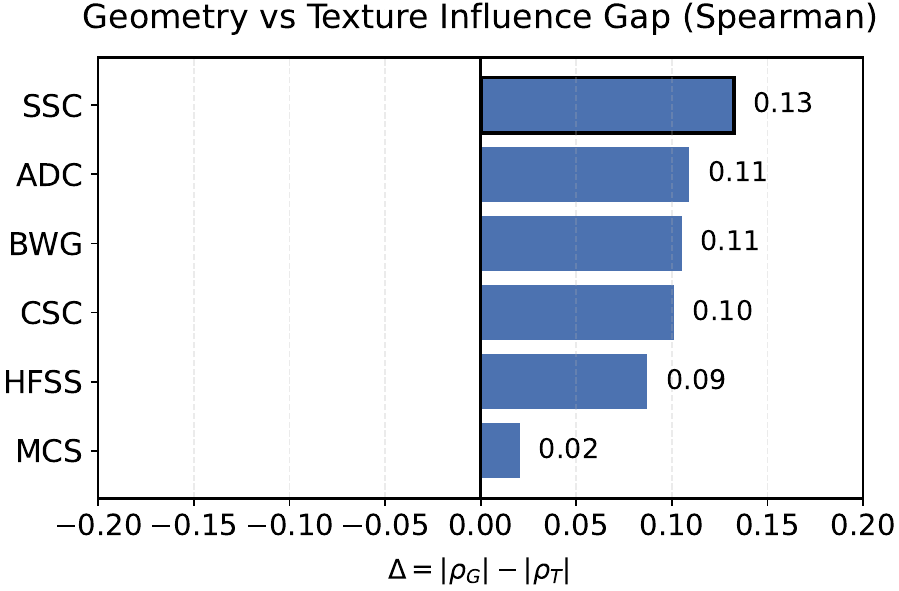}
  \vspace{-0.3em}
  \caption{Left: CLIP+Bilinear+DUSt3R~\cite{wang2024dust3r}, right: DINO+Bilinear+DUSt3R~\cite{wang2024dust3r}.}
\end{subfigure}
\vspace{0.6em}
% ===================== (b) =====================
\begin{subfigure}[t]{\linewidth}
  \centering
  \includegraphics[width=0.49\linewidth]{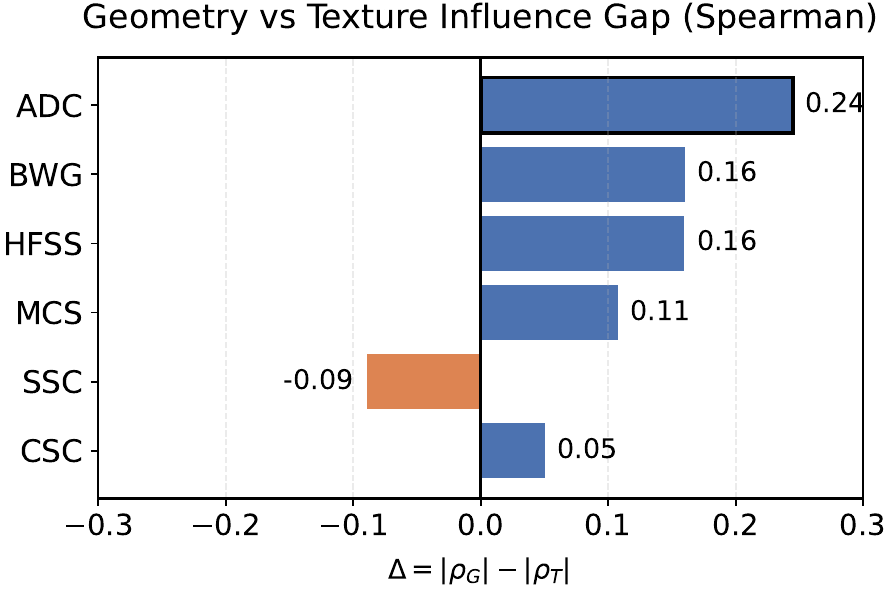}
  \hfill
  \includegraphics[width=0.49\linewidth]{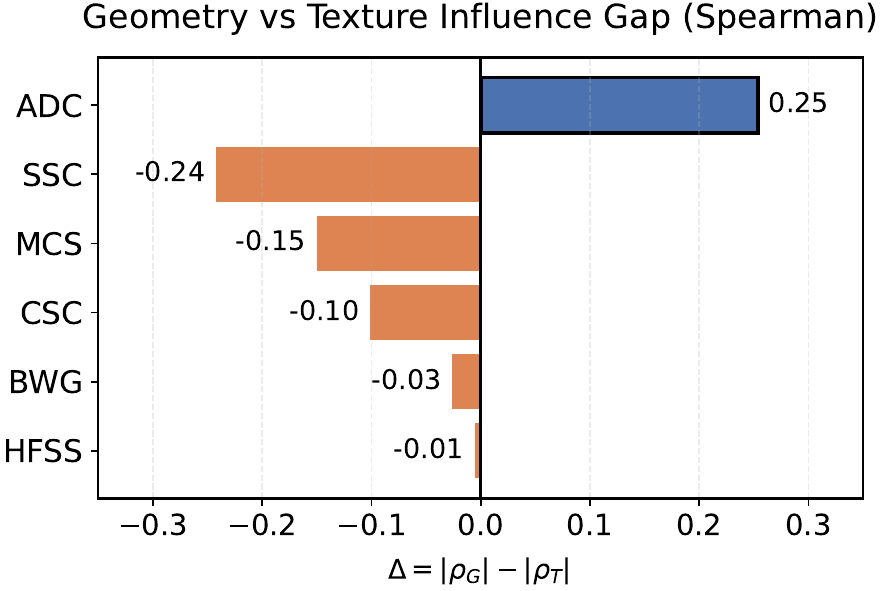}
  \vspace{-0.3em}
  \caption{Left: CLIP+Bilinear+MASt3R~\cite{leroy2024grounding}, right: DINO+Bilinear+MASt3R~\cite{leroy2024grounding}.}
\end{subfigure}
\vspace{0.6em}
% ===================== (c) =====================
\begin{subfigure}[t]{\linewidth}
  \centering
  \includegraphics[width=0.49\linewidth]{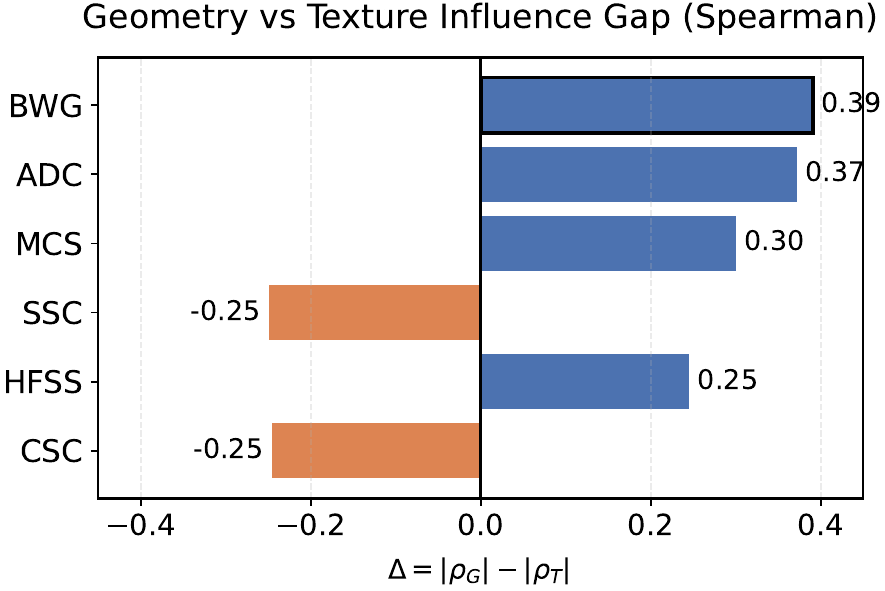}
  \hfill
  \includegraphics[width=0.49\linewidth]{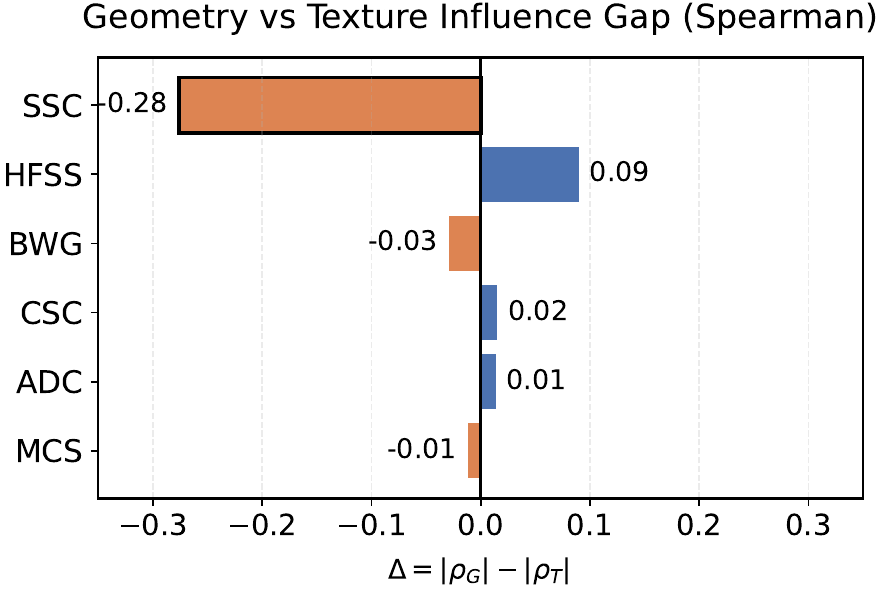}
  \vspace{-0.3em}
  \caption{Left: CLIP+NSM+DUSt3R~\cite{wang2024dust3r}, right: DINO+NSM+DUSt3R~\cite{wang2024dust3r}.}
\end{subfigure}
\vspace{0.6em}
% ===================== (d) =====================
\begin{subfigure}[t]{\linewidth}
  \centering
  \includegraphics[width=0.49\linewidth]{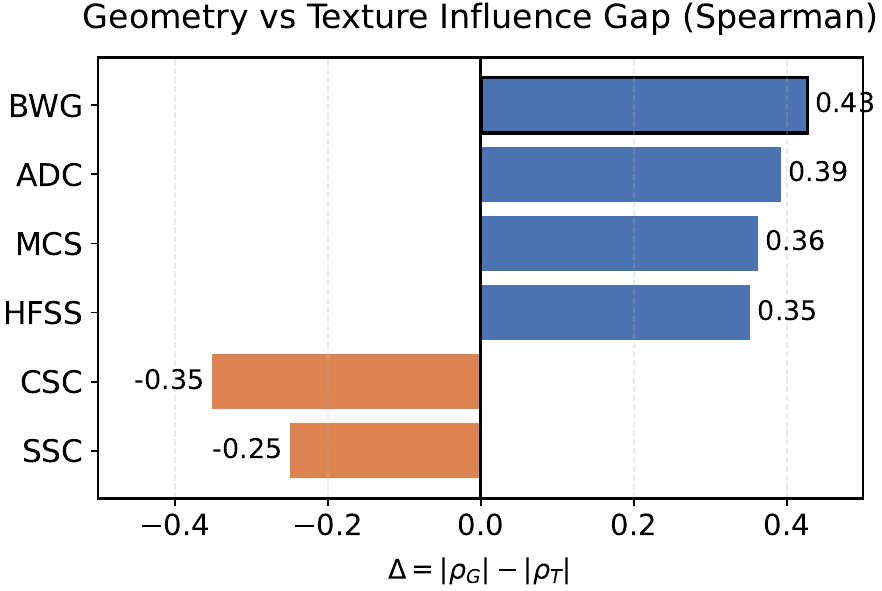}
  \hfill
  \includegraphics[width=0.49\linewidth]{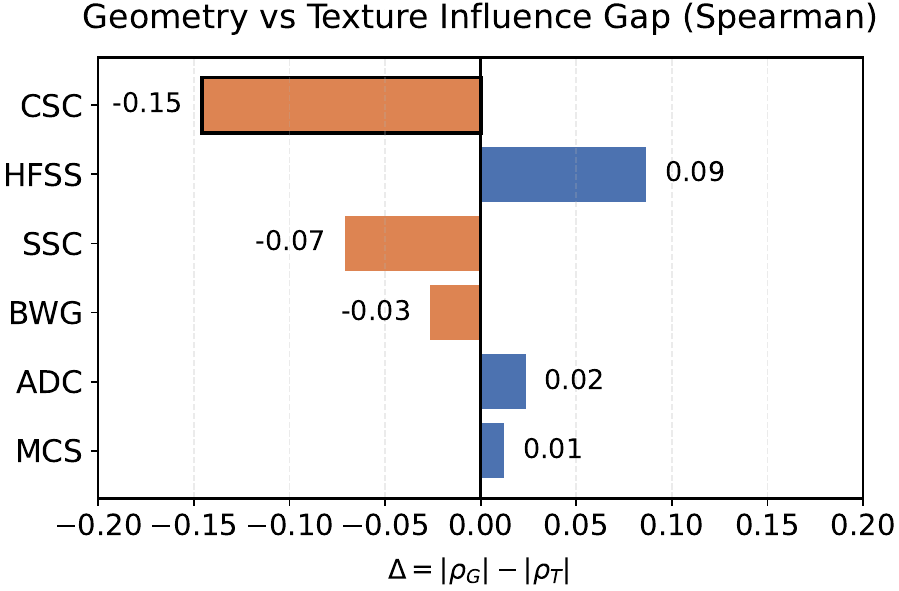}
  \vspace{-0.3em}
  \caption{Left: CLIP+NSM+MASt3R~\cite{leroy2024grounding}, right: DINO+NSM+MASt3R~\cite{leroy2024grounding}.}
\end{subfigure}

\caption{
\textbf{Geometry--texture influence gap.}
}
\label{fig:gap-bilinear-nsm}
\end{figure}

\begin{figure}[t]
\centering
\captionsetup[subfigure]{font=small, labelfont=bf}
% ===================== (a) =====================
\begin{subfigure}[t]{\linewidth}
  \centering
  \includegraphics[width=0.49\linewidth]{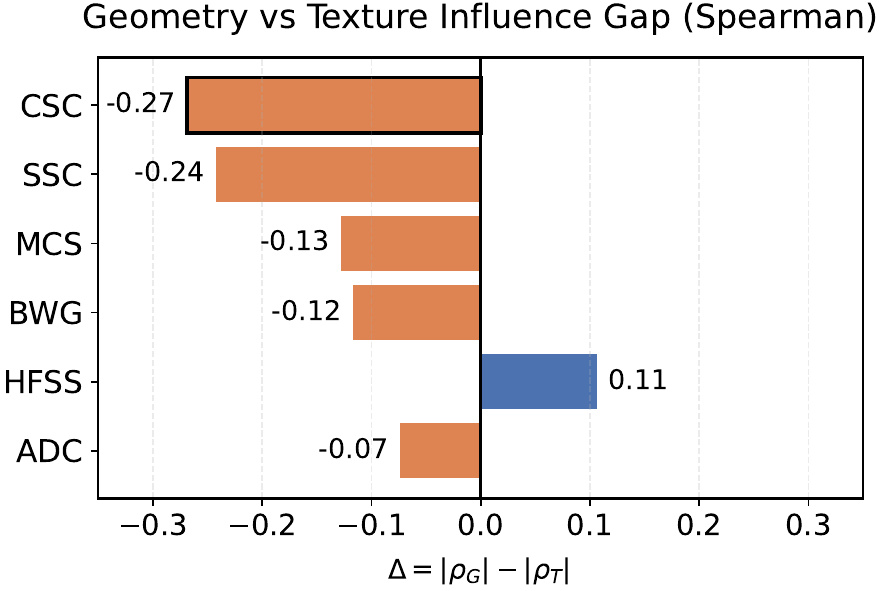}
  \hfill
  \includegraphics[width=0.49\linewidth]{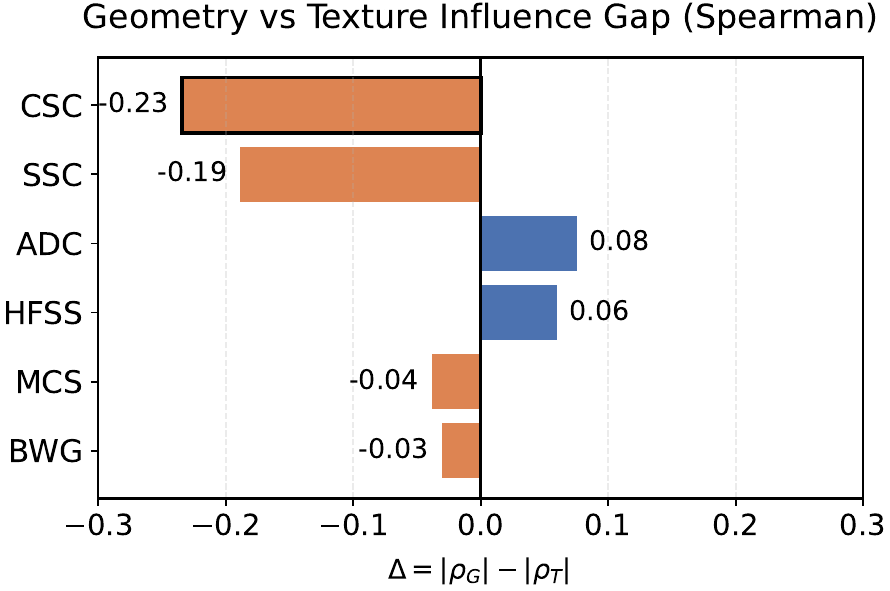}
  \vspace{-0.3em}
  \caption{Left: CLIP+AnyUp~\cite{wimmer2025anyup}+DUSt3R~\cite{wang2024dust3r}, right: DINO+AnyUp~\cite{wimmer2025anyup}+DUSt3R~\cite{wang2024dust3r}.}
\end{subfigure}
\vspace{0.6em}
\caption{
\textbf{Geometry--texture influence gap.}
}
\label{fig:gap-anyup-featup-loftup}
\end{figure}

Additional geometry--texture influence gap results are provided in 
Figs.~\ref{fig:gap-lanczos-jafar-mast3r},~\ref{fig:gap-bilinear-nsm}, and~\ref{fig:gap-anyup-featup-loftup}. 
The gap plots illustrate whether a spectral diagnostic is more strongly associated with geometry- or texture-related reconstruction metrics across scenes. 
A positive $\Delta$ indicates stronger geometry coupling, whereas a negative value indicates stronger texture coupling.

Across different upsampling strategies, the geometry--texture sensitivity varies considerably. 
Nevertheless, several consistent patterns can be observed.

First, the amplitude distribution metric ADC frequently exhibits positive gaps across multiple interpolation and learned upsampling methods, indicating stronger coupling with geometry-related reconstruction metrics. 
However, the magnitude of this effect varies depending on the backbone encoder and the 3D reconstructor.

Second, structural spectral consistency metrics (SSC and CSC) often show small or negative gaps, suggesting that structural spectral alignment tends to influence texture fidelity slightly more than geometric accuracy.

Finally, the influence gap can change substantially across different reconstructors. 
For example, the same spectral diagnostic may exhibit different magnitudes or even sign changes when switching between DUSt3R and MASt3R, suggesting that the relationship between spectral characteristics and reconstruction quality is strongly dependent on how the reconstructor utilizes feature representations.

\bibliographystyle{splncs04}
\bibliography{main}
\end{document}